%% file: main.tex
\documentclass[10pt,twocolumn,letterpaper]{article}

\usepackage[pagenumbers]{cvpr} 

\input{preamble}

\definecolor{cvprblue}{rgb}{0.21,0.49,0.74}
\usepackage[pagebackref,breaklinks,colorlinks,citecolor=cvprblue]{hyperref}

\title{Atlantis: Enabling Underwater Depth Estimation with Stable Diffusion}

\author{Fan Zhang$^1$ \quad\quad Shaodi You$^2$ \quad\quad Yu Li$^{3}$ \quad\quad Ying Fu$^1$\\
	$^1$Beijing Institute of Technology \quad $^2$University of Amsterdam\\ $^3$International Digital Economy Academy
}

\begin{document}

\twocolumn[{
	\maketitle
	\vspace{-10mm}
	\begin{center}
		\centering
		\includegraphics[width=\linewidth]{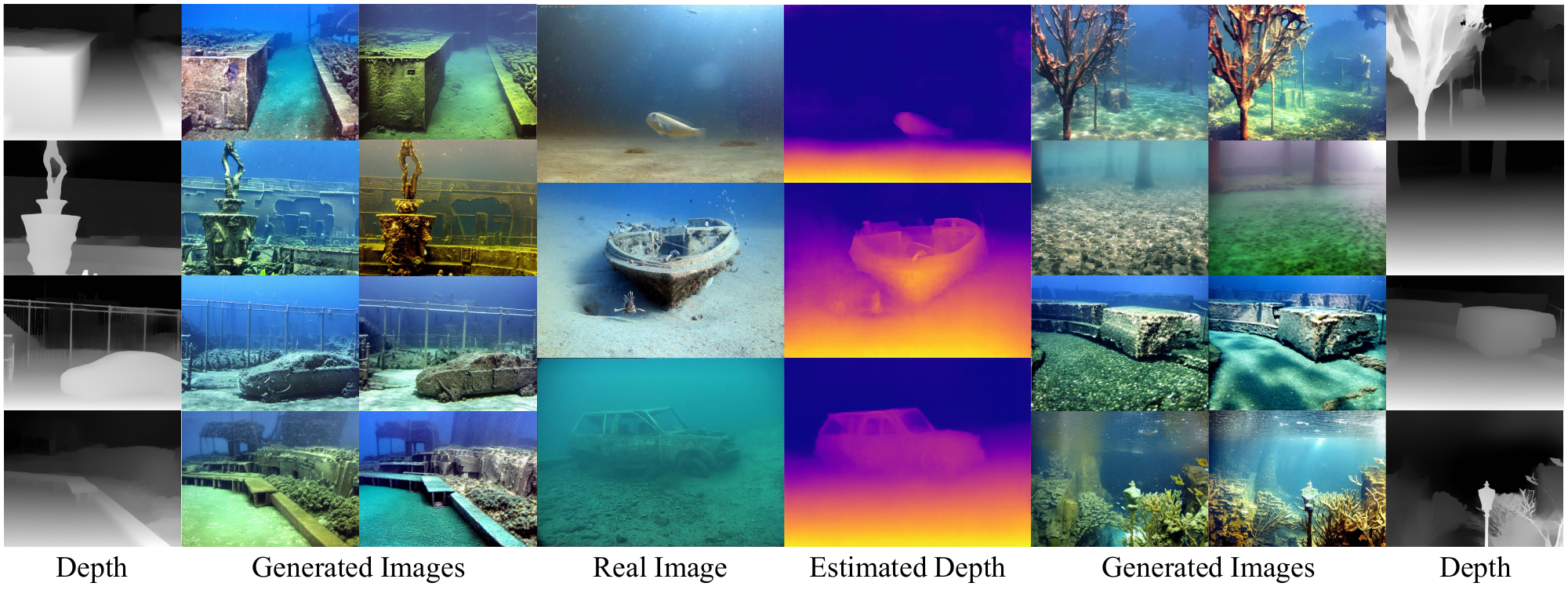}
		\vspace{-7mm}
		\captionof{figure}{The proposed method can generate various vivid non-existent underwater scenes by sampling, following the scene layout of conditioning depth map (on the left and right) for training depth estimation models. The depth model trained on the proposed dataset can well handle unseen real underwater scenes and get reliable depth maps (in the middle).
		}
		\vspace{1mm}
		\label{fig:teaser}
	\end{center}
}]
\input{sec/0_abstract}    
\input{sec/1_intro}

\input{sec/2_method}

\input{sec/3_experiment}

{
    \small
    \bibliographystyle{ieeenat_fullname}
    \bibliography{main}
}


\end{document}

%% file: preamble.tex
\usepackage[dvipsnames]{xcolor}
\usepackage{pifont}
\usepackage{multirow}
\usepackage{multicol}
\usepackage{subcaption}
\usepackage{array}
\usepackage{enumitem}
\usepackage{makecell}


%% file: sec/0_abstract.tex
\begin{abstract}
    Monocular depth estimation has experienced significant progress on terrestrial images in recent years, largely due to deep learning advancements. However, it remains inadequate for underwater scenes, primarily because of data scarcity. Given the inherent challenges of light attenuation and backscattering in water, acquiring clear underwater images or precise depth information is notably difficult and costly. Consequently, learning-based approaches often rely on synthetic data or turn to unsupervised or self-supervised methods to mitigate this lack of data. Nonetheless, the performance of these methods is often constrained by the domain gap and looser constraints. In this paper, we propose a novel pipeline for generating photorealistic underwater images using accurate terrestrial depth data. This approach facilitates the training of supervised models for underwater depth estimation, effectively reducing the performance disparity between terrestrial and underwater environments. Contrary to prior synthetic datasets that merely apply style transfer to terrestrial images without altering the scene content, our approach uniquely creates vibrant, non-existent underwater scenes by leveraging terrestrial depth data through the innovative Stable Diffusion model. Specifically, we introduce a unique Depth2Underwater ControlNet, trained on specially prepared \{Underwater, Depth, Text\} data triplets, for this generation task. Our newly developed dataset enables terrestrial depth estimation models to achieve considerable improvements, both quantitatively and qualitatively, on unseen underwater images, surpassing their terrestrial pre-trained counterparts. Moreover, the enhanced depth accuracy for underwater scenes also aids underwater image restoration techniques that rely on depth maps, further demonstrating our dataset's utility. The dataset will be publicly available at \url{https://github.com/zkawfanx/Atlantis}.
\end{abstract}

%% file: sec/1_intro.tex
\section{Introduction}
\label{sec:intro}

Occupying over two-thirds of Earth's surface, the sea is crucial for human exploration, where precise underwater depth acquisition is essential. This holds particularly true for fields such as autonomous underwater vehicles (AUV) \cite{auv1, auv2}, underwater robotics \cite{auv3}, marine biology, ecology \cite{marine3} and archaeology \cite{marine1, marine2}. Unlike costly and operationally complex active ranging equipment, such as underwater LiDARs \cite{lidar1, lidar2}, monocular depth estimation offers a more cost-effective and convenient deployment solution. Despite significant advancements in monocular depth estimation for terrestrial applications \cite{eigen2014depth, monodepth, monodepth2, dorn, midas, zoedepth}, underwater depth estimation remains challenging due to factors like light attenuation, backscatter, and water turbidity \cite{uieb2019li, squid2020berman, seathur2019akkaynak}, which lead to poor image quality and imprecise depth data. The scarcity of data hampers the training of powerful learning based models.

While some datasets like Sea-thru \cite{seathur2019akkaynak} and SQUID \cite{squid2020berman} offer real underwater data, they are costly to acquire thus are limited in scene diversity and scale. Their depth data, derived from stereo pairs or video sequences, is often sparse and not entirely reliable. GAN-based methods have emerged as an alternative, synthesizing underwater images by transferring styles from terrestrial scenes using image formation models \cite{uwcnn, depth_uwgan}. Despite they provide a remedy for the data scarcity issue because of easier acquisition and relatively larger scale and diversity, their domain gap and lack of realism limit their efficacy.

To address these challenges, our paper introduces a novel pipeline to generate underwater depth dataset, comprising diverse and realistic underwater images paired with accurate depth data. Compared to aforementioned real datasets, it is inexpensive and easy to obtain, featuring large diversity and theoretically unlimited scale.
Utilizing advancements in Stable Diffusion (SD) \cite{stablediffusion} and ControlNet \cite{controlnet2023zhang}, 
this approach allows for the generation of underwater imagery following the scene structure and layout of terrestrial depth. Despite their widespread applications in AI-generated content, these technologies have rarely been used for generating training data. We present a dataset that combines the accuracy of terrestrial depth with the lifelike depiction of underwater scenes, offering a robust resource for training reliable depth estimation models for unseen underwater scenes. This dataset not only serves as a bridge between terrestrial and underwater domains but also demonstrates its utility in image restoration techniques like Sea-Thru \cite{seathur2019akkaynak}.

Specifically, we first construct a dataset comprising underwater images, estimated depths, and captions that describe the image content. Then we train a \textit{Underwater2Depth} ControlNet targeting realistic underwater image generation using depth map. Using the pretrained SD and our trained ControlNet, we generate an underwater depth dataset comprising realistic underwater images and accurate depth, enabling the training of terrestrial depth estimation models for underwater depth estimation. Compared the terrestrial counterparts of KITTI \cite{kitti} and NYU Depthv2 \cite{nyudepthv2}, their performance are largely improved both quantitatively and qualitatively. We also show the value of our dataset in applying the trained depth model for underwater image enhancement using Sea-thru algorithm \cite{seathur2019akkaynak}. It is important to note that our goal is not necessarily to surpass the results of robust terrestrial depth models trained with abundant mixed sourced data and training tricks \eg, MiDaS \cite{midas} and ZoeDepth \cite{zoedepth} on underwater scenes, but to enable existing depth models on underwater scenes with our data and simple training.

To summarize, our contributions are three-fold:
\begin{itemize}
	\item We are the first, to the best of our knowledge, proposing to construct paired dataset for underwater depth estimation training, utilizing newly emerged SD and ControlNet.
	
	\item The proposed dataset, Atlantis, is easy to collect and extend, comprising realistic underwater images and reliable depth, and featuring large diversity and theoretically unlimited scale.
	
	\item We propose to improve the performance of existing depth models on unseen underwater scenes using our proposed dataset for training. The improved depth can further be applied for underwater image enhancment, which highlighting the effectiveness and utility of our dataset.
\end{itemize}

\section{Related Work}
\label{sec:related}

In the evolving field of monocular depth estimation, significant strides have been made through diverse methodologies. This section reviews key developments in terrestrial monocular depth estimation, explores current underwater depth estimation techniques, and introduces methods integrating underwater depth estimation with image restoration.

\subsection{Terrestrial Depth Estimation}

Eigen \etal \cite{eigen2014depth} pioneered the coarse-to-fine network approach for end-to-end monocular depth estimation, a significant breakthrough. Their Scale-Invariant log loss is widely adopted in subsequent methods. Monodepth \cite{monodepth} and Monodepth2 \cite{monodepth2} achieved impressive self-supervised performance and robustness. DORN \cite{dorn} and Adabins \cite{adabins2021bhat} represent methods that treat depth estimation as ordinal regression or classification, discretizing depth. Recently, MiDaS \cite{midas} set a new benchmark for robust zero-shot depth estimation using multi-source mixed data training and various optimization techniques. DPT \cite{dpt} and ZoeDepth \cite{zoedepth} further enhanced performance in relative and absolute depth metrics, respectively. NeWCRFs \cite{newcrfs2022yuan} and iDisc \cite{idisc2023piccinelli} introduced fully-connected CRFs and an Internal Discretization module, respectively, for depth estimation. However, these models' performance in underwater scenes is limited due to domain gaps and data scarcity.

\subsection{Underwater Depth Estimation}

Underwater, light attenuation and backscatter depends on the distance light travels through water. Image formation models \cite{duntley1963light, jaffe1990computer, mcglamery1980computer, seathur2019akkaynak}  that elucidate these relationships aid in estimating parameters such as attenuation coefficients and transmission.  Intriguingly, depth information often emerges as a secondary product of this process. Traditional techniques of DCP family \cite{dcp, udcp}, therefore, can estimate depth. Gupta and Mitra \cite{uwnet2019gupta} proposed UW-Net that utilizes the GAN for unsupervised training. Li \etal \cite{uwcnn} and Hambarde \etal \cite{depth_uwgan} proposed to synthesize different types of underwater images using the image formation model \cite{chiang2011underwater} and NYU Depthv2 \cite{nyudepthv2}, focusing on image enhancement and depth estimation, respectively. Recent work has also explored lightweight models \cite{udepth2023yu} and self-supervised learning \cite{depth_self2022yang, depth_self2023amitai}. Despite their effectiveness, these methods still lag behind terrestrial models in performance, underscoring the need for novel datasets that enable the training of powerful terrestrial modeling techniques.

\subsection{Underwater Image Enhancement}

Unlike underwater depth estimation, underwater image enhancement has been an actively investigated field since the era of traditional techniques, focusing on color correction, contrast enhancement, and backscatter removal. Early methods predominantly relied on physical models and handcrafted priors \cite{dcp, udcp, seathur2019akkaynak}, often integrating depth-related aspects. Recent learning-based methods \cite{depth_uie2016drews, depth_uie2018song, depth_self2023varghese} have shown a preference for jointly estimating underwater depth and image recovery. A notable advancement is Akkaynak and Treibitz's revised image formation model \cite{ifm2018akkaynak} and their Sea-thru algorithm \cite{seathur2019akkaynak}, which achieves effective de-watering results using range maps. Our dataset's potential to enhance depth estimation performance is validated through its application in the Sea-thru algorithm.

%% file: sec/2_method.tex
\section{Method}
\label{sec:method}

\begin{figure*}[ht]\small
	\centering
	\includegraphics[width=1\linewidth]{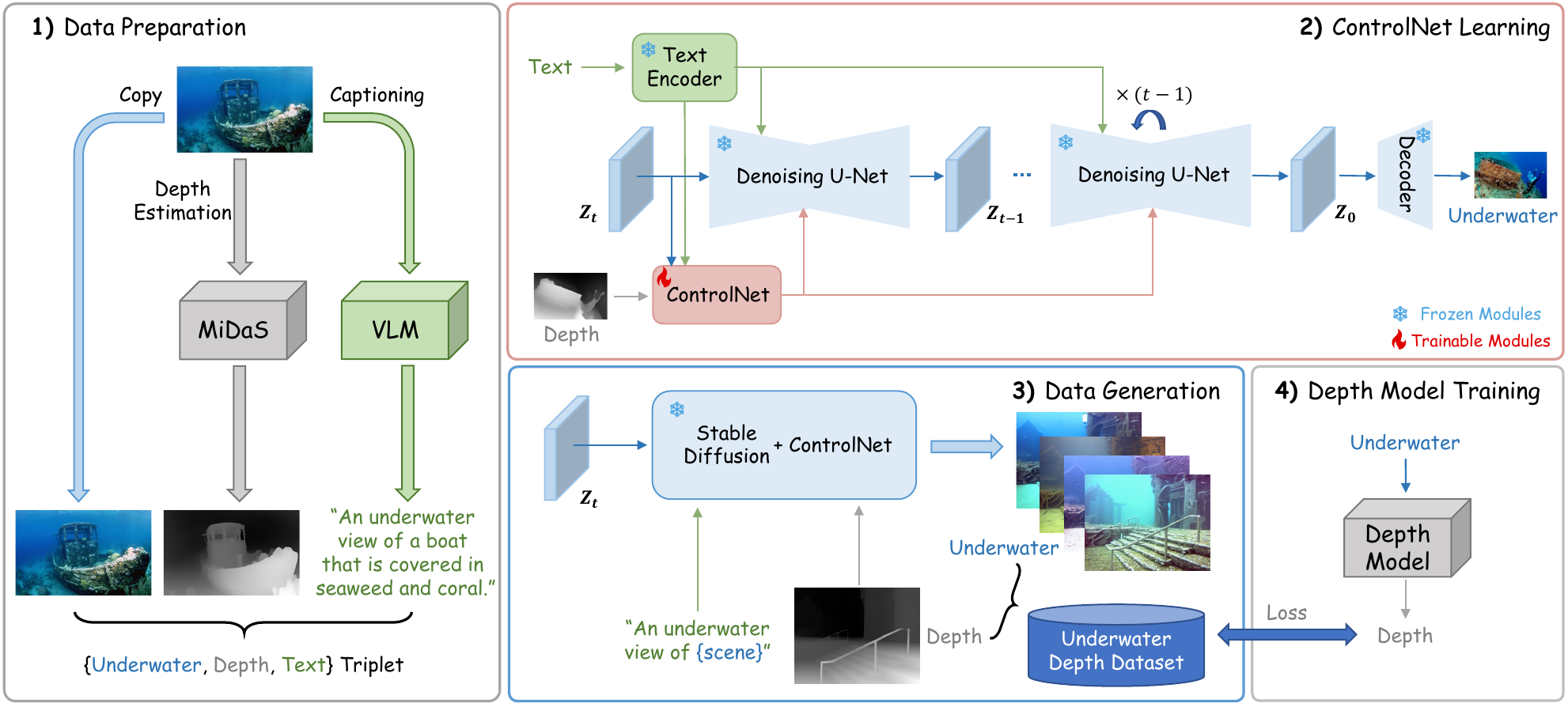}
	\vspace{-5mm}
	\caption{Overview of our method for generating the underwater depth dataset. The process begins by creating an intermediate dataset containing \textcolor{NavyBlue}{underwater} images, \textcolor{gray}{depth} maps, and \textcolor{OliveGreen}{text} descriptions. This dataset is used to train the \textit{Depth2Underwater} ControlNet for generating underwater images from depth maps. The resulting dataset facilitates the training and performance improvement of terrestrial depth models for unseen underwater scenes, as well as application to underwater image enhancement.}
	\label{fig:method}
	\vspace{-5mm}
\end{figure*}

In this section, we first detail the motivation, then introduce our pipeline for data generation as depicted in Figure \ref{fig:method}.
\begin{figure}[t]\small
	\centering
	\includegraphics[width=\linewidth]{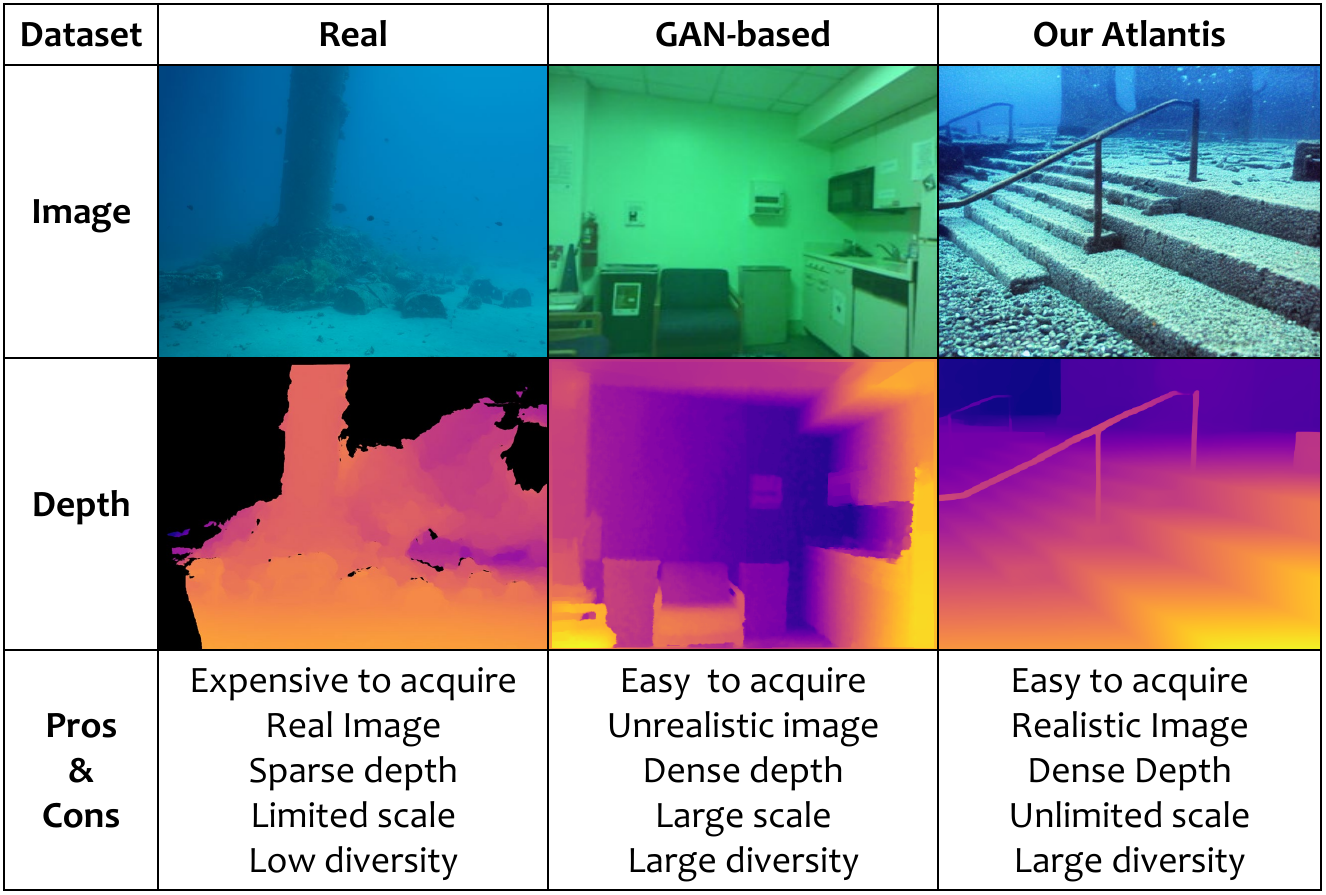}
	\vspace{-6mm}
	\caption{Comparisions of real dataset \cite{squid2020berman}, GAN-based synthetic dataset \cite{depth_uwgan} and ours proposed underwater depth dataset Atlantis.}
	\label{fig:motivation}
	\vspace{-5mm}
\end{figure}

\subsection{Motivation}
In the pursuit of accurate underwater depth estimation, one of the primary challenges is the labor-intensive and complex task of collecting real underwater data, including both imagery and precise depth information. Existing datasets like Sea-thru \cite{seathur2019akkaynak} and SQUID \cite{squid2020berman}, although valuable, are limited in the diversity of scenes and in scale. The depth data obtained from stereo pairs in these datasets is often sparse and compromised in reliability due to the inherently low quality of underwater images.

As an alternative, GAN-based methods have been utilized to synthesize underwater images by transferring the style of terrestrial images, combining terrestrial depth and image formation models, aiming to alleviate the scarcity of real underwater data. However, this approach, while being less costly and in larger diversity and scale, typically results in unrealistic synthetic images with significant domain gap, as the transformation is more akin to style transfer than to the creation of authentic underwater scenes.

This is where our proposed dataset comes into play. We offer a solution that generates vivid, non-existent underwater scenes using only depth maps and textual descriptions. This approach not only provides an infinite range of sampling possibilities but also ensures the ease of depth map acquisition. The resulting images exhibit a smaller domain gap compared to traditional methods (Section \ref{sec:domain}). Our dataset, therefore, stands out for its advantages in terms of easy acquisition, diversity and scale, realism, and practicality, marking a significant improvement over existing datasets and synthesized underwater imagery methods.

\subsection{Underwater Depth Dataset: Atlantis}
In the creation of our underwater depth dataset, illustrated in Figure \ref{fig:method}, we initiate by constructing an intermediary dataset that is instrumental in training a specialized ControlNet \cite{controlnet2023zhang}. This tailored ControlNet is then utilized to guide the pretrained Stable Diffusion v1.5 \cite{stablediffusion} in generating underwater images informed by outdoor depth maps.
\vspace{-2mm}

\paragraph{Data Preparation.} Our process begins with the utilization of the robust MiDaS \cite{midas} model to estimate inverse relative depth for images from the UIEB dataset \cite{uieb2019li}, following ControlNet \cite{controlnet2023zhang} procedure. For each underwater image $U$,  a corresponding depth map $D$ is obtained as follows:
\begin{equation}
	D = \mathcal{F}_{MiDaS}(U),
\end{equation}
where $\mathcal{F}_{MiDaS}$ denotes the pretrained MiDaS model.
Additionally, each image $U$ undergoes captioning using the pretrained BLIP2 model \cite{blip2} to generate descriptive text $T$:
\begin{equation}
	T = \mathcal{F}_{BLIP2}(U).
\end{equation}
This leads to the formation of our intermediate dataset, comprising \{\textcolor{NavyBlue}{Underwater}, \textcolor{gray}{Depth}, \textcolor{OliveGreen}{Text}\} triplets.
Here, the depth map $D$ serves as the conditioning input, with $U$ as the target image and $T$ providing the textual narrative for SD's content generation. During the training stage, only ControlNet is set as trainable and other parts of SD is freezed in the whole process.
\vspace{-4mm}

\paragraph{Data Generation.} Post training our \textit{Depth2Underwater} ControlNet, we can now generate underwater images based on provided depth maps. For instance, with a text prompt \textcolor{OliveGreen}{\textit{``an underwater view of Atlantis"}} and a corresponding outdoor depth map $D$, a vivid non-existent underwater scene is created. The process is as follows:
\begin{equation}
	\textbf{c} = \mathcal{F}_{CtrlNet}(\textbf{z}_t, D, T),
\end{equation}
where $\mathcal{F}_{CtrlNet}$ represents our trained ControlNet and $\textbf{c}$ is conditioning feature extracted from the depth map. $t$ denotes the $t$-th step of the backward diffusion process. This feature $\textbf{c}$ is then utilized in the SD generation process:
\begin{equation}
	\bar{U} = \mathcal{F}_{SD}(\textbf{z}_t, T | \textbf{c}),
\end{equation}
yielding the generated underwater image $\bar{U}$. $\mathcal{F}_{SD}(\cdot|\cdot)$ denotes the generation process of pretrained SD conditioned on a ControlNet. This methodology allows for the creation of a diverse array of underwater images, all adhering to the predetermined scene structure but with varied appearances.
\vspace{-5mm}
\paragraph{Underwater Depth Dataset.} The final dataset is produced by conditioning the generation process of the pretrained SD model with our \textit{Depth2Underwater} ControlNet. Utilizing 400 terrestrial images from the DIODE-outdoor dataset \cite{diode} for depth estimation, we employ text prompts such as \textcolor{OliveGreen}{\textit{``an underwater view of Atlantis"}} and \textcolor{OliveGreen}{\textit{``a corner of lost Atlantis"}} to guide the generation of unique underwater scenes. Sampling four times for each prompt and depth map results in a dataset comprising 3,200 data pairs. This dataset is pivotal in training and enhancing the performance of state-of-the-art terrestrial depth estimation models, particularly for unseen underwater scenes. The final output is an estimated depth map $D'$ for any given unseen underwater image $U'$:
\begin{equation}
	D' = \mathcal{F}_{Depth}(U'),
\end{equation}
where $\mathcal{F}_{Depth}$ denotes the depth estimation model trained on our dataset.

\begin{table*}[t]\footnotesize
	\caption{Quantitative comparisons on real underwater images from D3 and D5 subsets of Sea-thru dataset \cite{seathur2019akkaynak}.}
	\label{tab:quantitative1}
	\centering
	\vspace{-2mm}
	\begin{tabular}{lccccccccc}
		\toprule
		Models         & $RMSE$$\downarrow$    &  $RMSE_{log}$$\downarrow$ & $A.Rel$$\downarrow$   & $S.Rel$$\downarrow$    & $log_{10}$$\downarrow$& $SI_{log}$$\downarrow$ & $\delta_1$$\uparrow$     & $\delta_2$$\uparrow$     & $\delta_3$$\uparrow$     \\ \midrule
		IDisc-KITTI        & 5.891          & 1.192          & 4.702          & 44.288         & 0.489          & 35.846          & 0.093          & 0.241          & 0.359          \\
		IDisc-NYUDepthv2   & 3.144          & 0.845          & \textbf{0.819} & \textbf{2.471} & 0.338          & 37.296          & 0.215          & 0.403          & 0.504          \\
		IDisc-Atlantis     & \textbf{1.371} & \textbf{0.354} & 1.630          & 14.279         & \textbf{0.109} & \textbf{34.654} & \textbf{0.553} & \textbf{0.850} & \textbf{0.955} \\ \midrule
		NewCRFs-KITTI      & 3.251          & 0.934          & 2.874          & 15.768         & 0.365          & 42.341          & 0.213          & 0.375          & 0.465          \\
		NewCRFs-NYUDepthv2 & 3.390          & 0.955          & \textbf{0.770} & \textbf{2.350} & 0.372          & 47.667          & 0.179          & 0.365          & 0.479          \\
		NewCRFs-Atlantis   & \textbf{1.435} & \textbf{0.378} & 1.683          & 14.764         & \textbf{0.120} & \textbf{37.101} & \textbf{0.476} & \textbf{0.837} & \textbf{0.952} \\ \bottomrule
	\end{tabular}
	\vspace{-3mm}
\end{table*}

\begin{table*}[t]\footnotesize
	\caption{Quantitative comparisons on real underwater images from SQUID dataset \cite{squid2020berman}.}
	\label{tab:quantitative2}
	\centering
	\vspace{-2mm}
	\begin{tabular}{lccccccccc}
		\toprule
		Models         & $RMSE$$\downarrow$    &  $RMSE_{log}$$\downarrow$ & $A.Rel$$\downarrow$   & $S.Rel$$\downarrow$    & $log_{10}$$\downarrow$& $SI_{log}$$\downarrow$ & $\delta_1$$\uparrow$     & $\delta_2$$\uparrow$     & $\delta_3$$\uparrow$     \\ \midrule
		IDisc-KITTI        & 7.265          & 0.736          & 1.039          & 8.040          & 0.289          & 35.827          & 0.156          & 0.349          & 0.555          \\
		IDisc-NYUDepthv2   & 8.752          & 1.638          & 0.737          & 6.454          & 0.683          & 41.097          & 0.016          & 0.046          & 0.093          \\
		IDisc-Atlantis     & \textbf{2.663} & \textbf{0.277} & \textbf{0.249} & \textbf{0.920} & \textbf{0.094} & \textbf{27.221} & \textbf{0.637} & \textbf{0.900} & \textbf{0.960} \\ \midrule
		NewCRFs-KITTI      & 6.692          & 0.779          & 0.579          & 3.930          & 0.294          & 52.091          & 0.197          & 0.381          & 0.541          \\
		NewCRFs-NYUDepthv2 & 8.957          & 1.764          & 0.766          & 6.740          & 0.734          & 46.791          & 0.013          & 0.029          & 0.064          \\
		NewCRFs-Atlantis   & \textbf{2.563} & \textbf{0.256} & \textbf{0.229} & \textbf{0.830} & \textbf{0.088} & \textbf{25.189} & \textbf{0.675} & \textbf{0.902} & \textbf{0.964} \\ \bottomrule
	\end{tabular}
	\vspace{-3mm}
\end{table*}

\begin{figure*}[!t]\small
	\centering
	\setlength{\tabcolsep}{1pt}
	\renewcommand{\arraystretch}{0.5}
	\begin{tabular}{ccccccc}
		\includegraphics[width=0.14\linewidth]{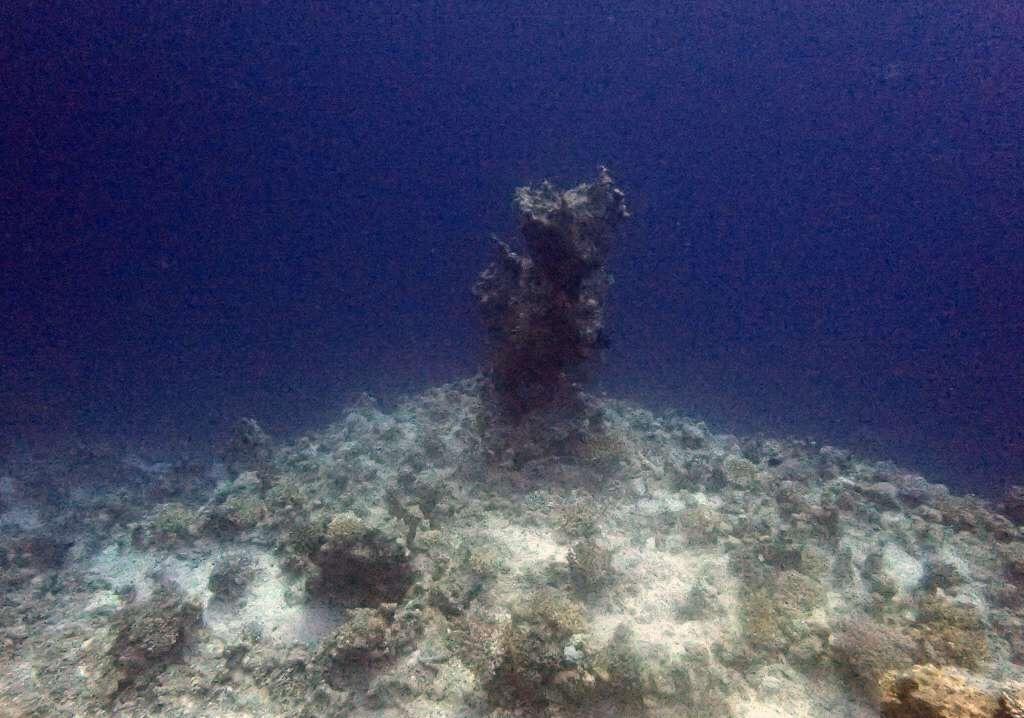} &
		\includegraphics[width=0.14\linewidth]{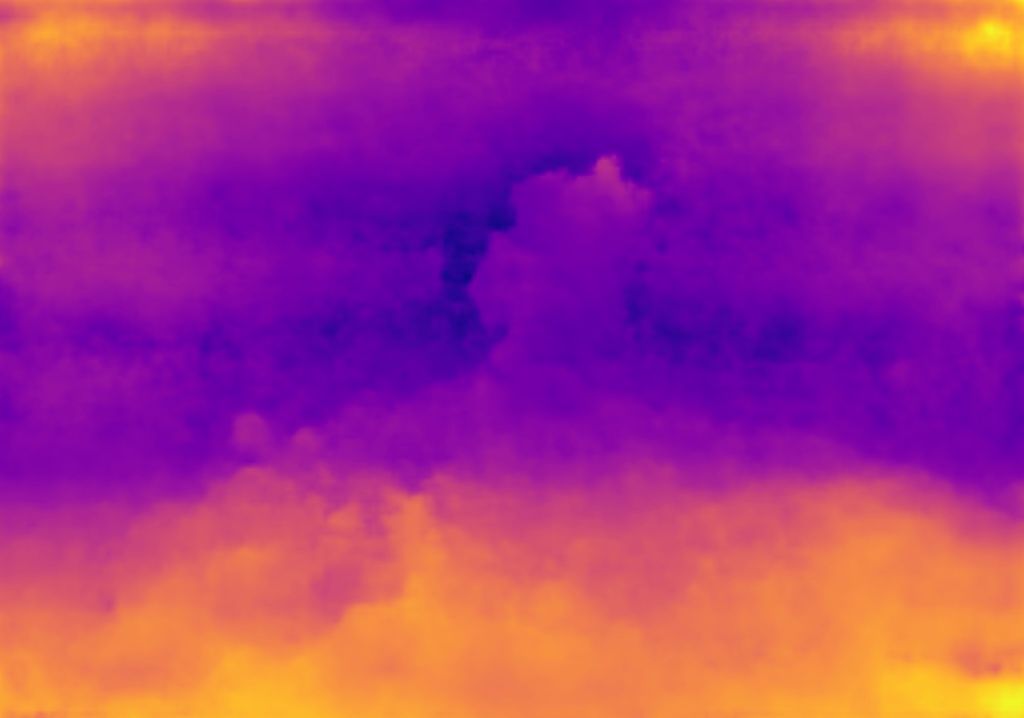} & 
		\includegraphics[width=0.14\linewidth]{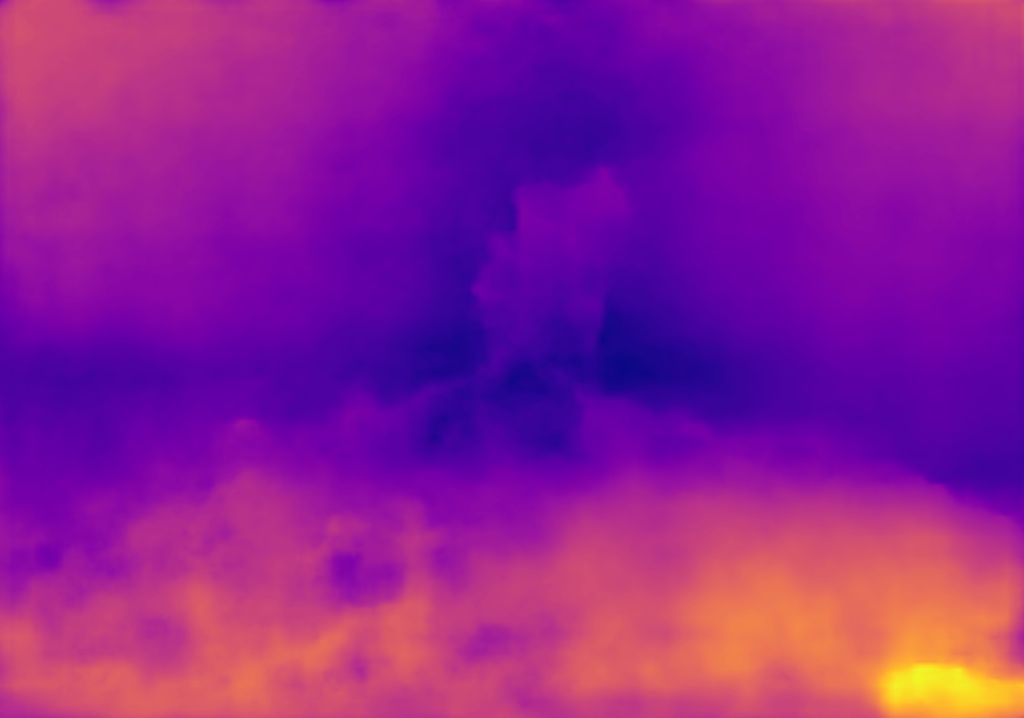} & 
		\includegraphics[width=0.14\linewidth]{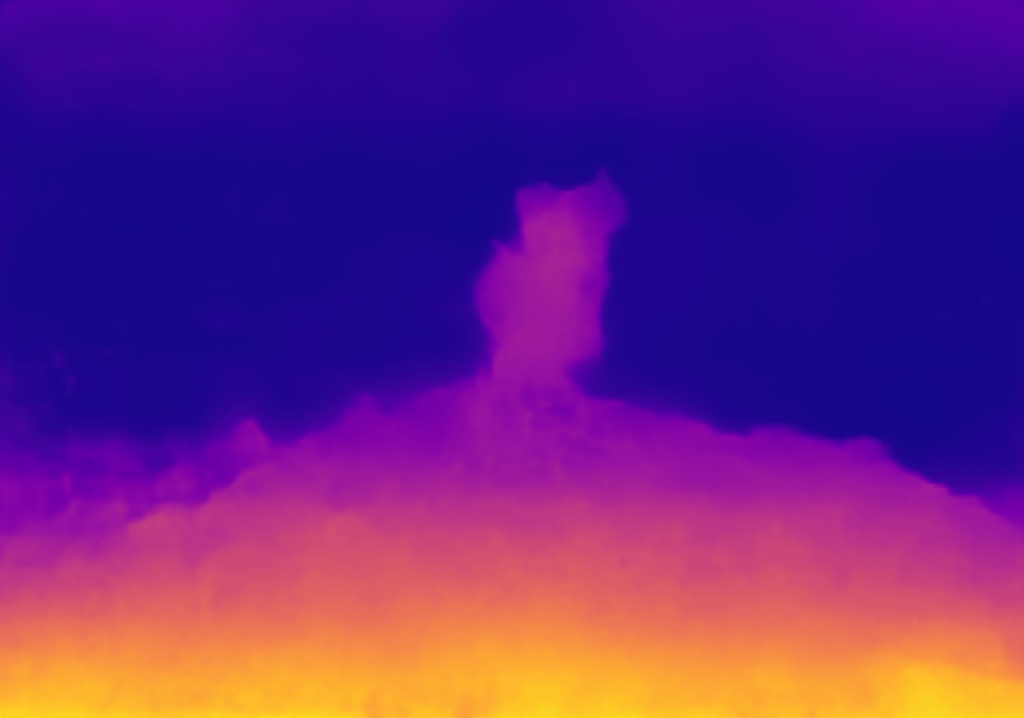} & 
		\includegraphics[width=0.14\linewidth]{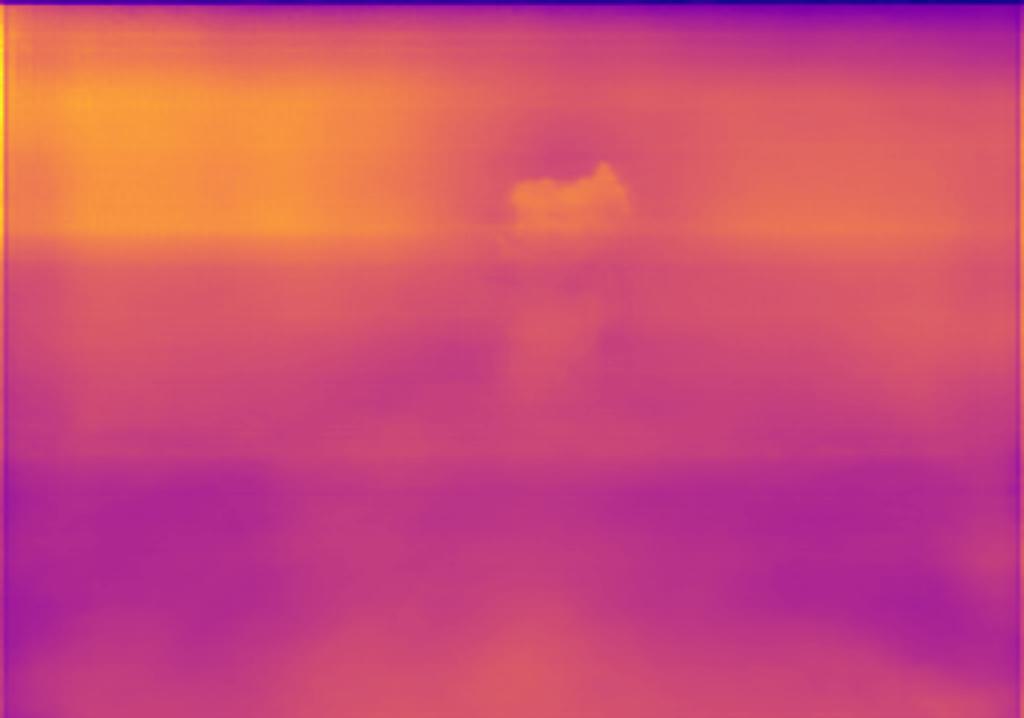} & 
		\includegraphics[width=0.14\linewidth]{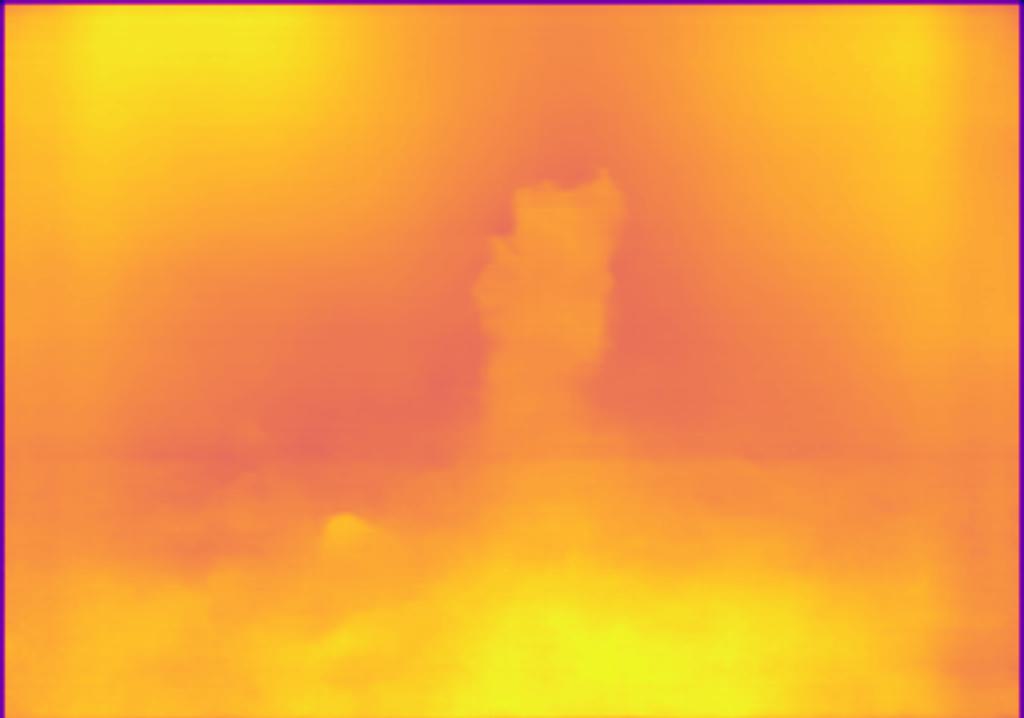} & 
		\includegraphics[width=0.14\linewidth]{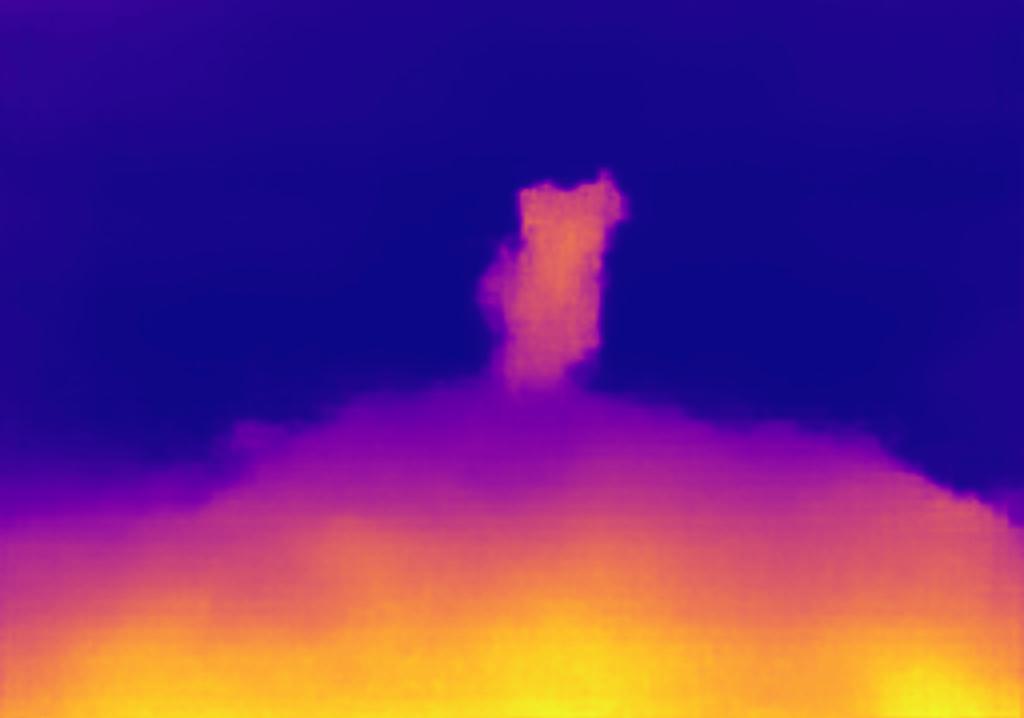}\\	
		
		\includegraphics[width=0.14\linewidth]{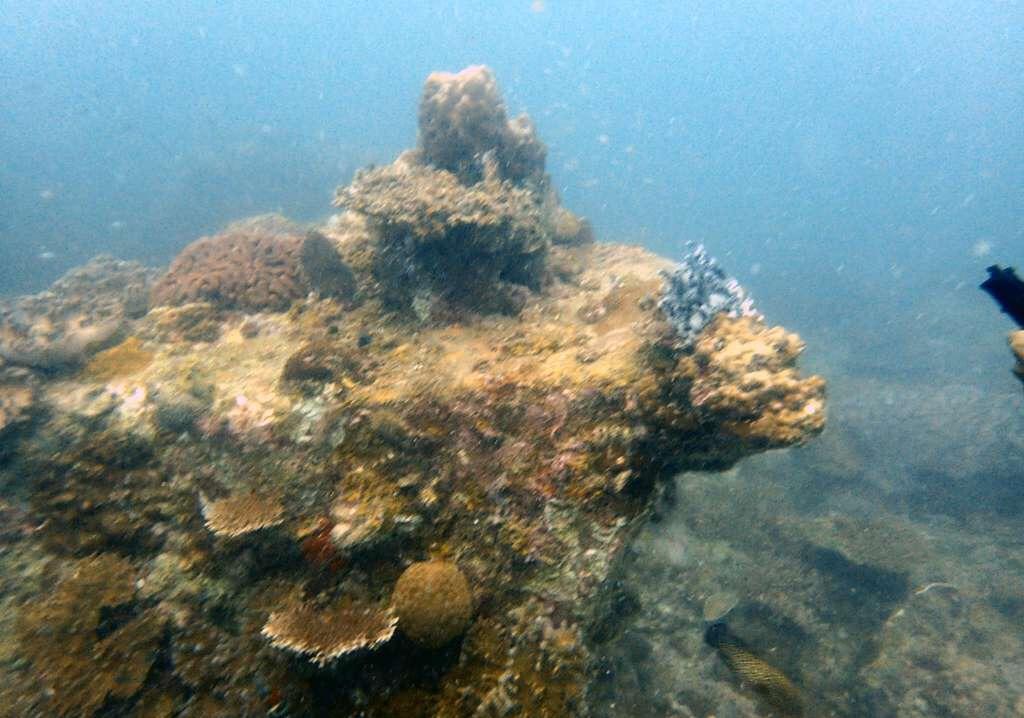} &
		\includegraphics[width=0.14\linewidth]{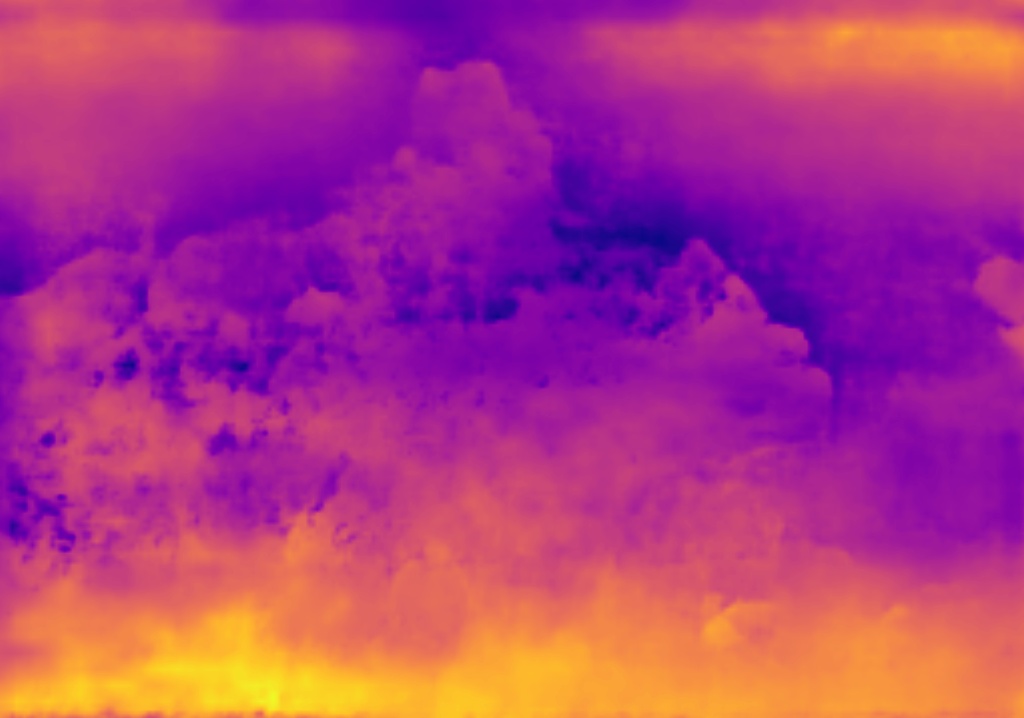} & 
		\includegraphics[width=0.14\linewidth]{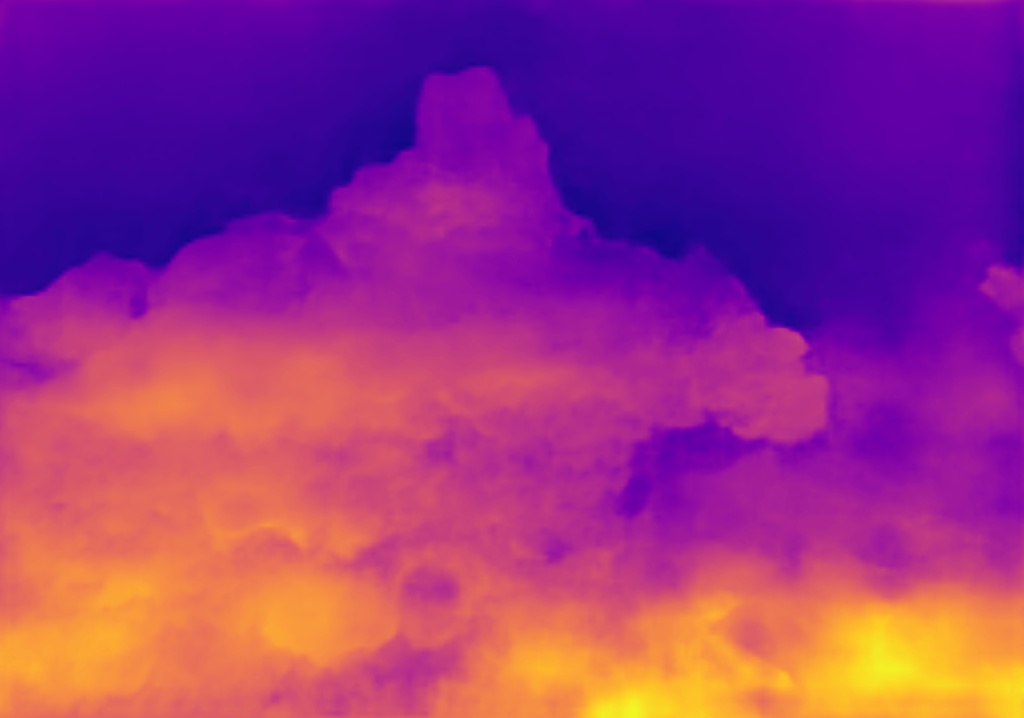} & 
		\includegraphics[width=0.14\linewidth]{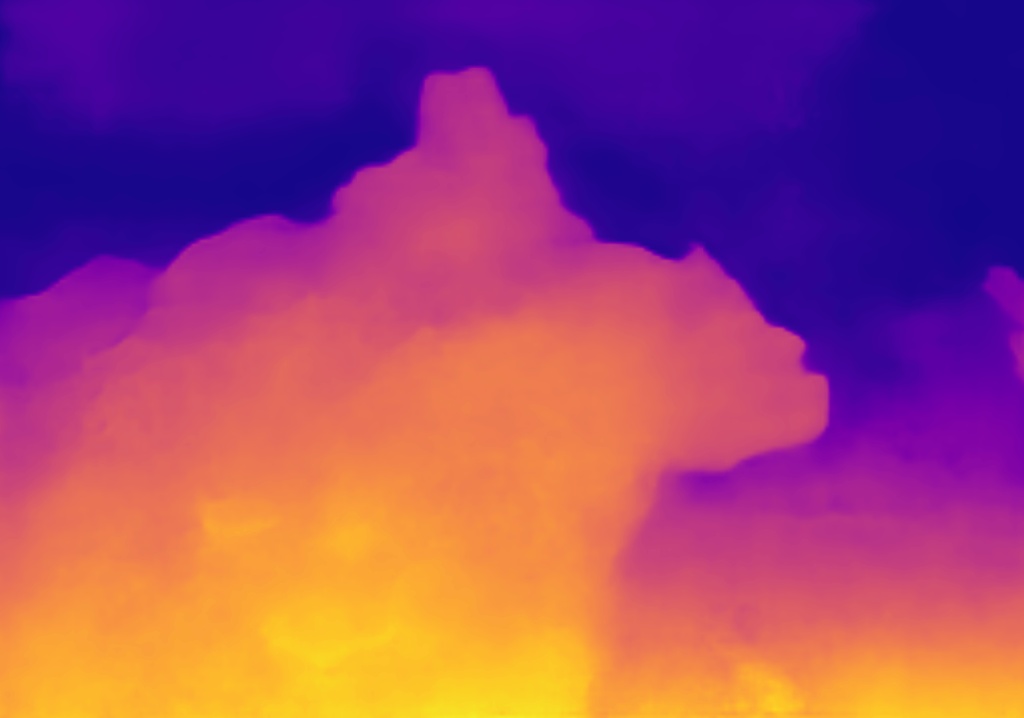} & 
		\includegraphics[width=0.14\linewidth]{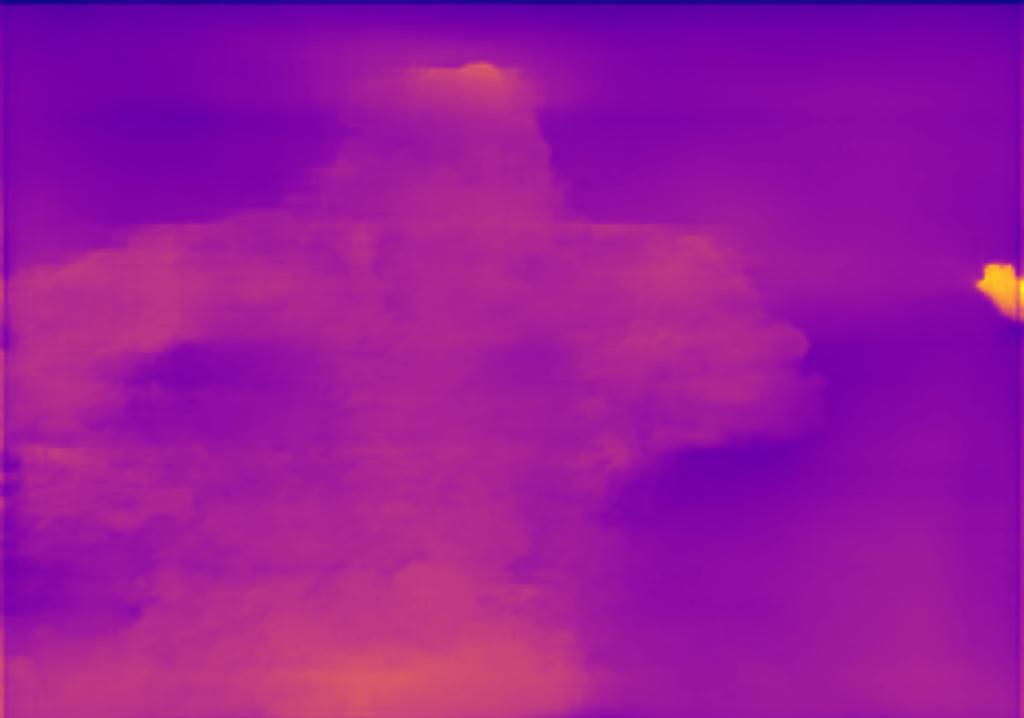} & 
		\includegraphics[width=0.14\linewidth]{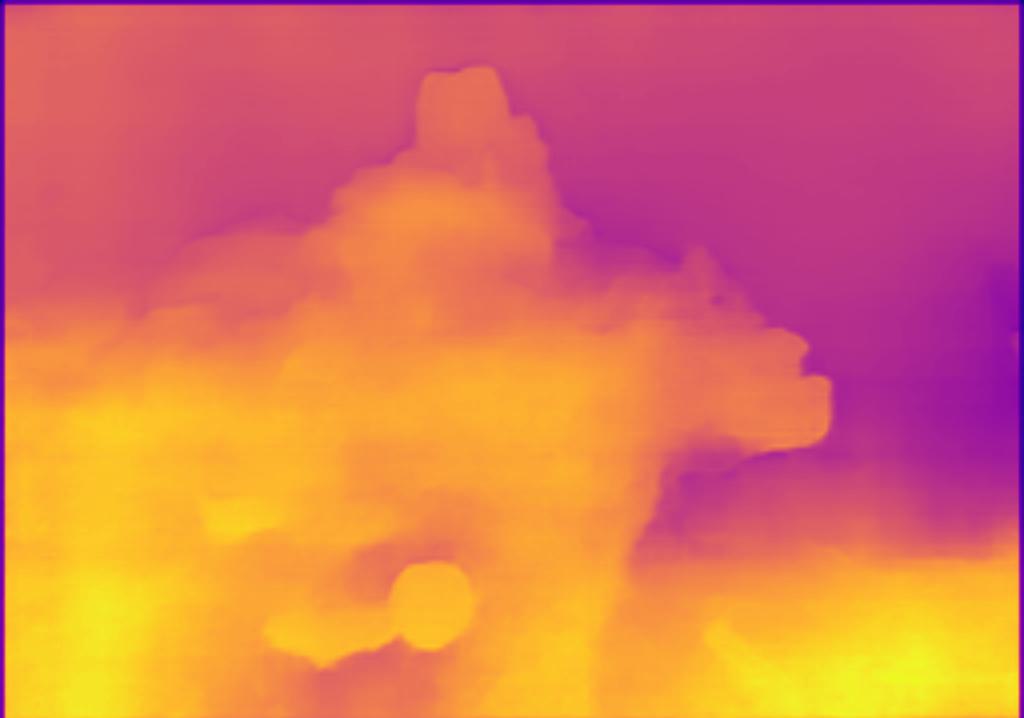} & 
		\includegraphics[width=0.14\linewidth]{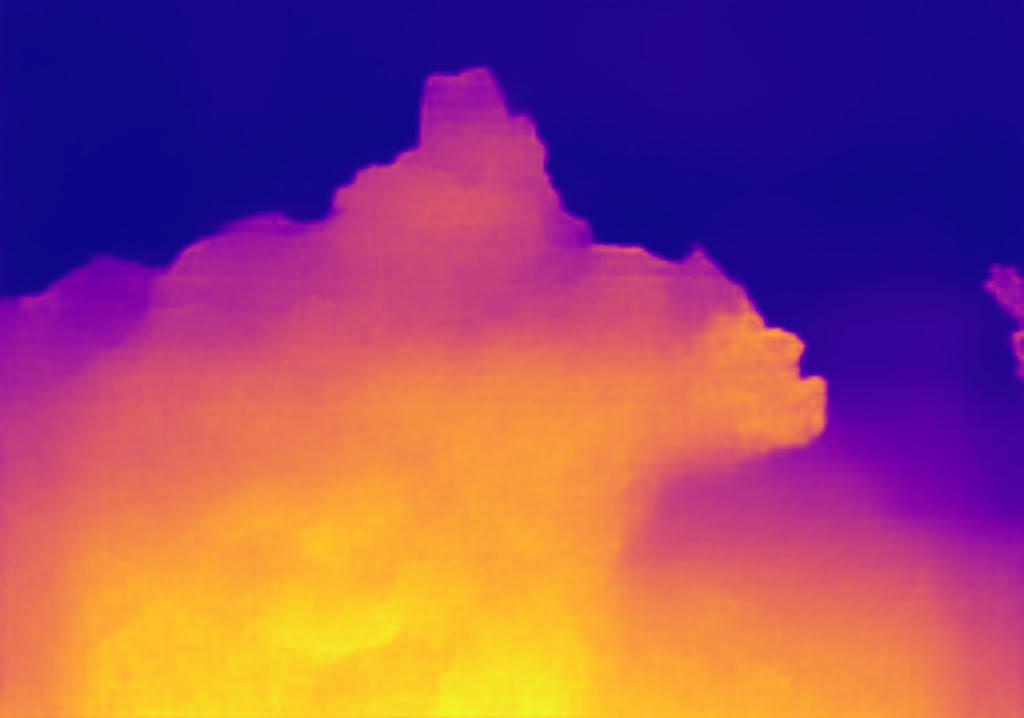}\\
		
		\includegraphics[width=0.14\linewidth]{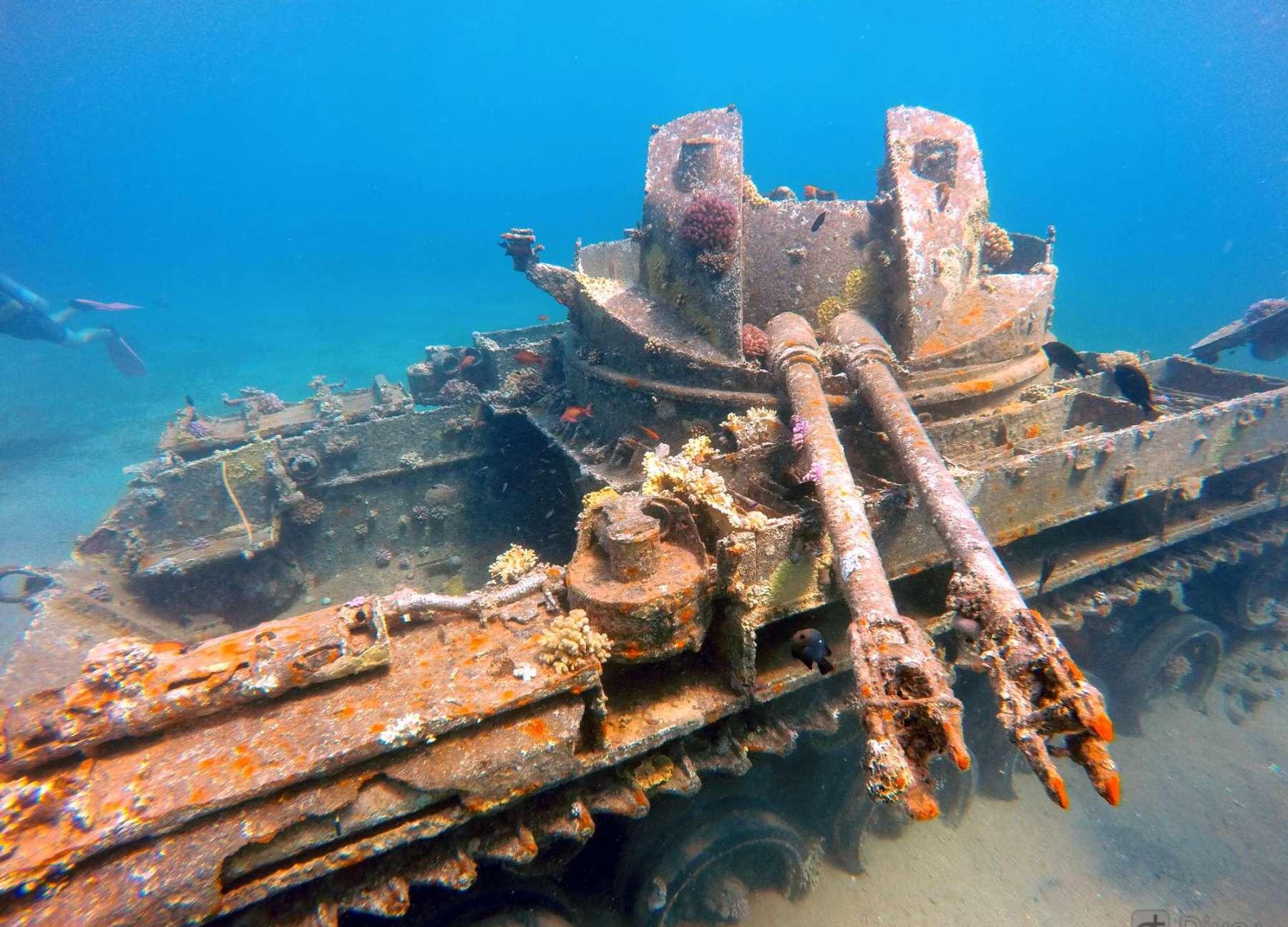} &
		\includegraphics[width=0.14\linewidth]{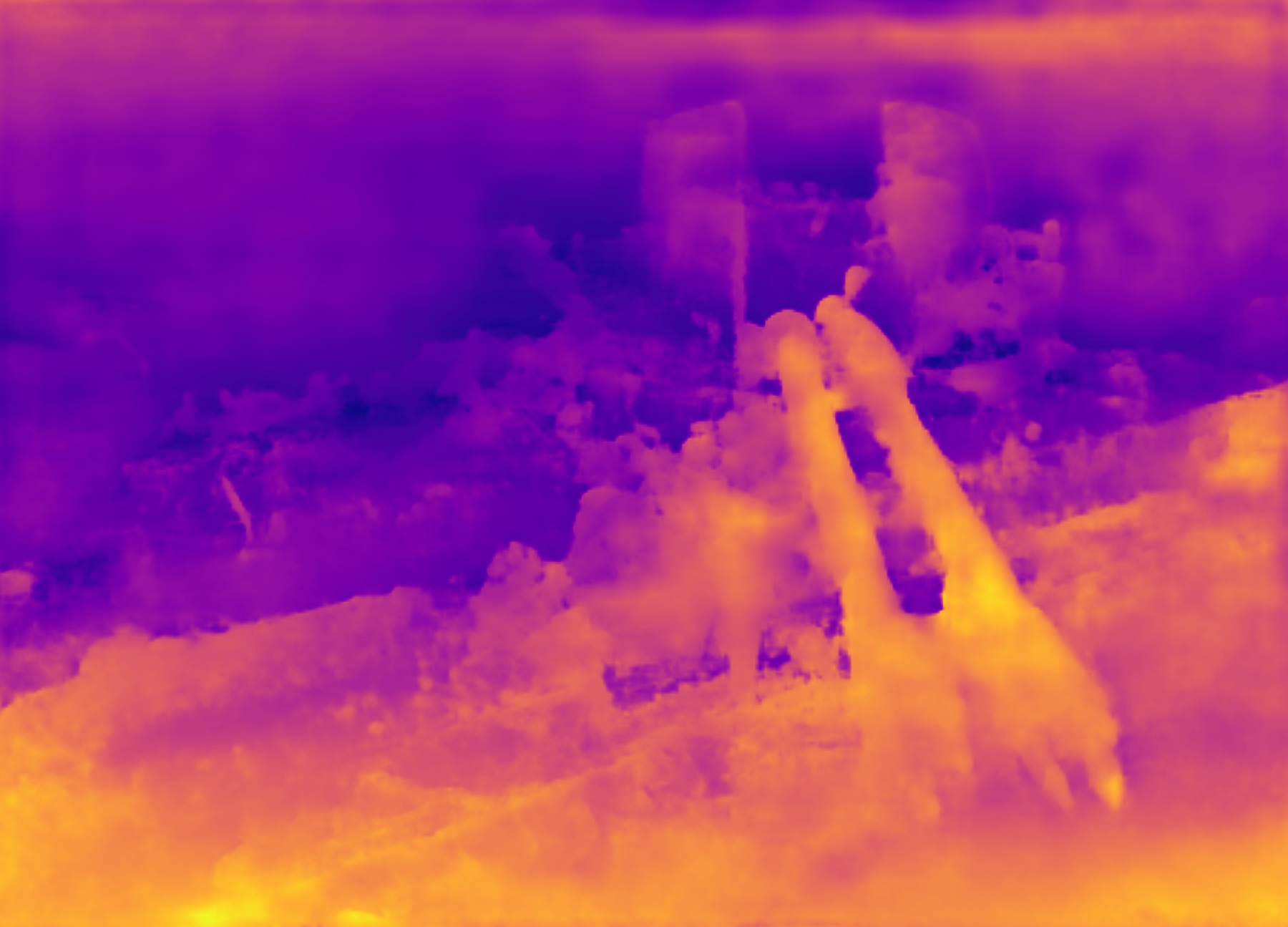} & 
		\includegraphics[width=0.14\linewidth]{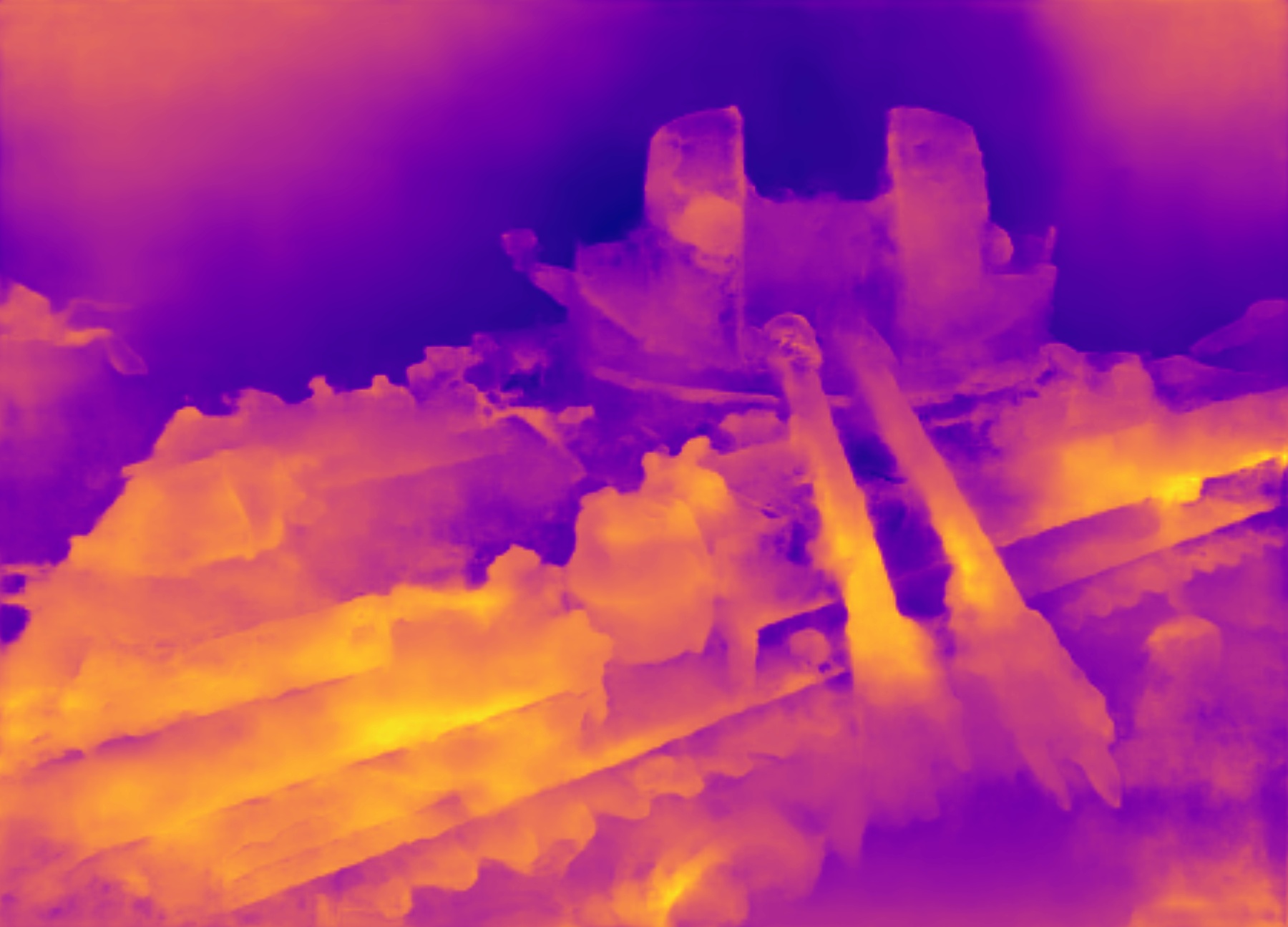} & 
		\includegraphics[width=0.14\linewidth]{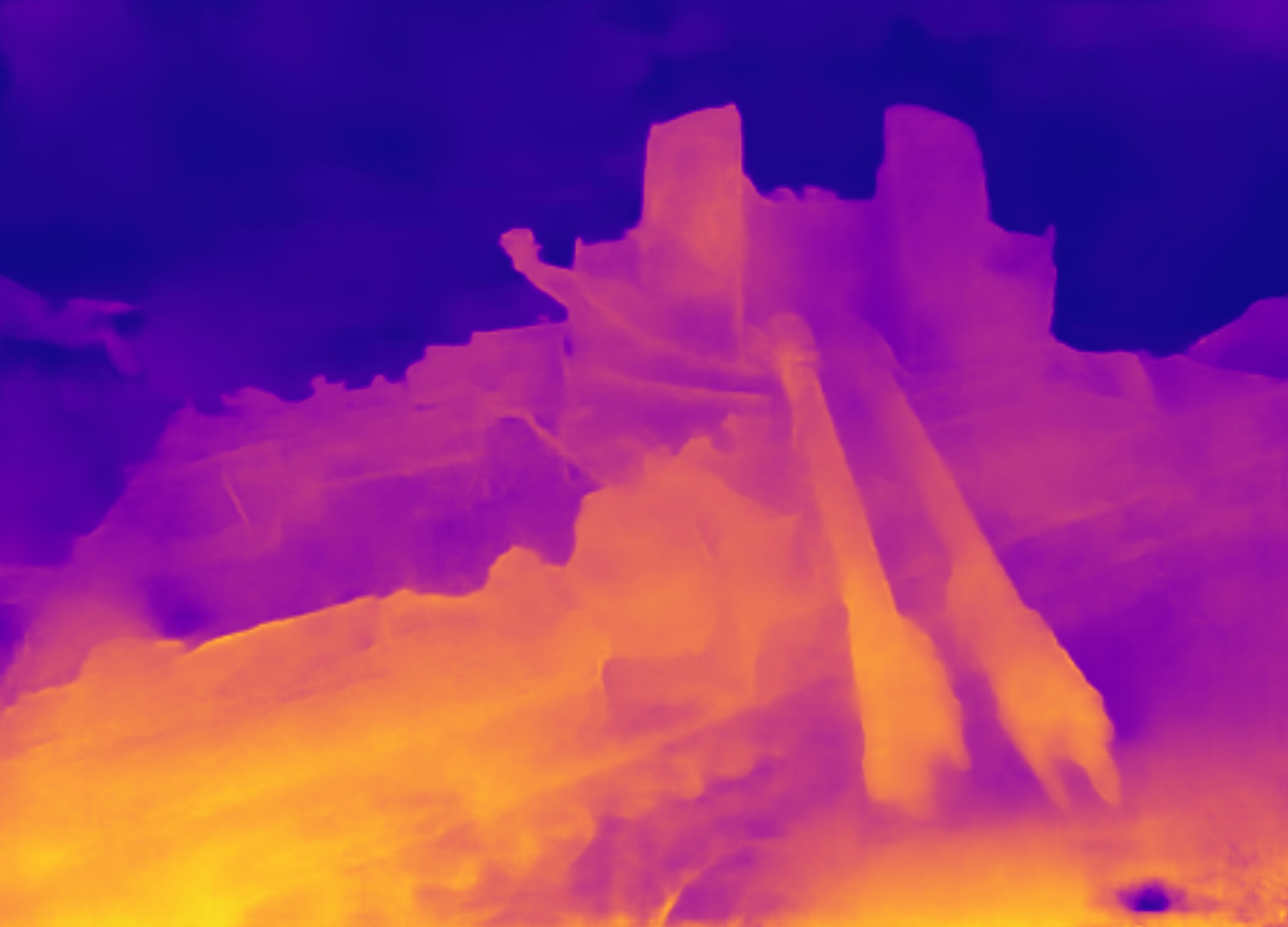} & 
		\includegraphics[width=0.14\linewidth]{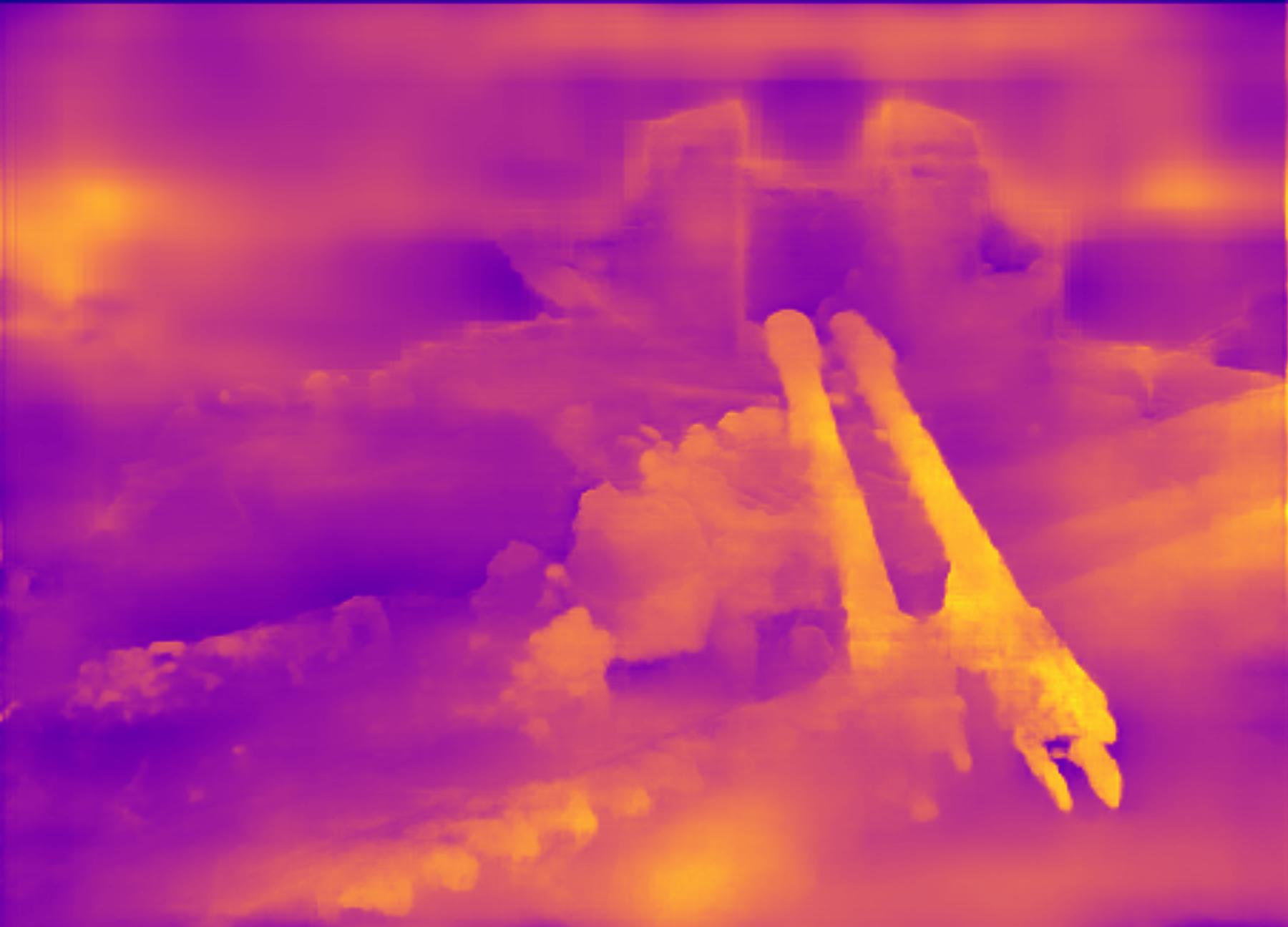} & 
		\includegraphics[width=0.14\linewidth]{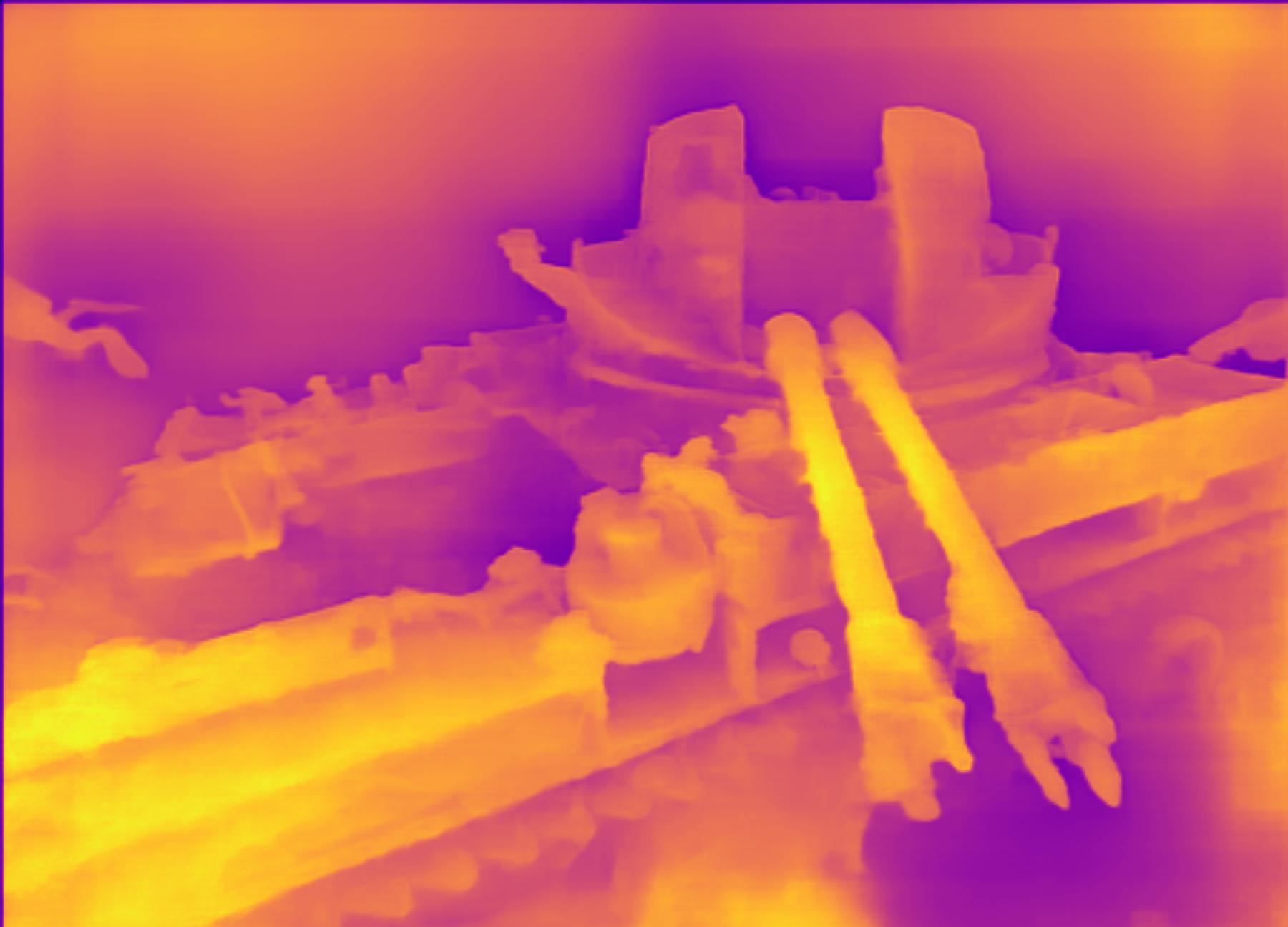} & 
		\includegraphics[width=0.14\linewidth]{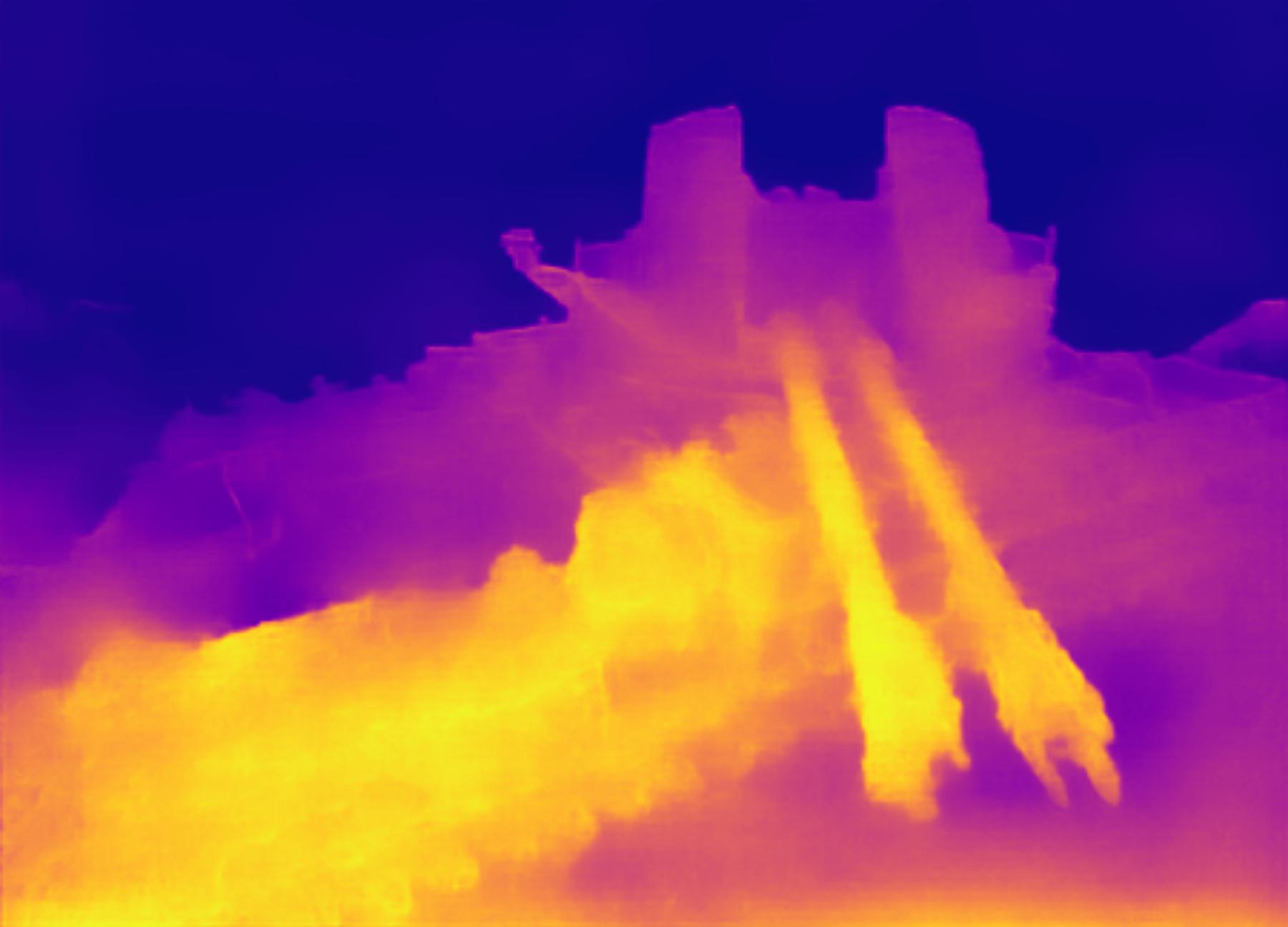}\\
		
		\includegraphics[width=0.14\linewidth]{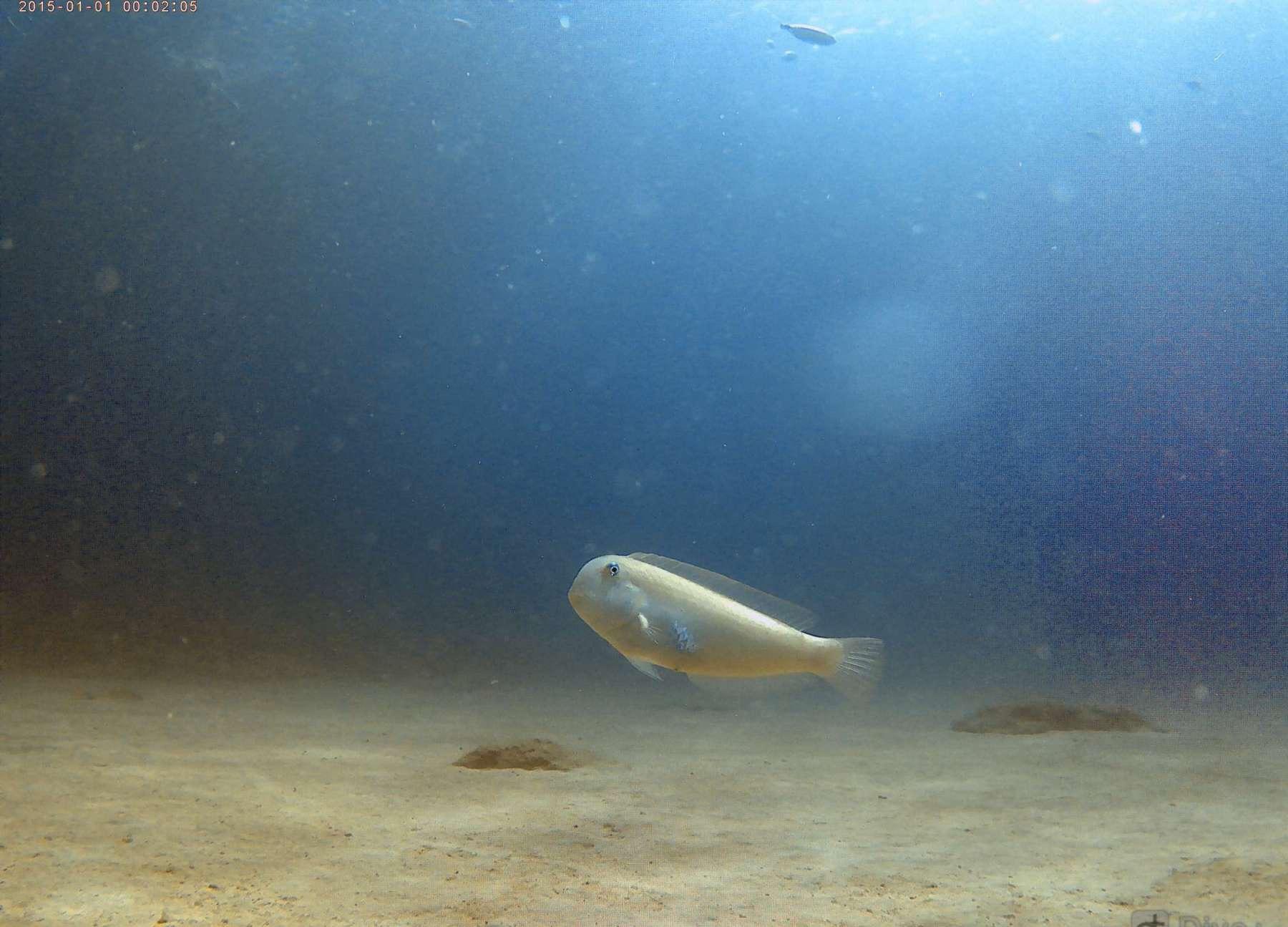} &
		\includegraphics[width=0.14\linewidth]{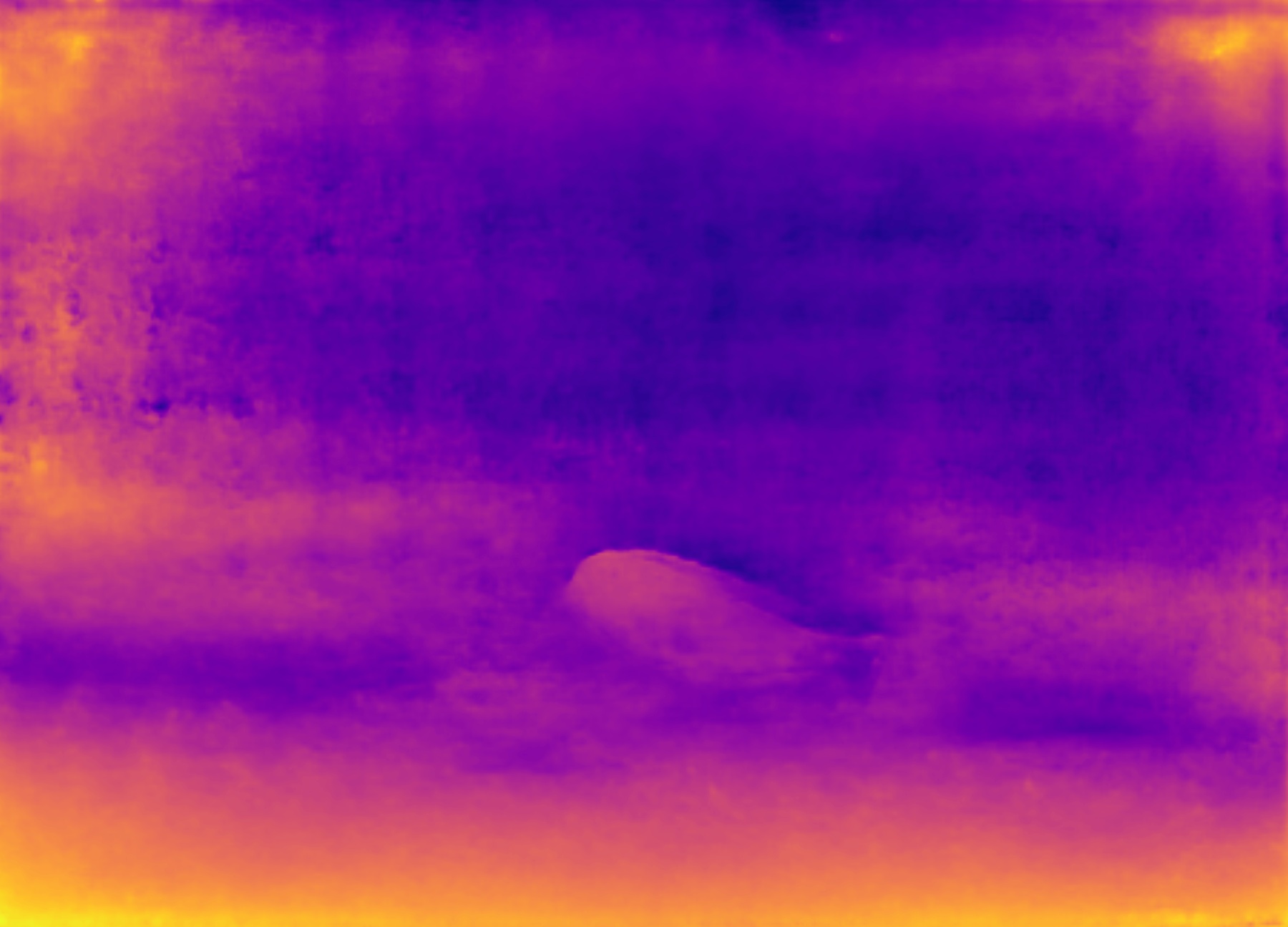} & 
		\includegraphics[width=0.14\linewidth]{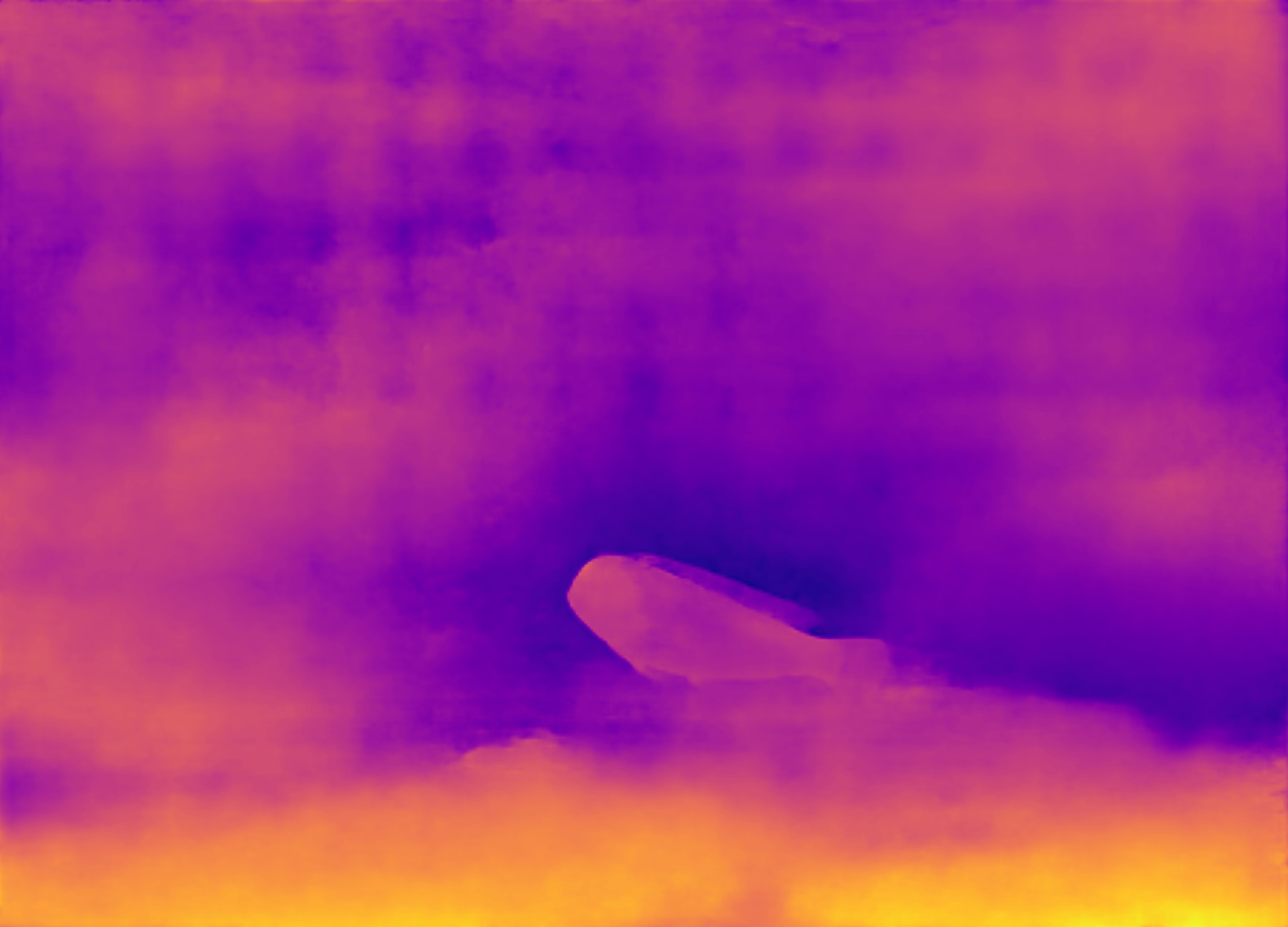} & 
		\includegraphics[width=0.14\linewidth]{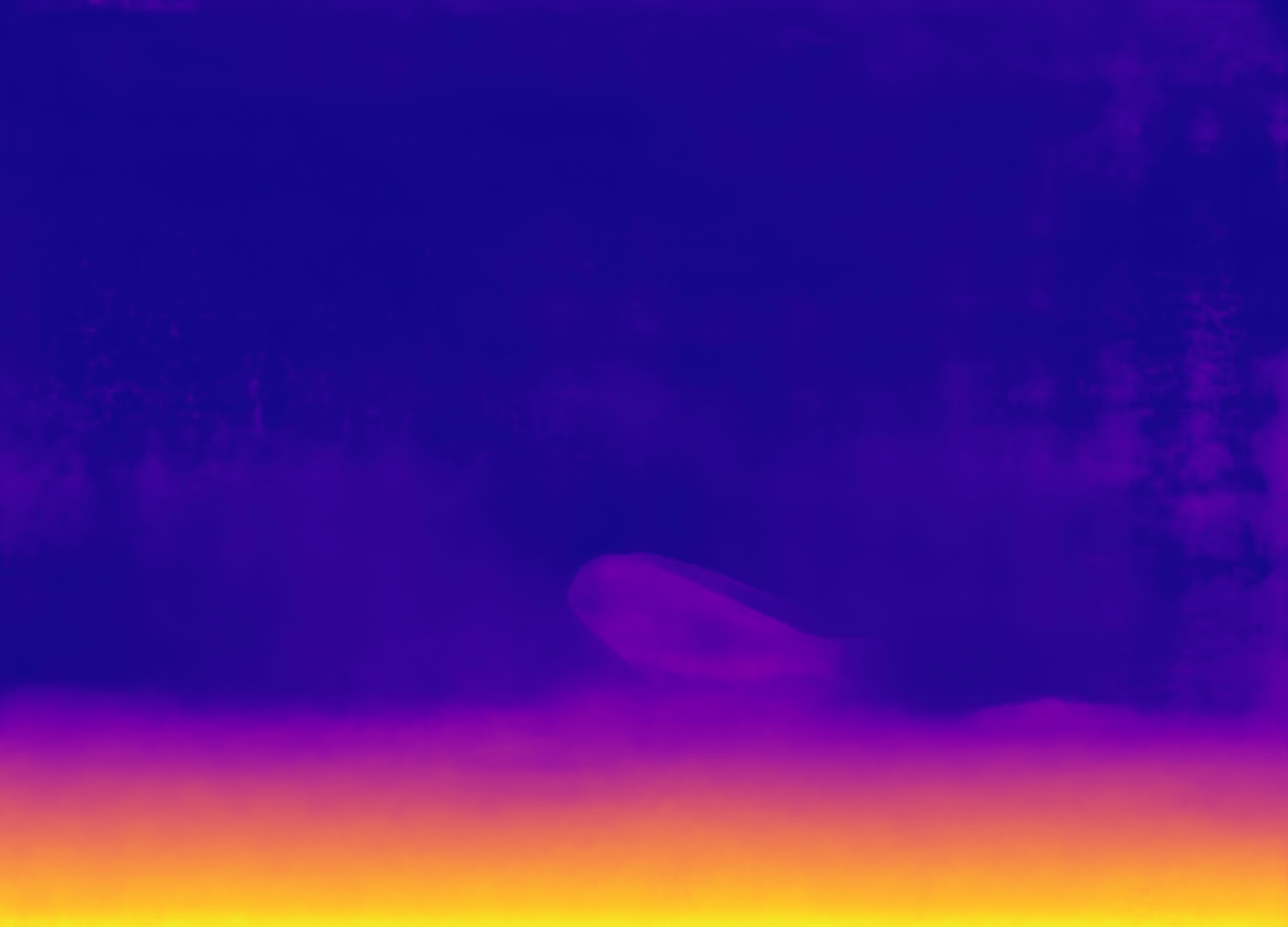} & 
		\includegraphics[width=0.14\linewidth]{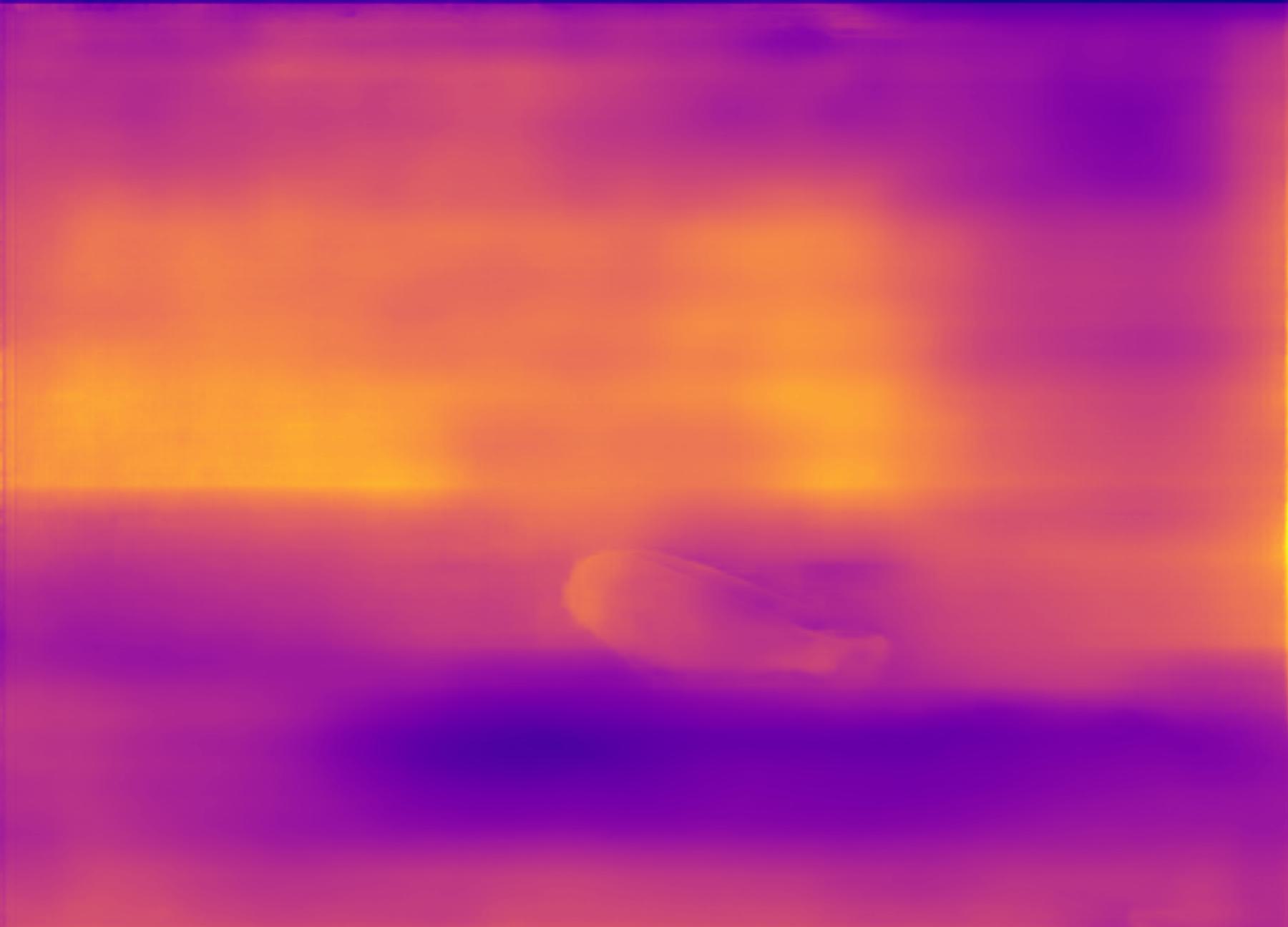} & 
		\includegraphics[width=0.14\linewidth]{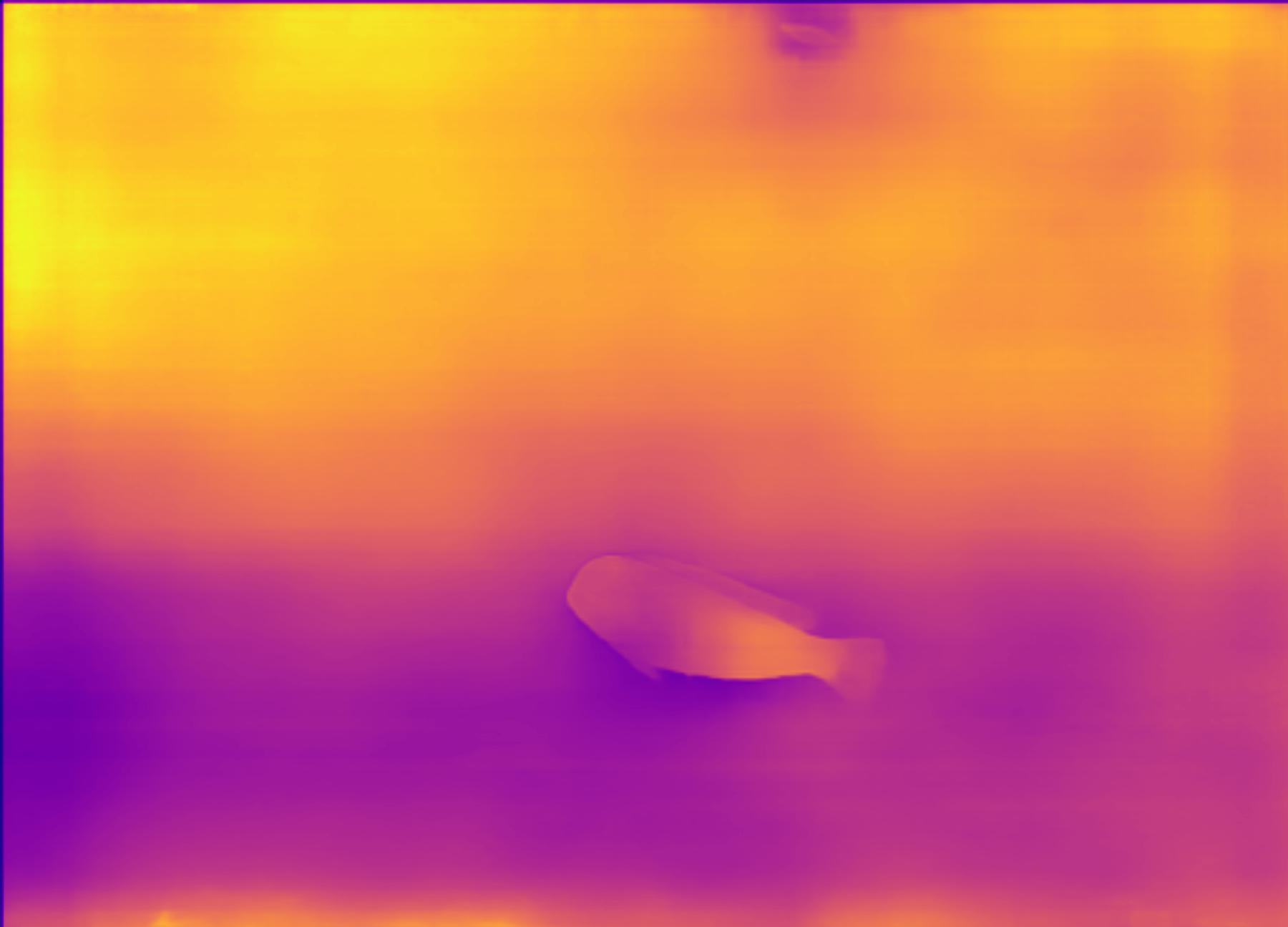} & 
		\includegraphics[width=0.14\linewidth]{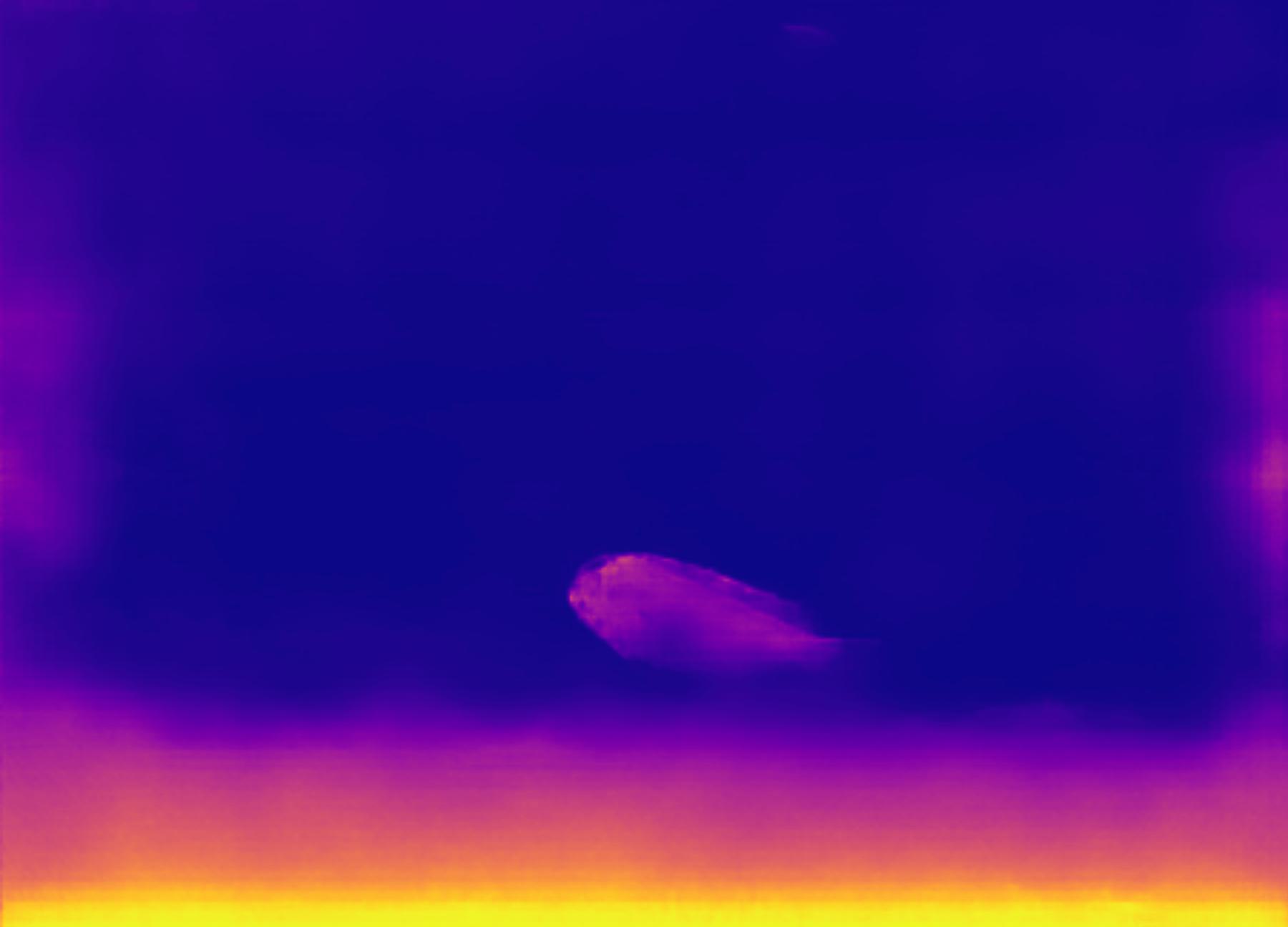}\\
		
		\includegraphics[width=0.14\linewidth]{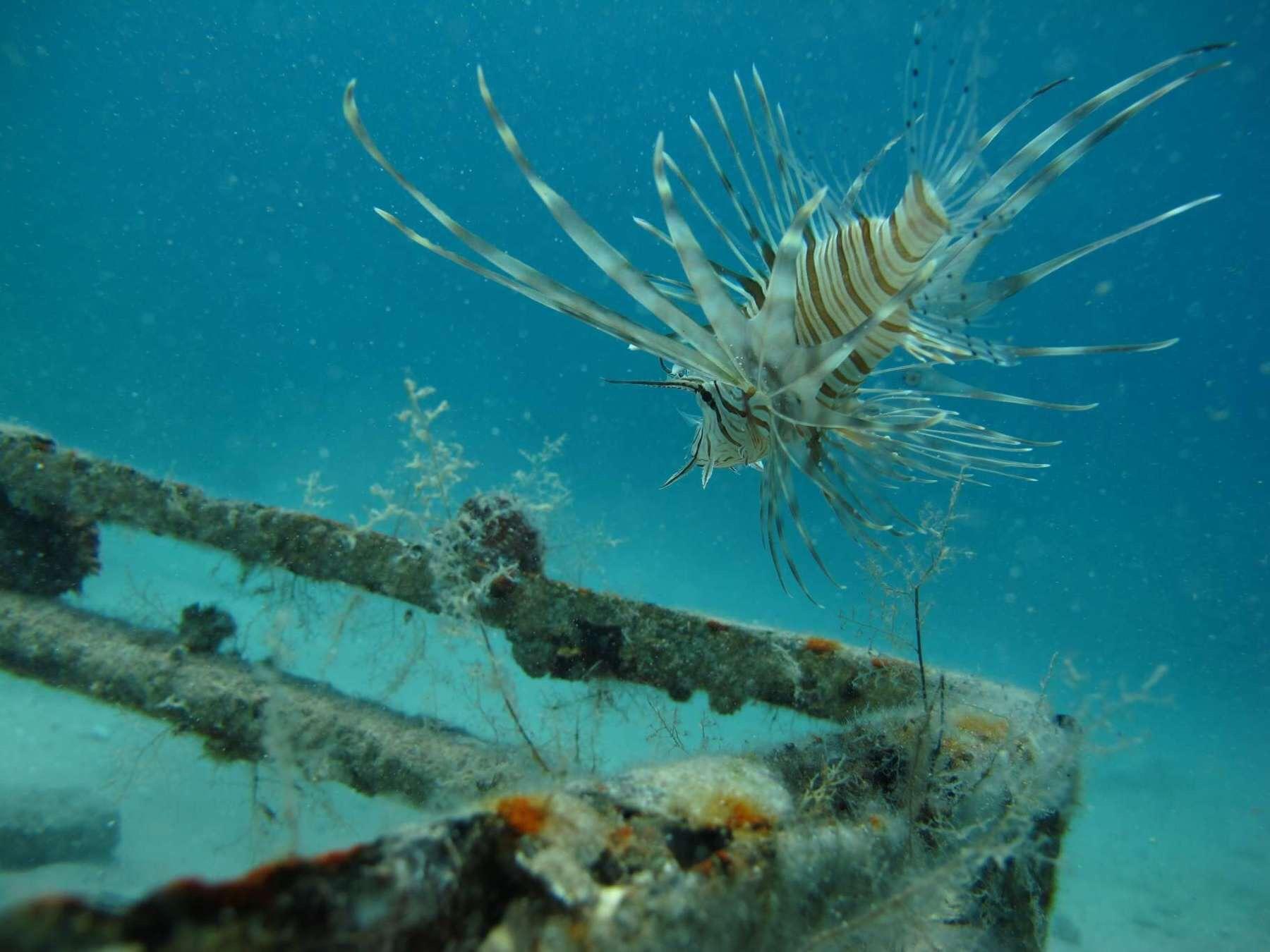} &
		\includegraphics[width=0.14\linewidth]{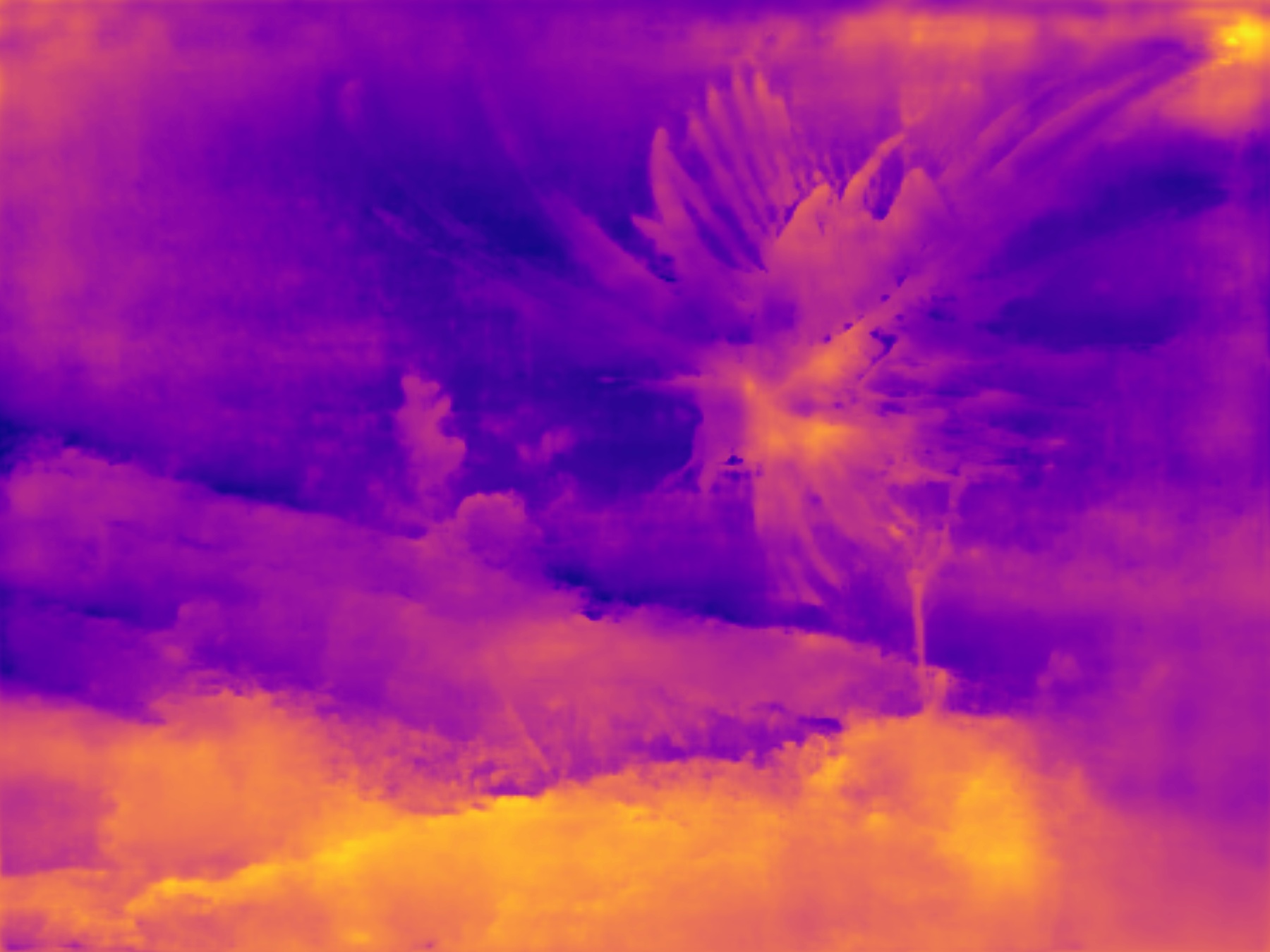} & 
		\includegraphics[width=0.14\linewidth]{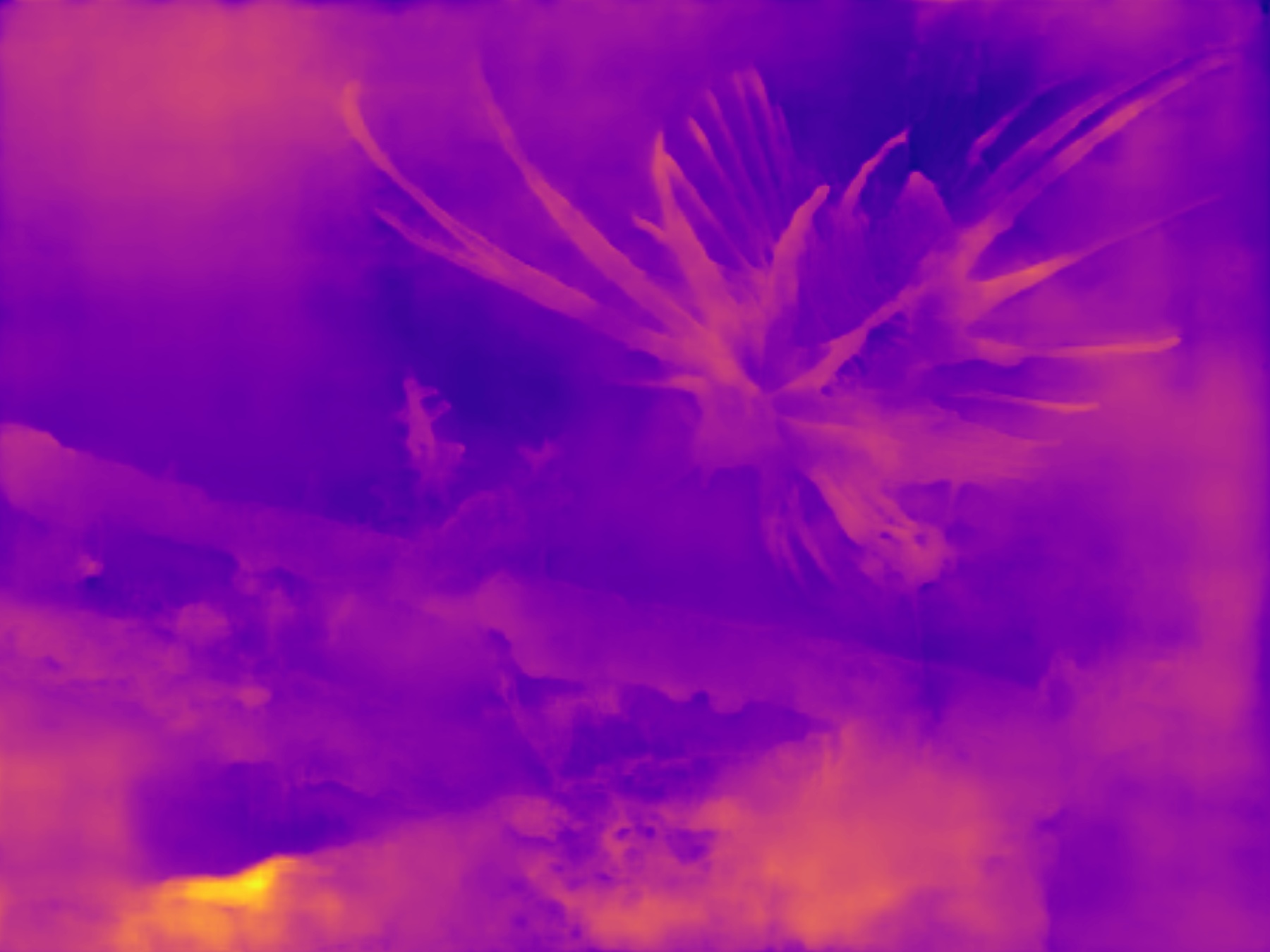} & 
		\includegraphics[width=0.14\linewidth]{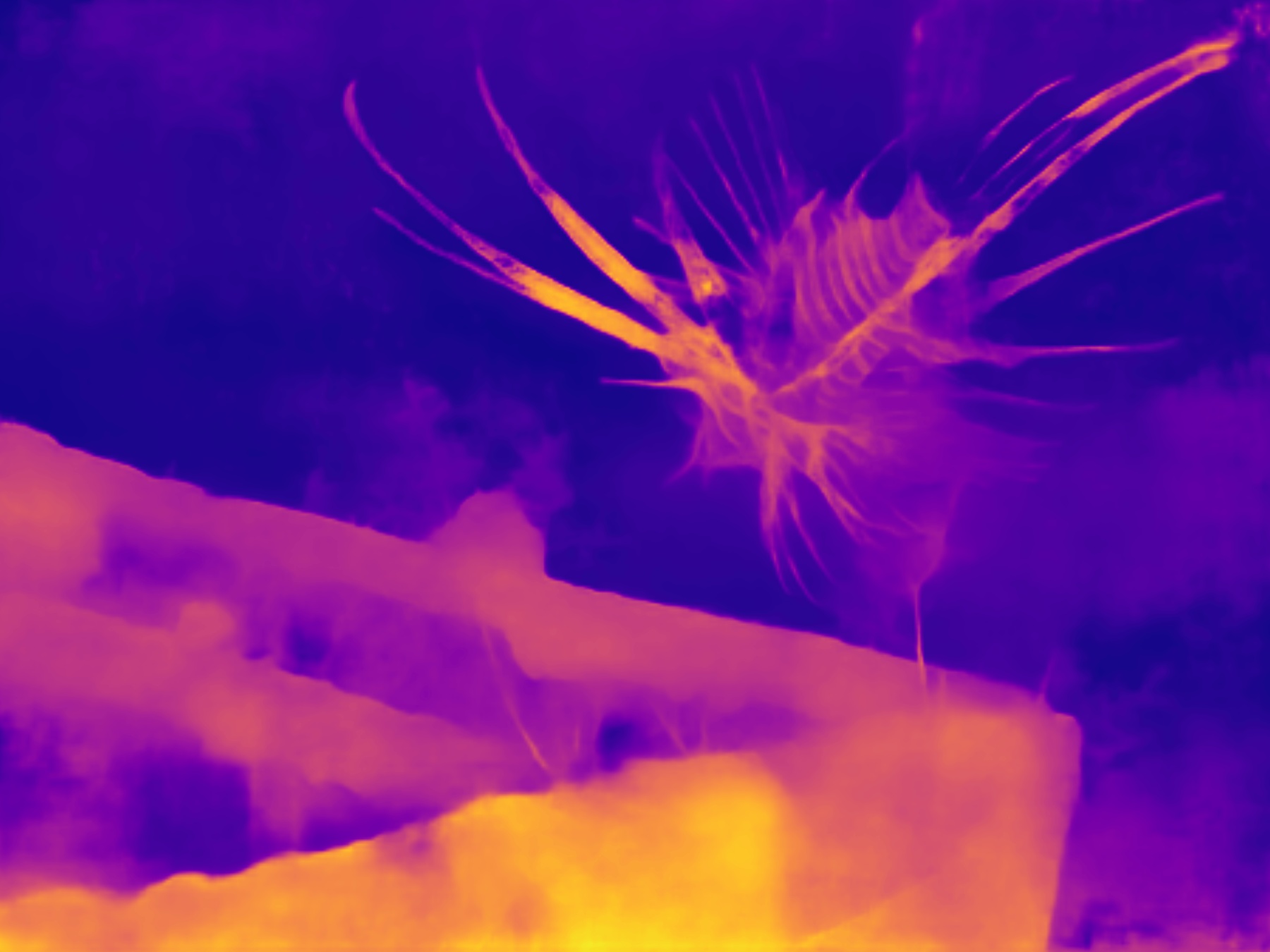} & 
		\includegraphics[width=0.14\linewidth]{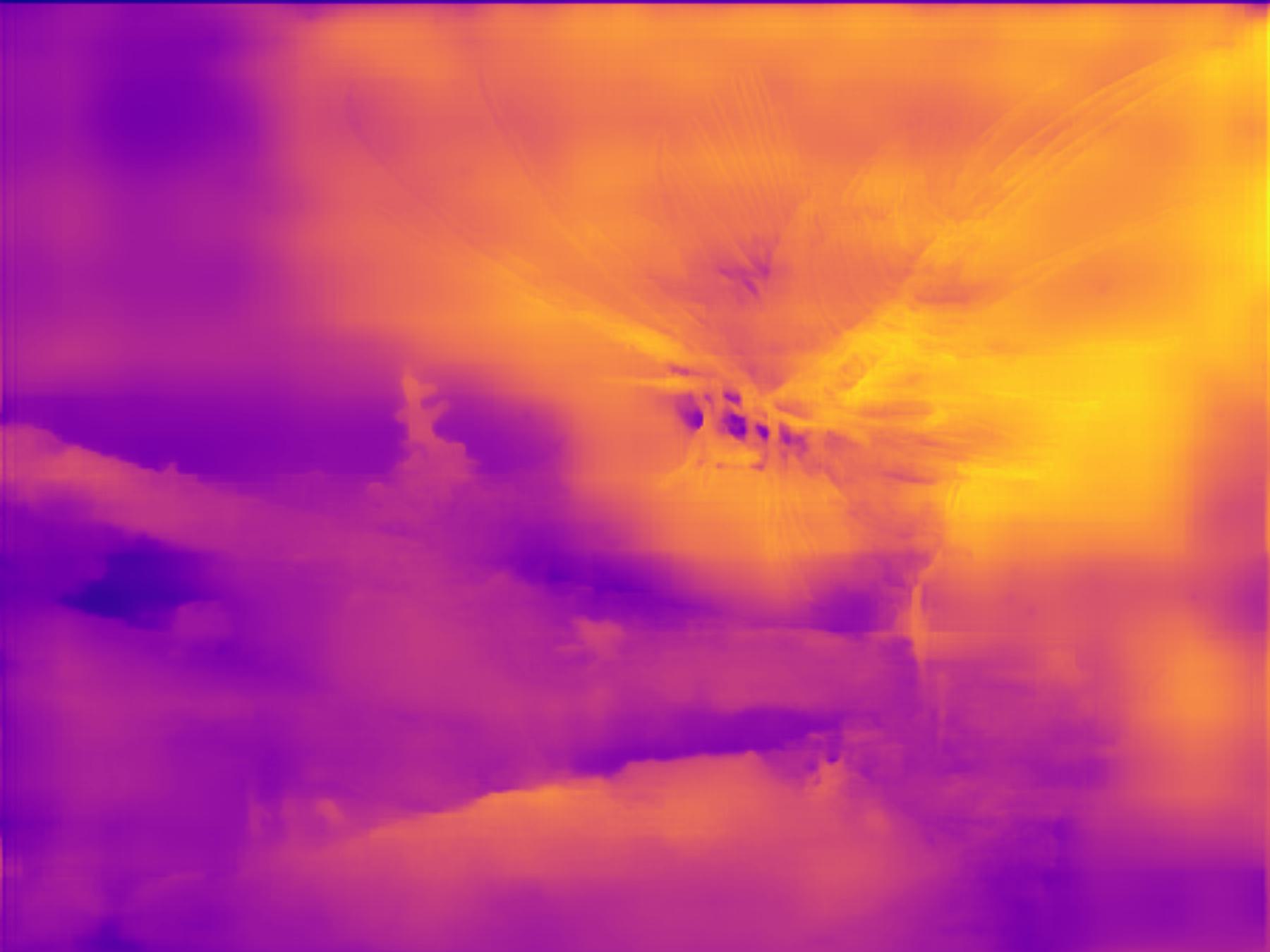} & 
		\includegraphics[width=0.14\linewidth]{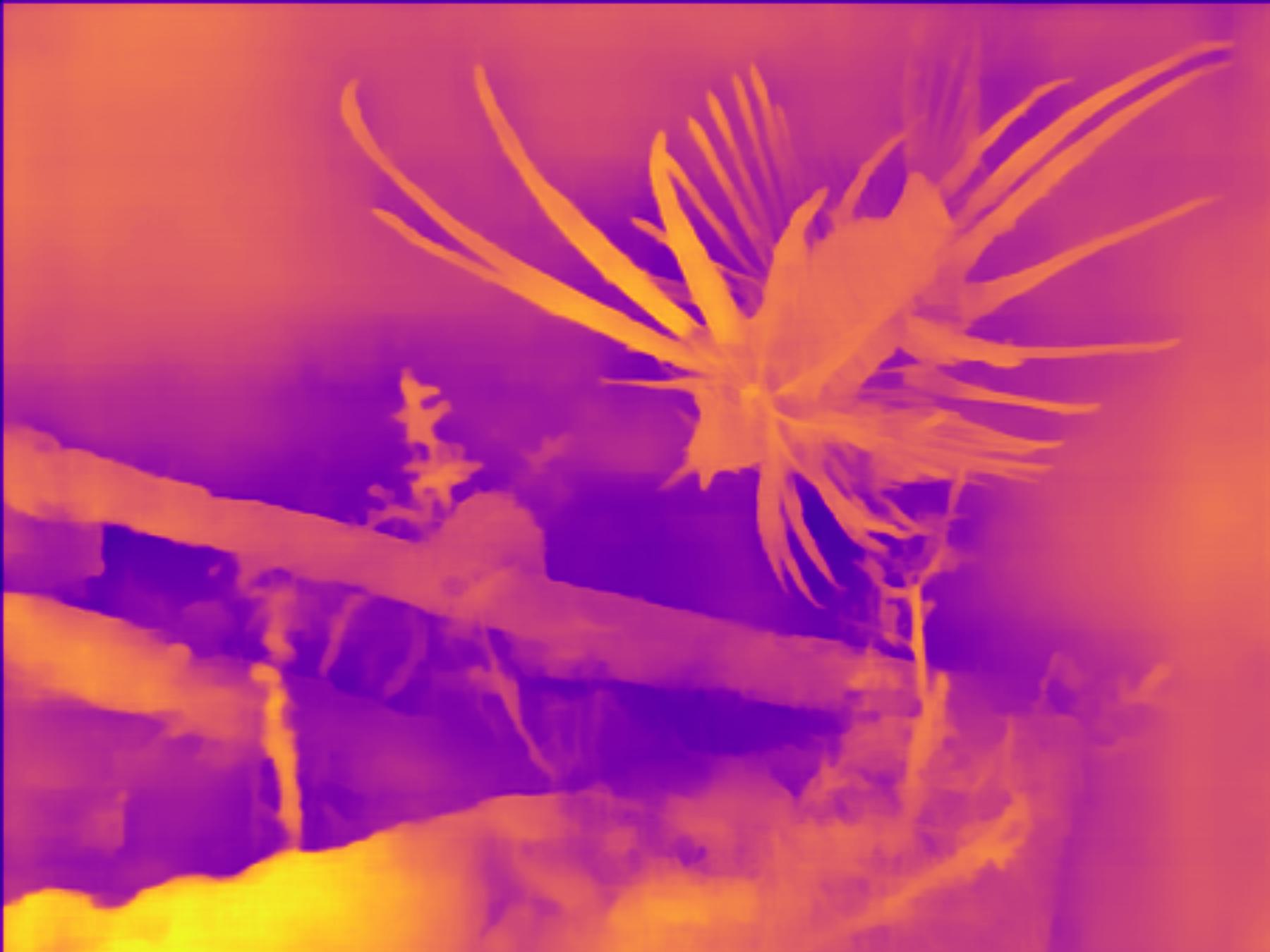} & 
		\includegraphics[width=0.14\linewidth]{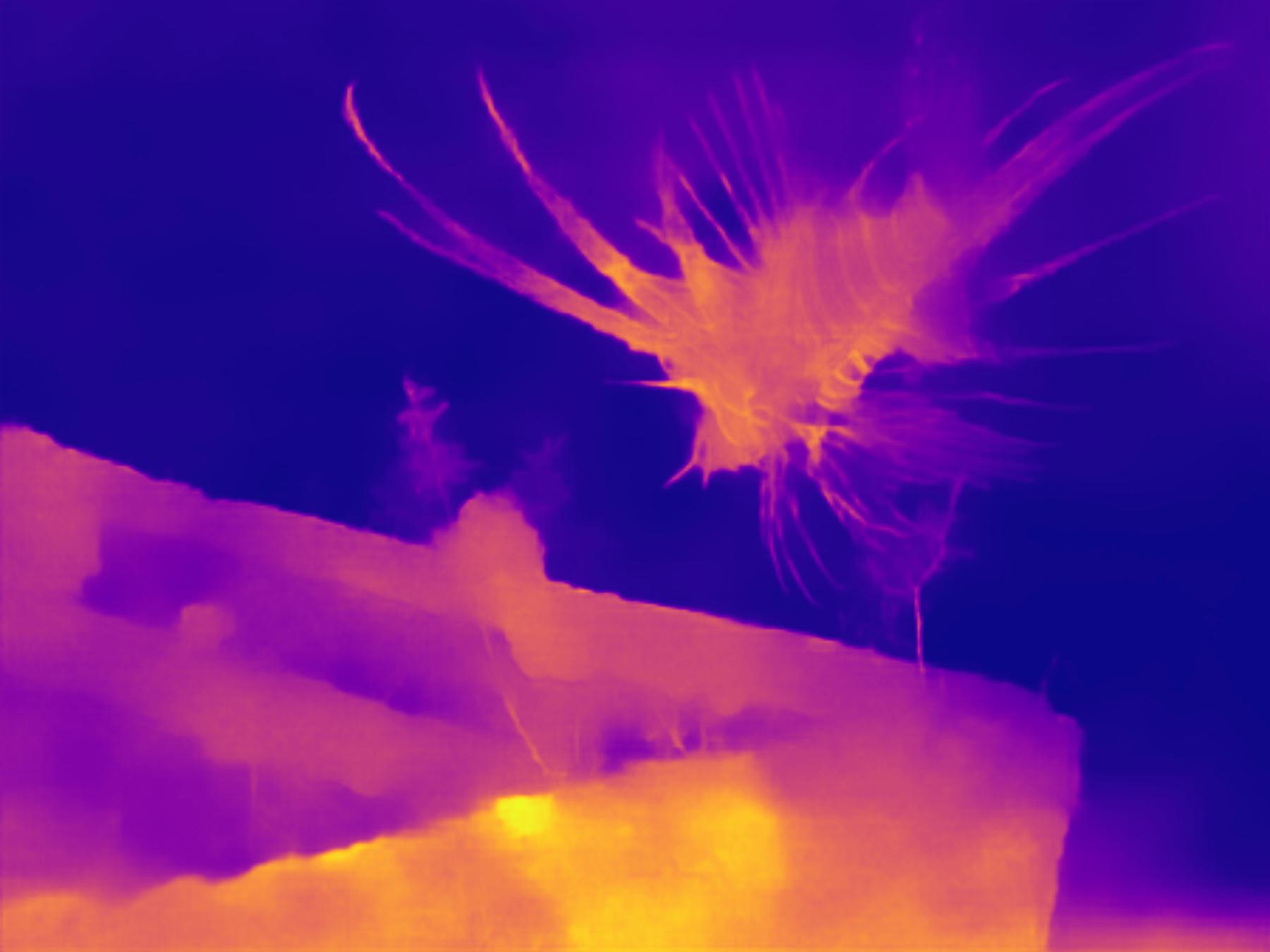}\\
		
		\includegraphics[width=0.14\linewidth]{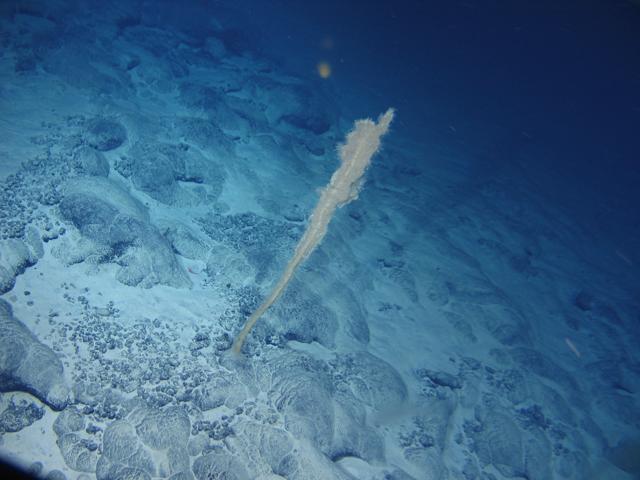} &
		\includegraphics[width=0.14\linewidth]{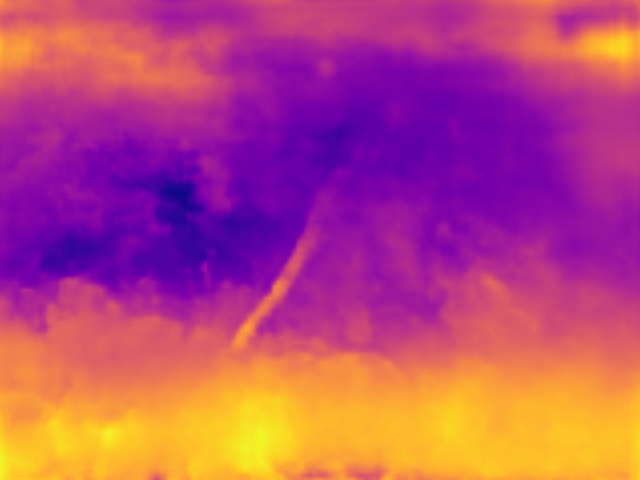} & 
		\includegraphics[width=0.14\linewidth]{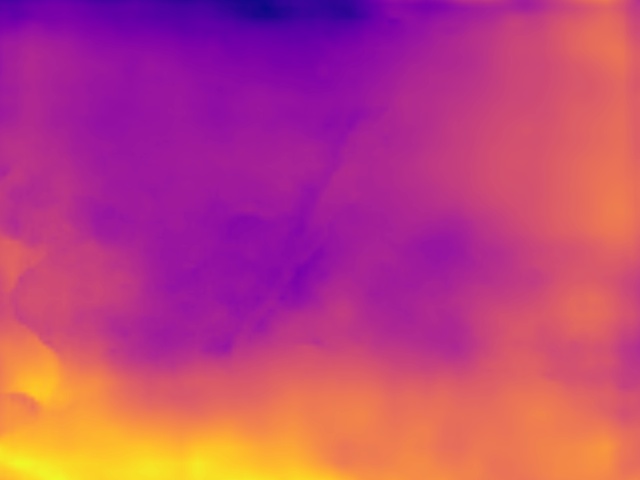} & 
		\includegraphics[width=0.14\linewidth]{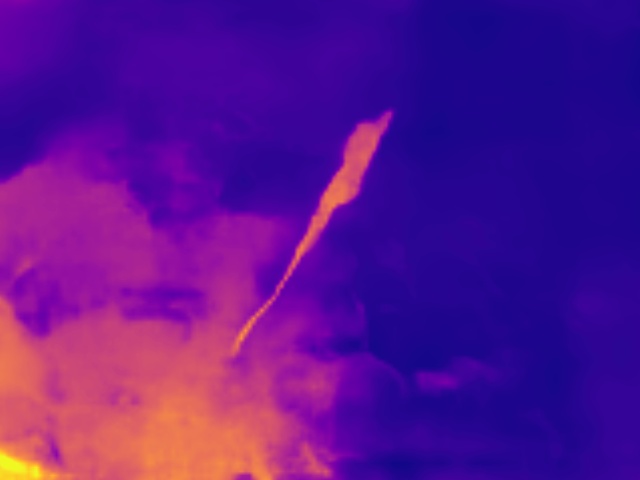} & 
		\includegraphics[width=0.14\linewidth]{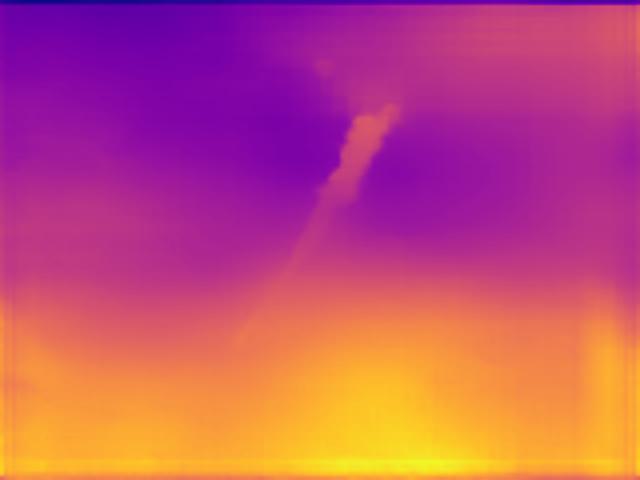} & 
		\includegraphics[width=0.14\linewidth]{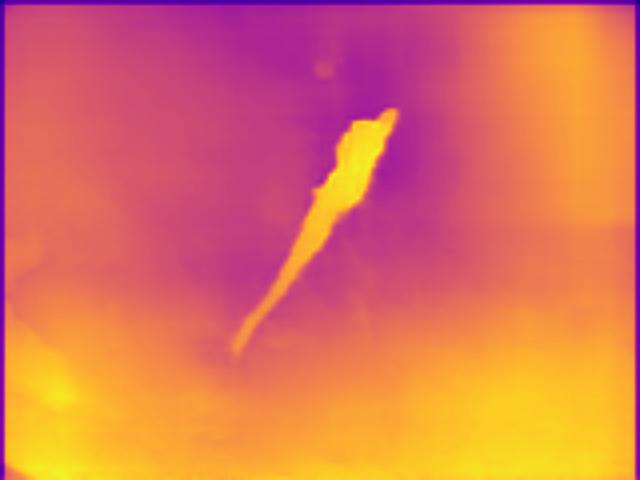} & 
		\includegraphics[width=0.14\linewidth]{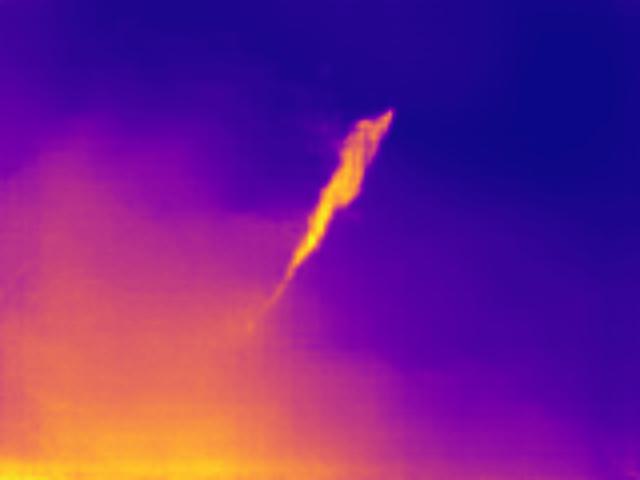}\\
		Input & iDisc (K)  & iDisc (N)  & iDisc (A)   & NeWCRFs (K)  & NeWCRFs (N)  & NeWCRFs (A) \\
		
	\end{tabular}
	\vspace{-3mm}
	\caption{Qualitative results on test set of UIEB dataset \cite{uieb2019li}. K and N denote models pretrained on KITTI \cite{kitti} and NYU Depthv2 \cite{nyudepthv2} datasets. A represents the models trained on our dataset Atlantis. Our method gets the best visual results. Please zoom in for details.} 
	\label{fig:uieb}
	\vspace{-5mm}
\end{figure*}
\subsection{Depth Uncertainty}
Despite the excellent performance and robustness of MiDaS \cite{midas} model in general depth estimation tasks, handling underwater scenes invariably introduces a domain gap. This gap stems from the distinct and challenging visual characteristics of underwater environments, which are not typically represented in the model's training data. It can lead to inaccuracies in depth maps estimated for out-of-distribution underwater images. To address this challenge, we propose the Depth Uncertainty (DU) metric as a means to quantify and mitigate the inaccuracies arising from this domain gap.
\vspace{-6mm}

\paragraph{Depth Uncertainty (DU).} The DU metric measures the variance in depth estimations produced by the MiDaS model for both the original underwater images and their horizontally flipped counterparts. This variance reflects the model's consistency under varied input conditions, as depth models often exhibit inconsistent results for flipped images, a phenomenon leveraged in self-supervised training methods like Monodepth \cite{monodepth}. This is a crucial factor considering depth models like MiDaS are typically less exposed to underwater imaging conditions. For each original underwater image $\bar{U}$ and its flipped version $\bar{U}^{lr}$, the DU for each pixel location $p$ is calculated as follows:
\begin{equation}
	DU_p = Var(D_p, D_p^{lr}),
\end{equation}
where $DU_p$ represents the per-pixel variance between the depth maps $\bar{D}$ and $\bar{D}^{lr}$. This variance provides a quantitative measure of the depth estimation reliability \cite{uncertainty} in the face of the domain gap.

\vspace{-4mm}
\paragraph{Validity Mask.} To ensure the reliability of our depth data, we introduce a Validity Mask, filtering out depth values at pixel locations where the DU exceeds a certain threshold. This threshold is empirically set at 0.15, allowing us to identify and discard depth values with high uncertainty. Consequently, only depth values with $DU<0.15$ are considered reliable, enhancing the overall quality and dependability of the depth information used in our dataset.

\subsection{Implementation Details}
This subsection outlines the key implementation aspects of our data generation pipeline, ensuring a comprehensive understanding of the process and techniques involved.

\noindent\textbf{Data Preparation.} We leveraged the training set of UIEB dataset \cite{uieb2019li}, which consists of 700 underwater images, for initial depth estimation and captioning. The robust MiDaS model \cite{midas} was employed for depth estimation, while the BLIP2 model \cite{blip2} facilitated image captioning. These steps resulted in 700 data triplets comprising underwater images, depth maps, and textual descriptions, forming the foundation of our training data for ControlNet.

\noindent\textbf{ControlNet Training and Deployment.} We utilized the diffusers library \cite{diffusers} for the modification and efficient deployment of both SD and ControlNet. We train the ControlNet using standard training settings. For inference, we set the guidance scale to 5, avoiding unrealistic lighting styles, and sample for 20 steps for each image generation.

\noindent\textbf{Depth Estimation Model Training.} For the training of depth estimation models, we employed recent iDisc \cite{idisc2023piccinelli} and NeWCRFs \cite{newcrfs2022yuan}. These models were trained on our generated underwater depth dataset. Given that MiDaS outputs inverse relative depth, we capped the depth values at a maximum of 20 meters. This aligns with the understanding that scene radiance in underwater environments is predominantly affected by backscattering beyond this depth \cite{seathur2019akkaynak}.

\noindent\textbf{Hardware and Accessibility.} All experiments and model trainings were conducted on an NVIDIA RTX 3090 GPU. We plan to make both the intermediate triplet data and underwater depth dataset, as well as the customized \textit{Depth2Underwater} ControlNet publicly available, contributing to the broader research community in this field.

%% file: sec/3_experiment.tex
\begin{figure*}[t]\small
	\centering
	\setlength{\tabcolsep}{1pt}
	\begin{tabular}{cccc}
		\includegraphics[width=0.21\linewidth]{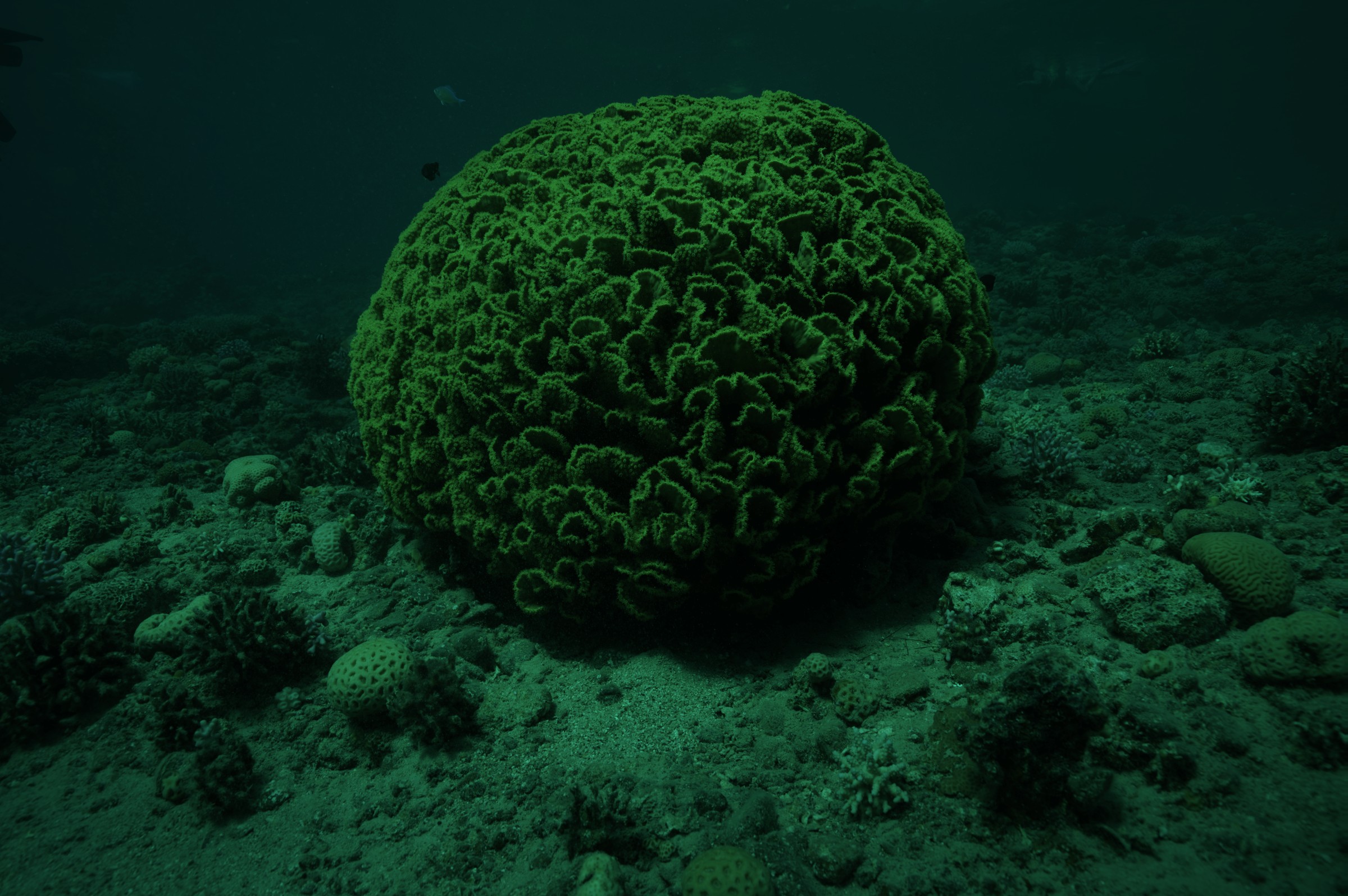}&
		\includegraphics[width=0.21\linewidth]{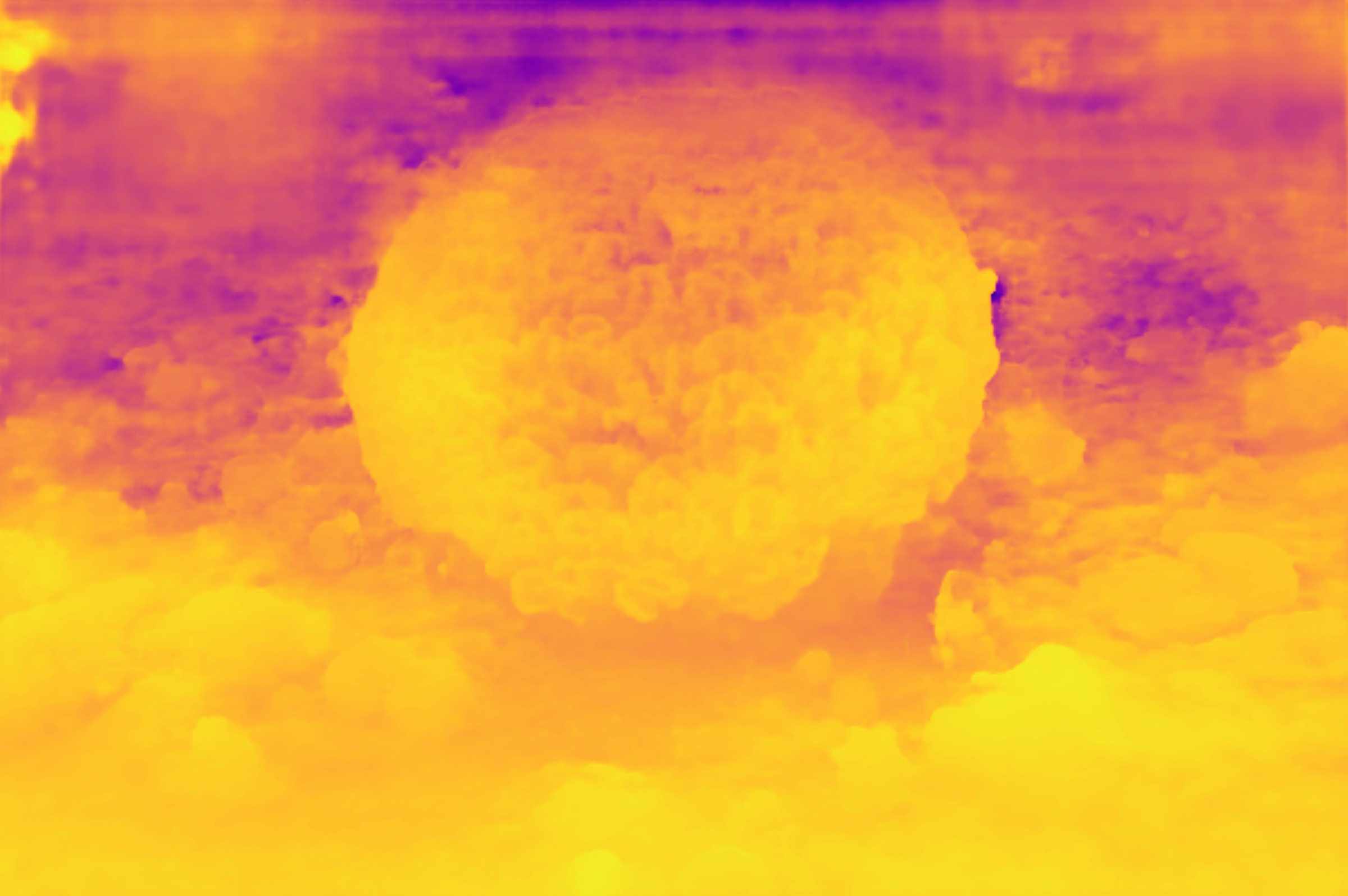} & 
		\includegraphics[width=0.21\linewidth]{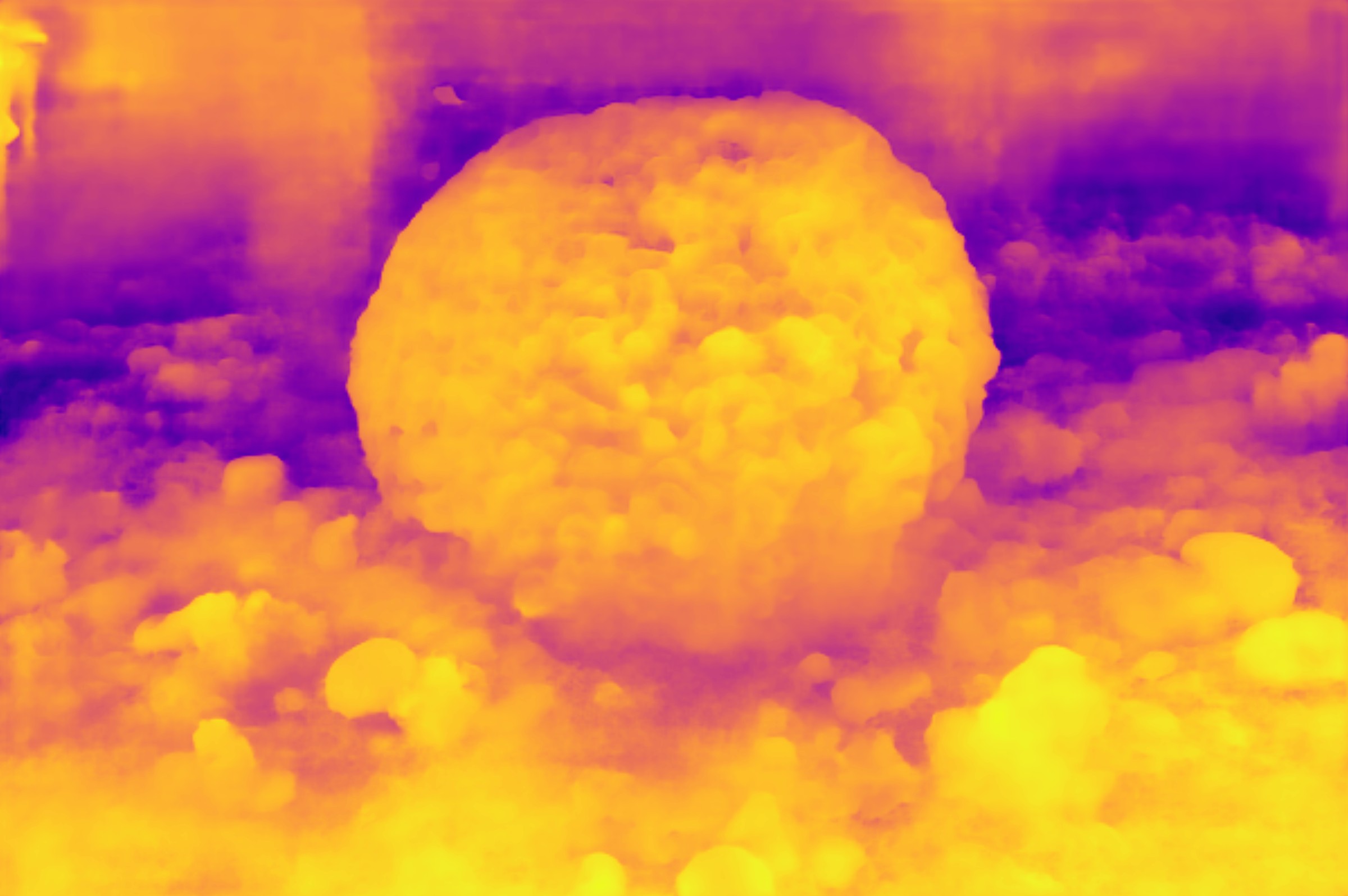} & 
		\includegraphics[width=0.21\linewidth]{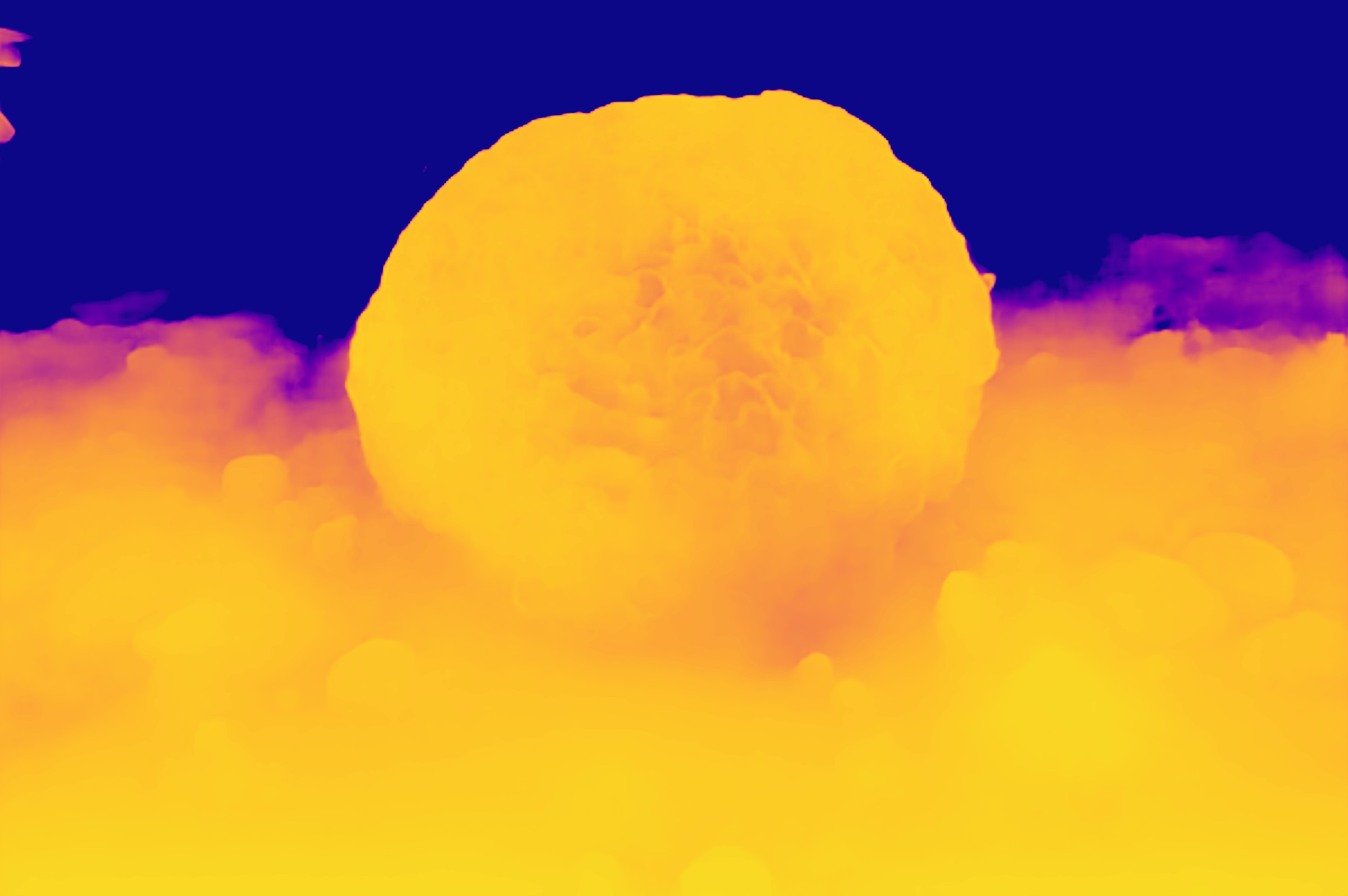}\\
		Input  & iDisc (K)  & iDisc (N)  & iDisc (A)\\
		
		\includegraphics[width=0.21\linewidth]{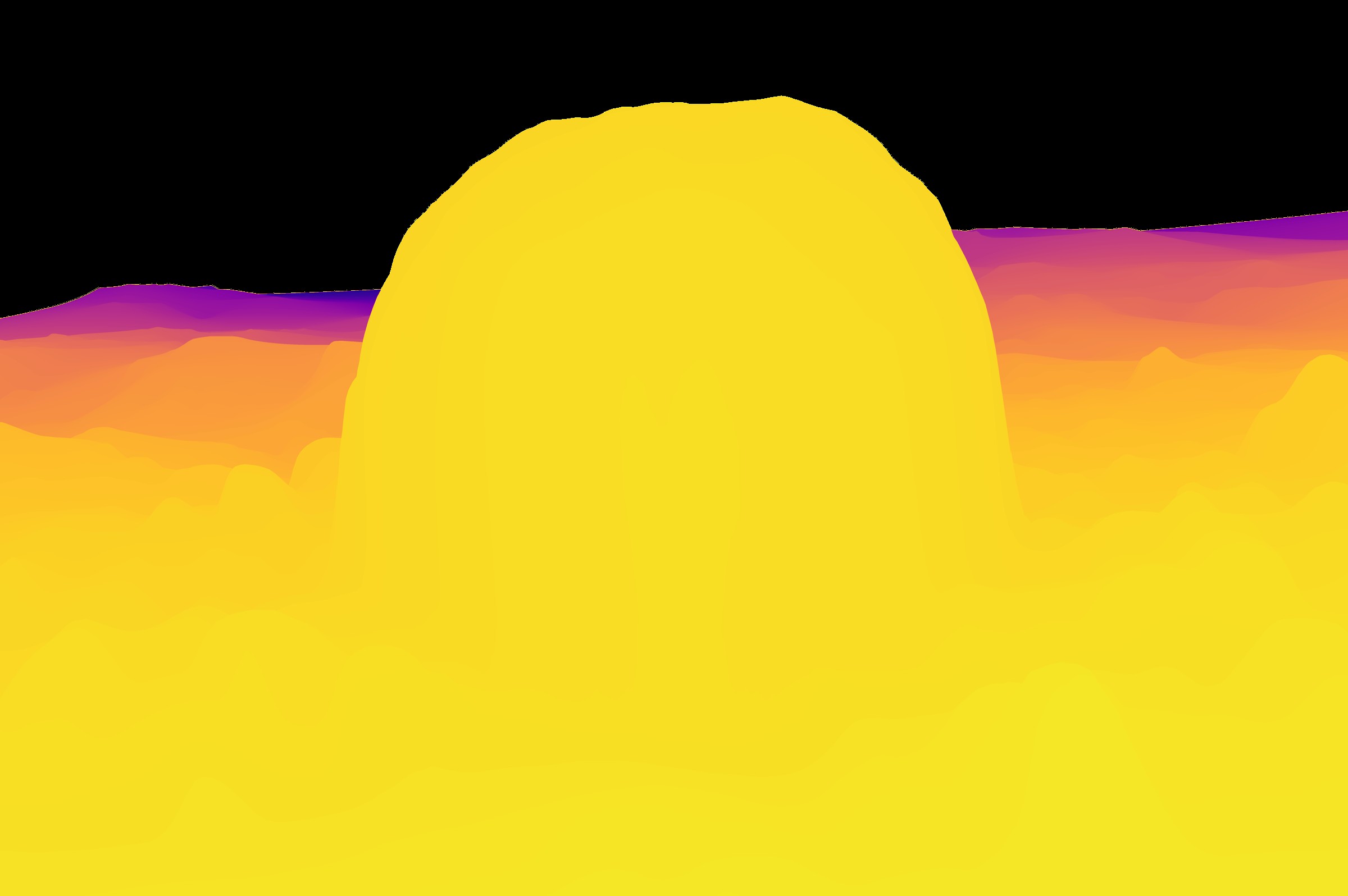}&		
		\includegraphics[width=0.21\linewidth]{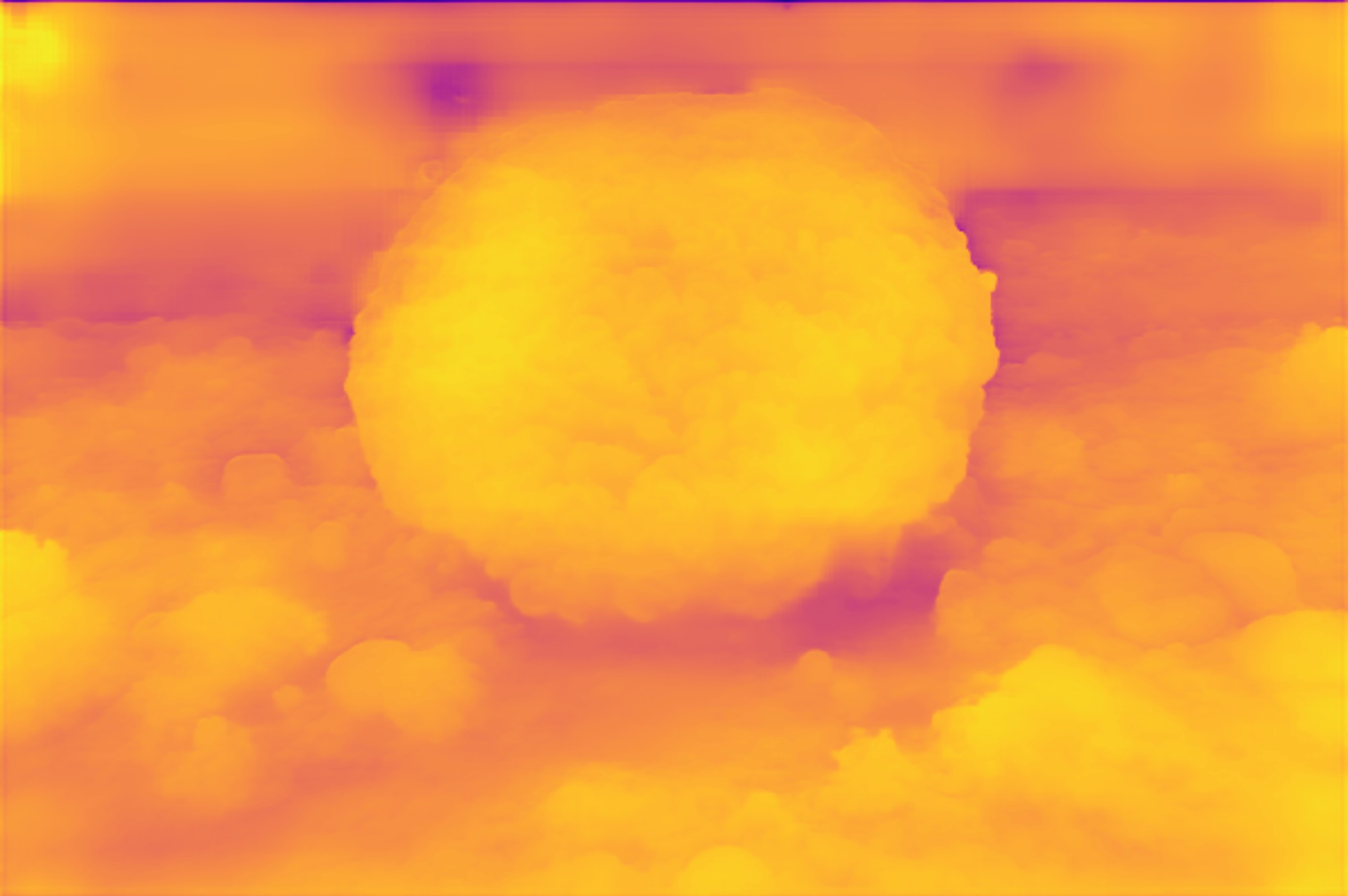} & 
		\includegraphics[width=0.21\linewidth]{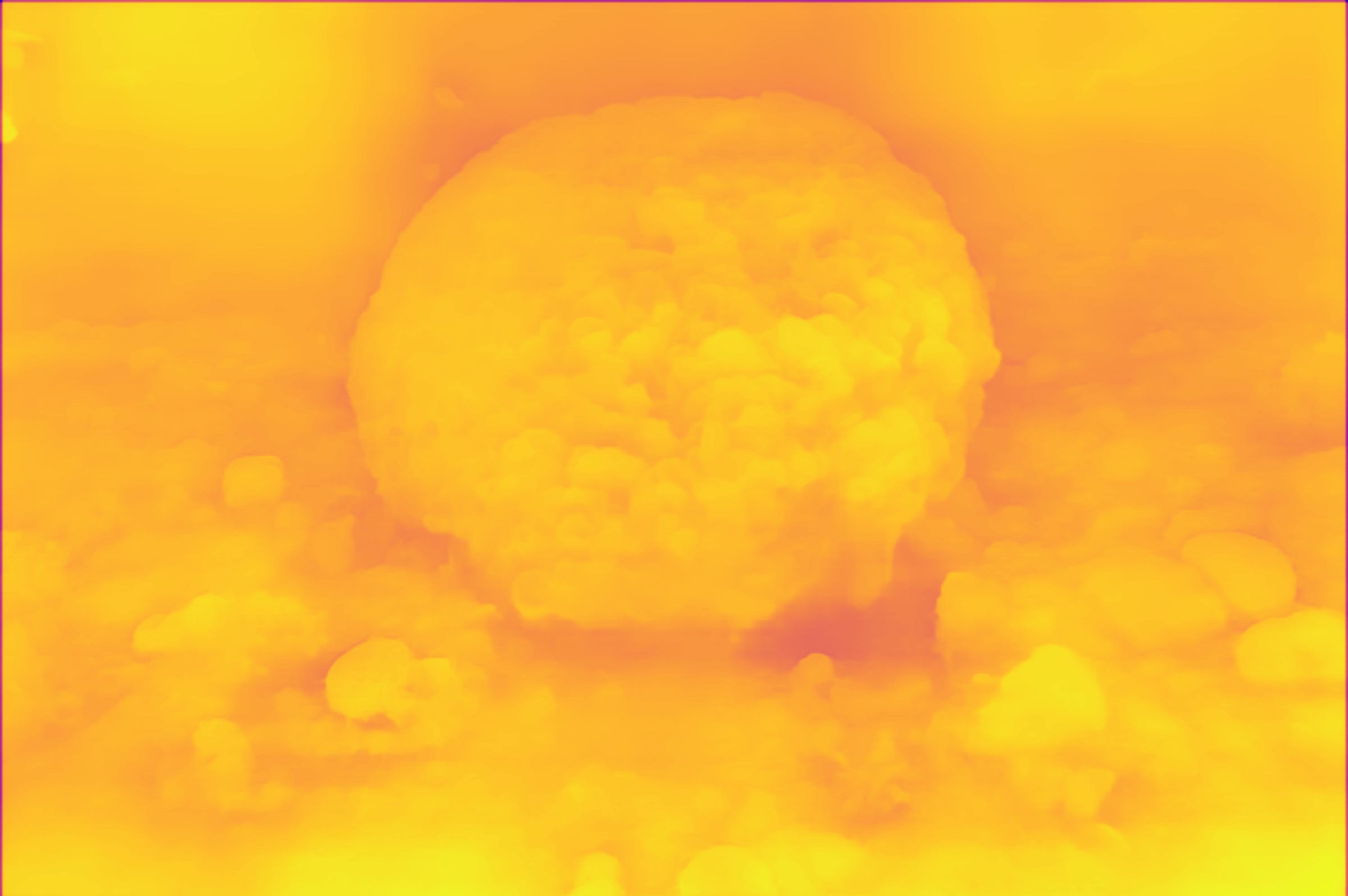} & 
		\includegraphics[width=0.21\linewidth]{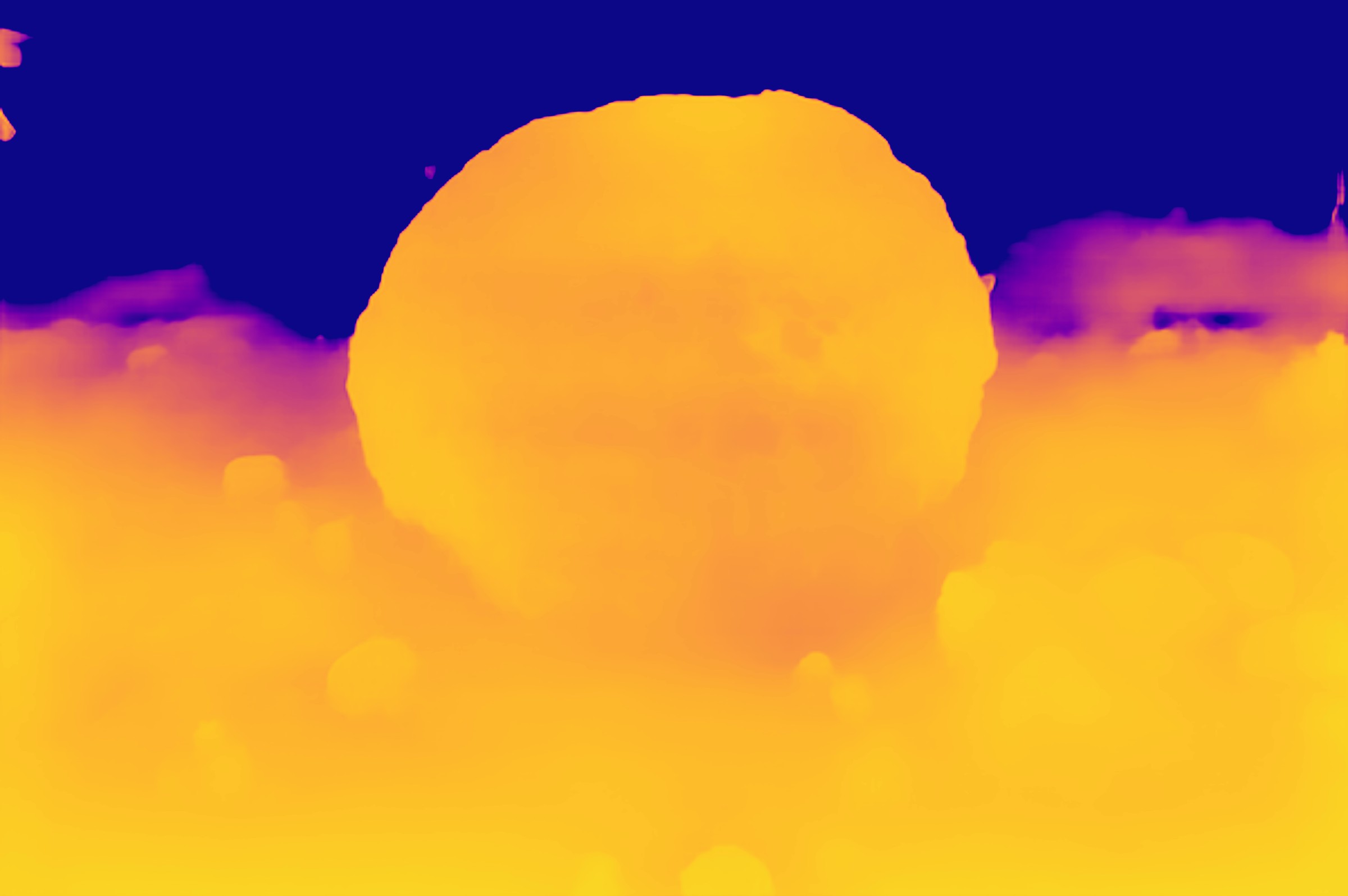}\\		
		 GT    & NeWCRFs (K)  & NeWCRFs (N)  & NeWCRFs (A) \\ 
		 
		 \includegraphics[width=0.21\linewidth]{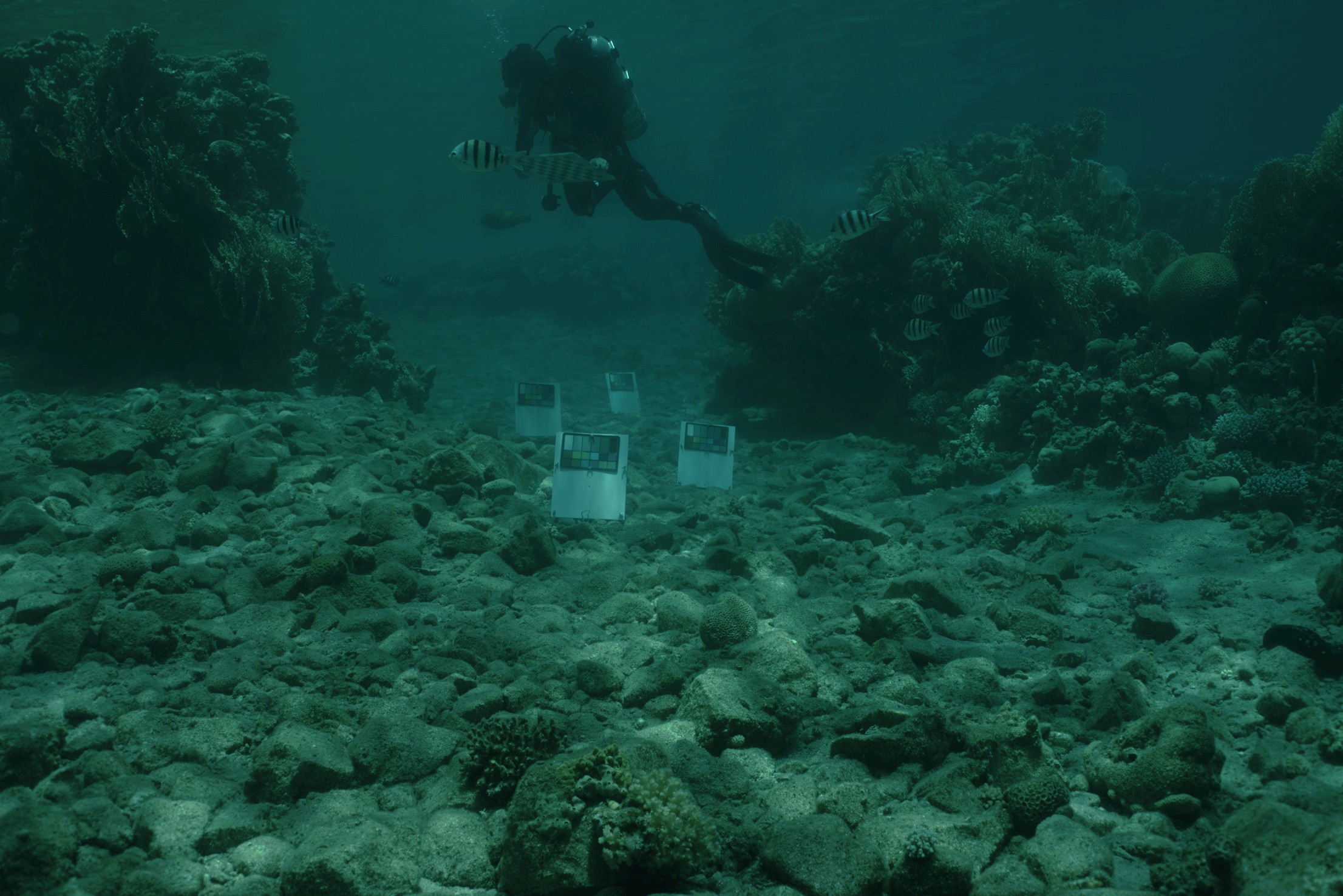}&
		 \includegraphics[width=0.21\linewidth]{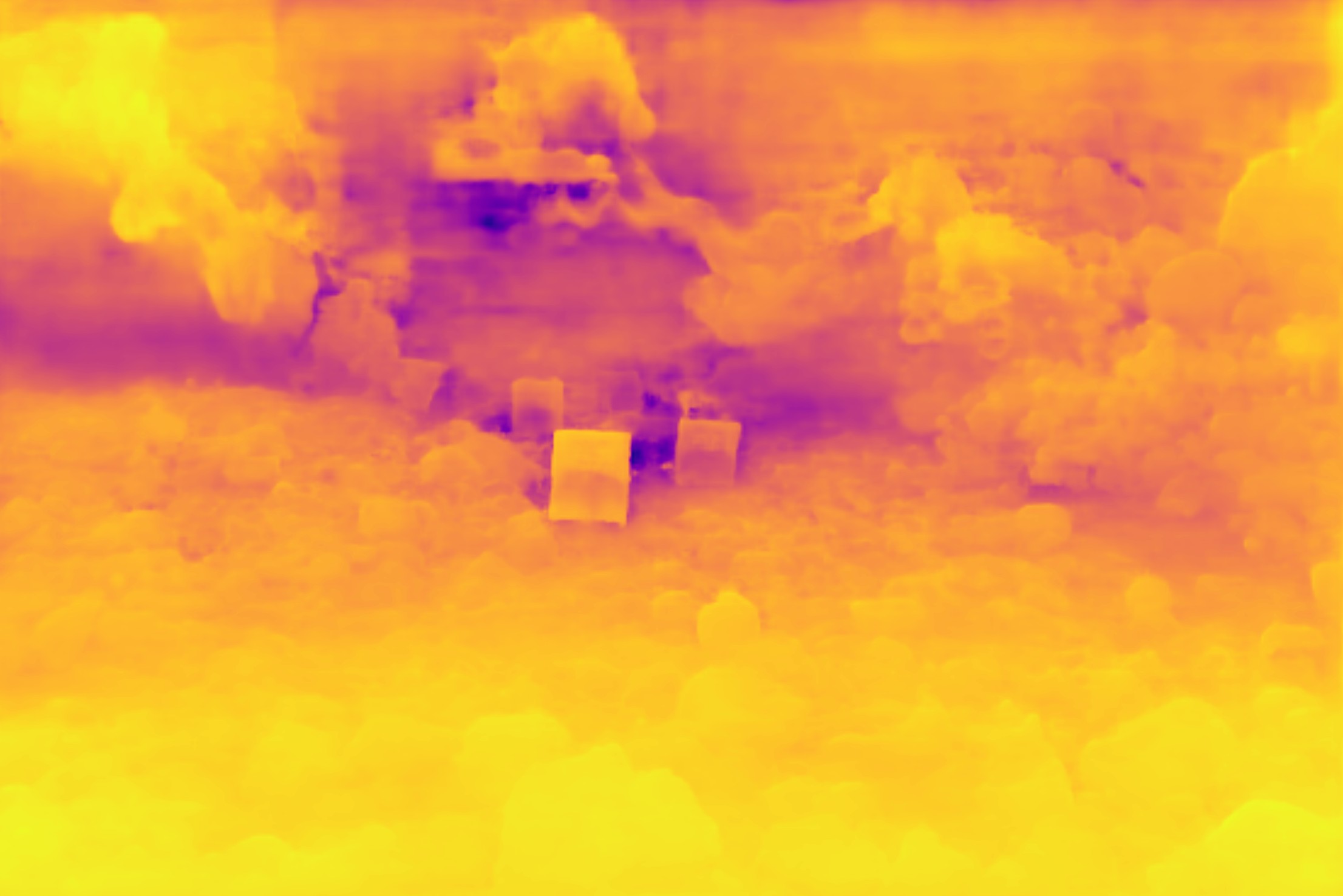} & 
		 \includegraphics[width=0.21\linewidth]{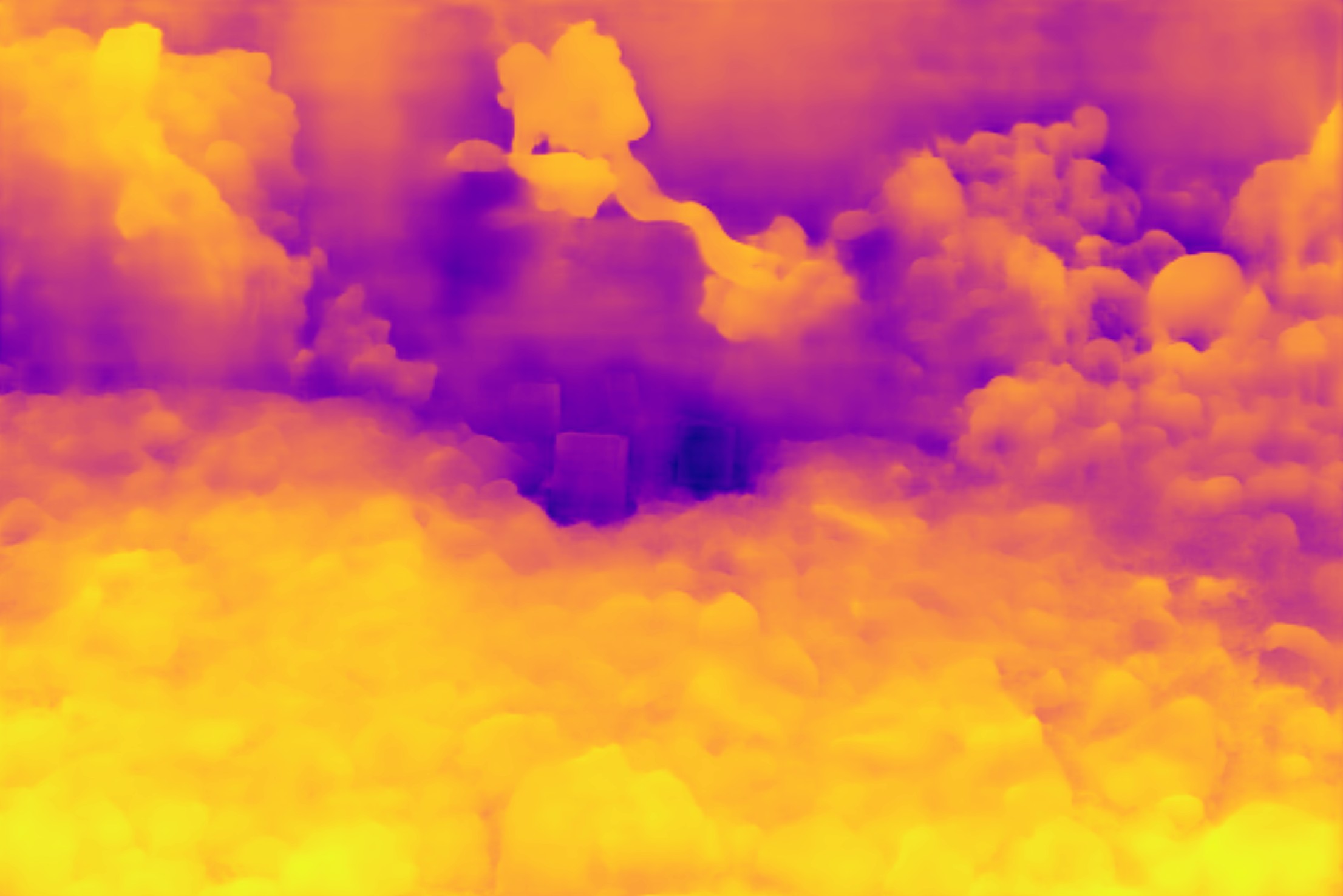} & 
		 \includegraphics[width=0.21\linewidth]{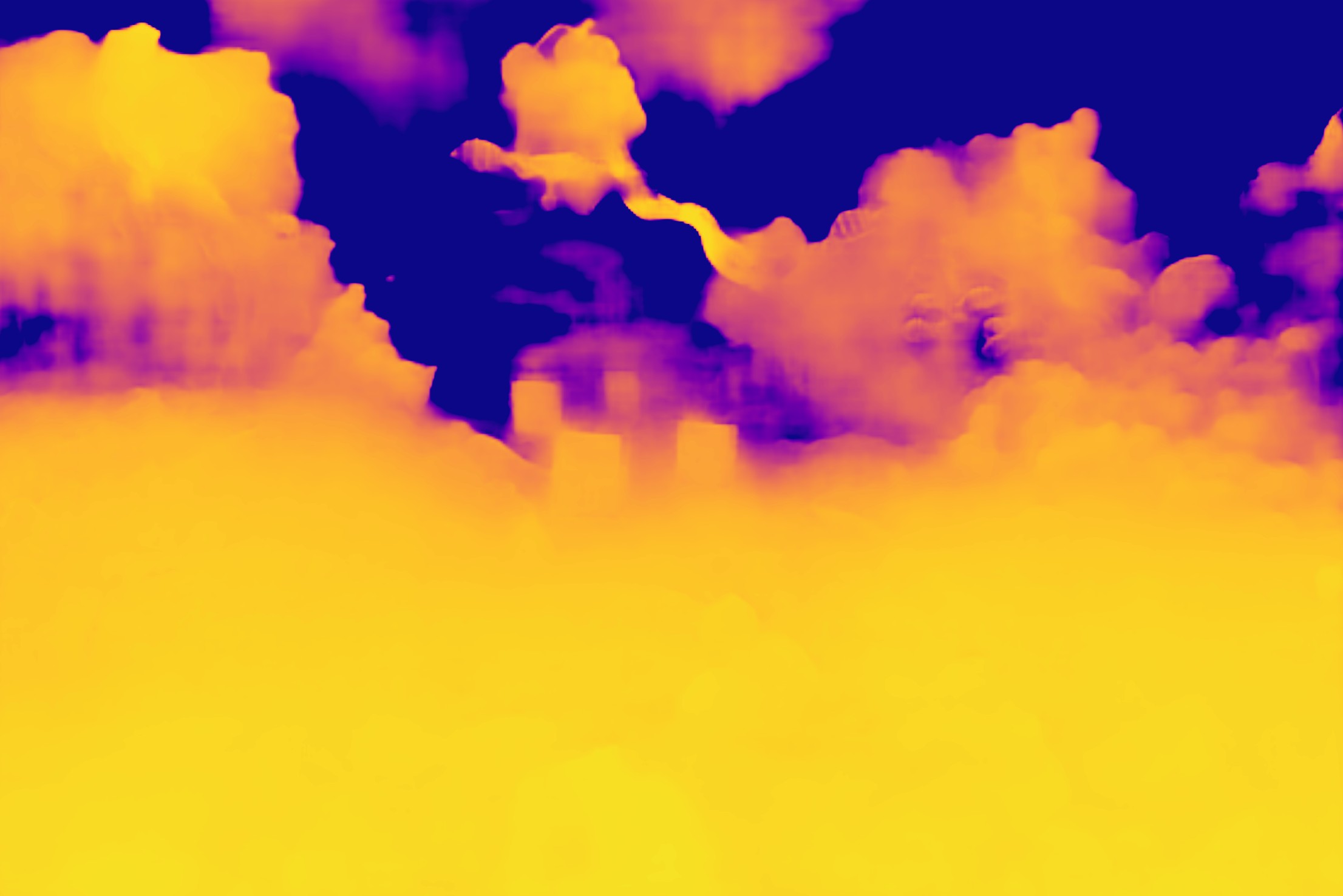}\\
		 Input  & iDisc (K)  & iDisc (N)  & iDisc (A)\\
		 
		 \includegraphics[width=0.21\linewidth]{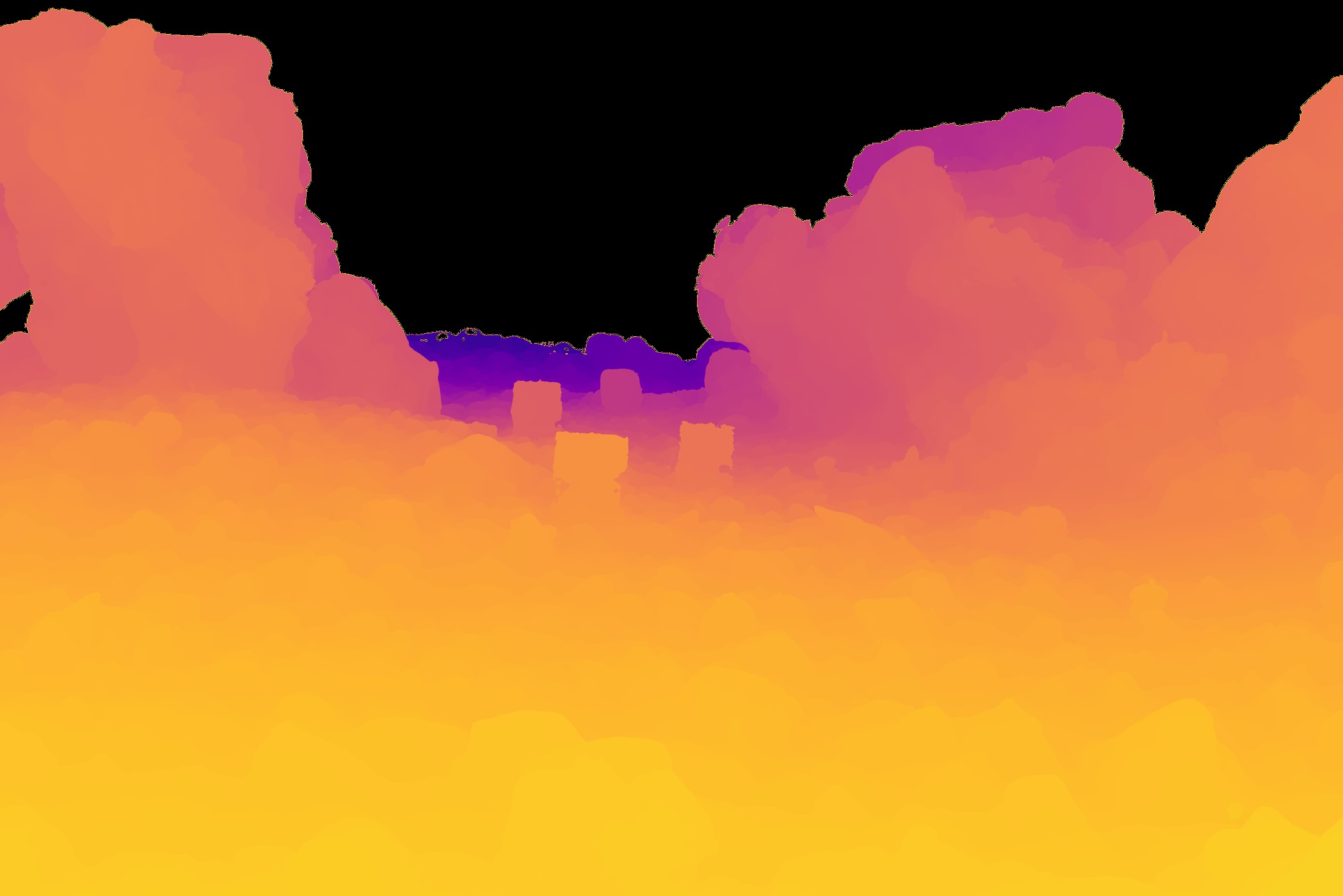}&		
		 \includegraphics[width=0.21\linewidth]{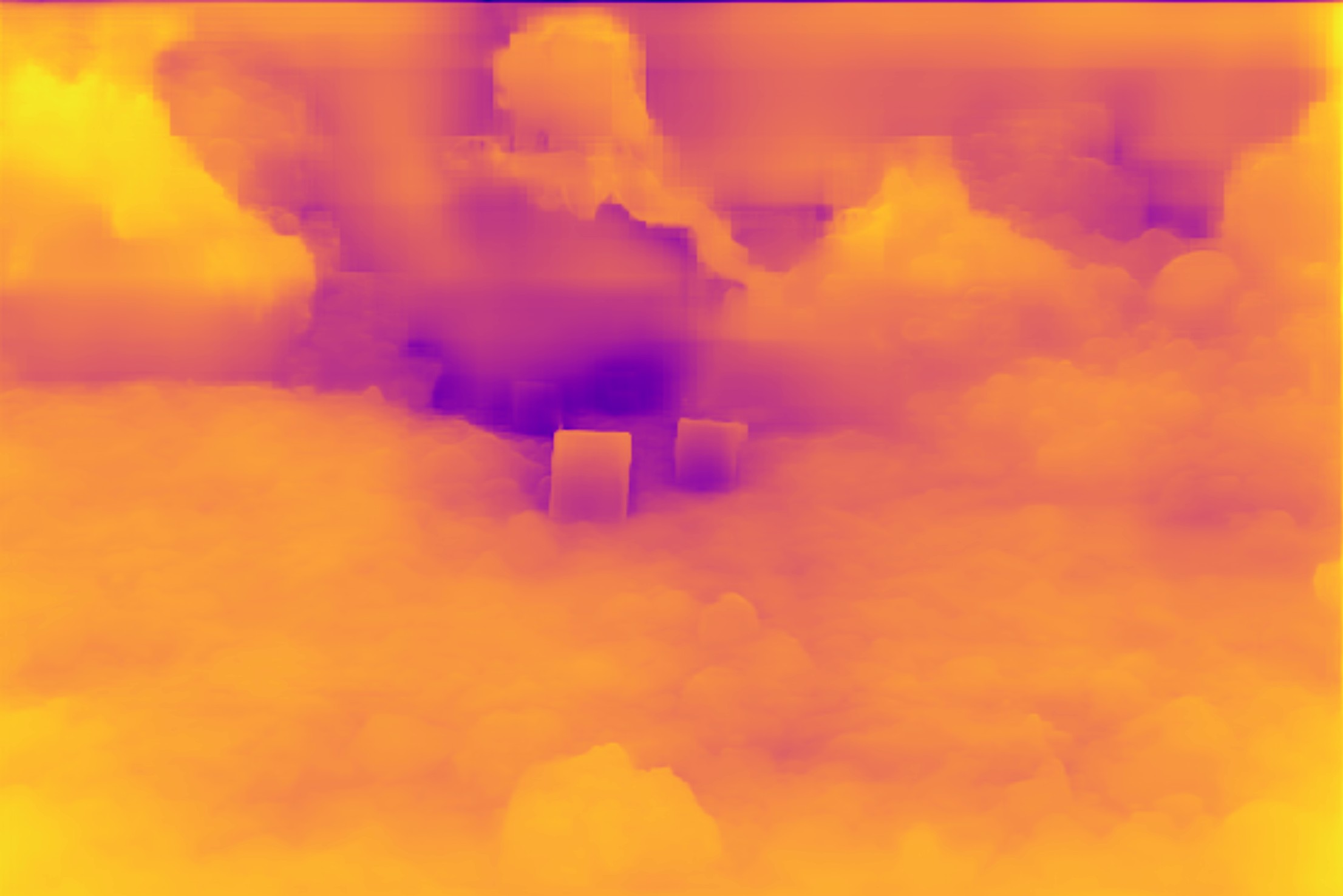} & 
		 \includegraphics[width=0.21\linewidth]{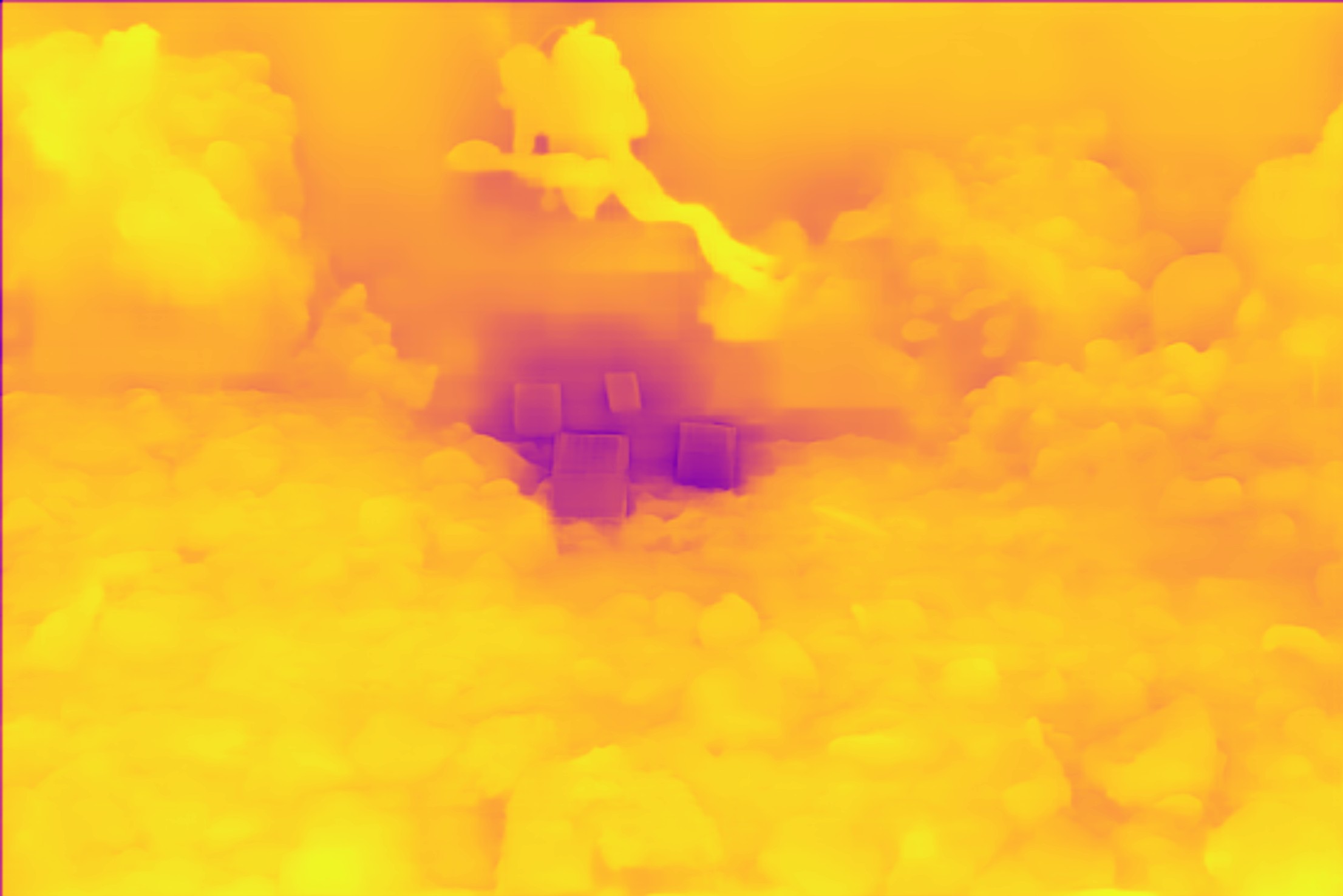} & 
		 \includegraphics[width=0.21\linewidth]{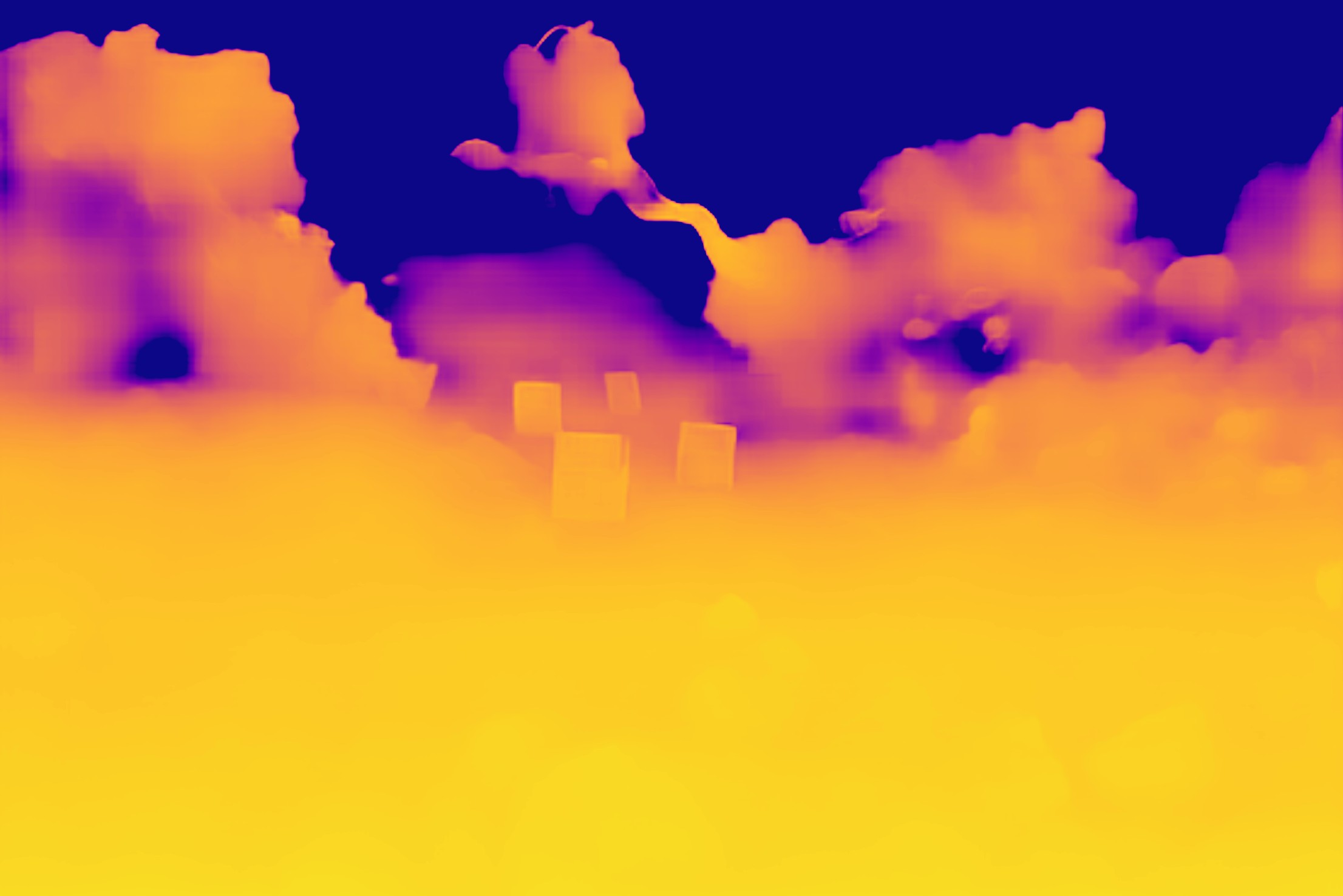}\\		
		 GT    & NeWCRFs (K)  & NeWCRFs (N)  & NeWCRFs (A) \\ 
		
	\end{tabular}
	\vspace{-2mm}
	\caption{Qualitative results on test set of Sea-thru dataset \cite{seathur2019akkaynak}. K and N denote models pretrained on KITTI \cite{kitti} and NYU Depthv2 \cite{nyudepthv2} datasets. A represents the models trained on our dataset Atlantis. Our method gets the best visual results. Please zoom in for details.} 
	\label{fig:seathru}
	\vspace{-3mm}
\end{figure*}

\begin{figure*}[ht]\small
	\centering
	\includegraphics[width=1\linewidth]{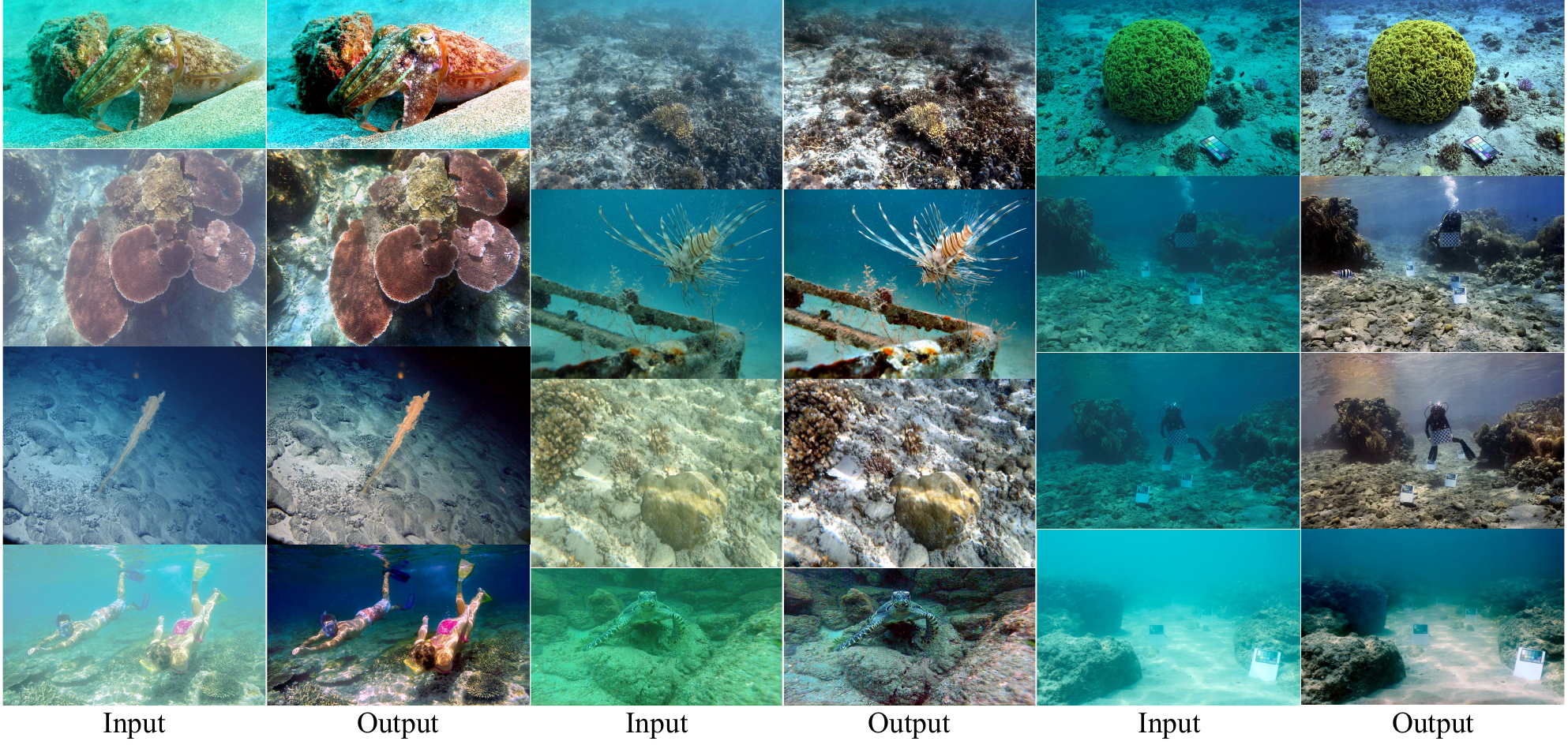}
	\vspace{-6mm}
	\caption{Qualitative results of the improved depth result applied to downstream underwater image enhancement. The left and middle parts are from UIEB dataset \cite{uieb2019li} and the right part contains images from Sea-thru dataset \cite{seathur2019akkaynak} (the above three) and SQUID dataset \cite{squid2020berman} (the bottom one). The enhancement method adopts the unofficial Sea-thru algorithm implementation\protect\footnotemark. Enhancement outputs well show the effectiveness of the proposed dataset on training depth models for reliable underwater depth estimation.}
	\label{fig:downstream}
	\vspace{-5mm}
\end{figure*}
\begin{figure}[t]\small
	\centering
	\setlength{\tabcolsep}{1pt}
	\begin{tabular}{c}
		\includegraphics[width=\linewidth]{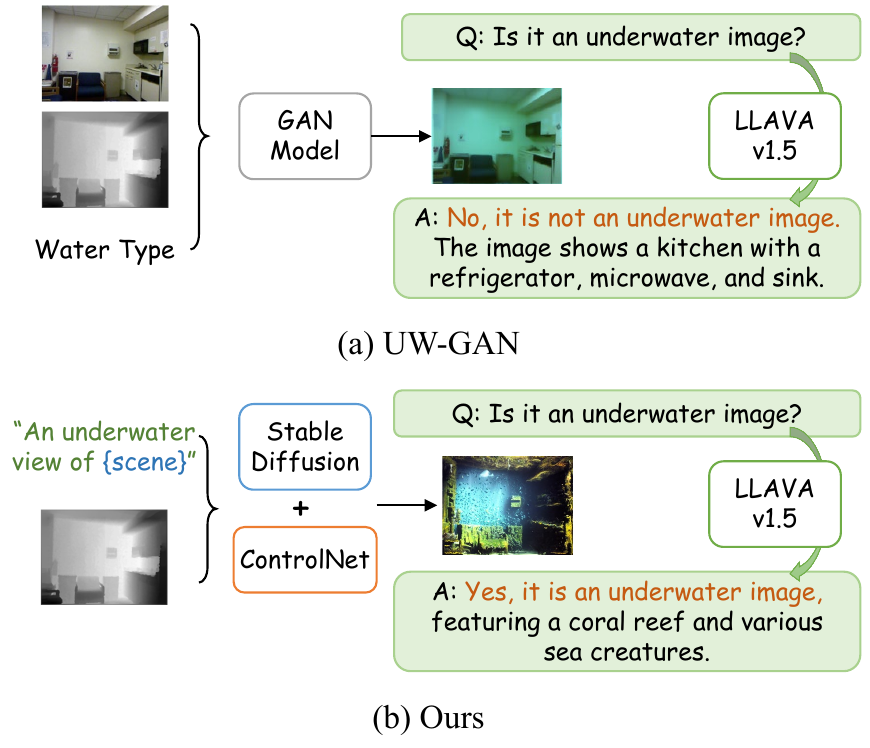}
	\end{tabular}
	\vspace{-2mm}
	\caption{(a) GAN-based methods style transfer images without changing the scene content, resulting in images still recognized as room scenes by LLVM. (b) In contrast, our method generates novel underwater images that maintain the original scene structure, correctly identified as underwater scenes by LLVM.}
	\label{fig:llava}
	\vspace{-5mm}
\end{figure}

\section{Experiments}
In this section, we demonstrate the effectiveness of our novel underwater depth dataset in training supervised depth estimation models, both quantitatively and qualitatively. We compare models trained from scratch on our dataset, specifically iDisc \cite{idisc2023piccinelli} and NeWCRFs \cite{newcrfs2022yuan}, with their officially pretrained counterparts on terrestrial datasets, namely KITTI \cite{kitti} and NYU Depthv2 \cite{nyudepthv2}. This comparison is conducted on unseen underwater datasets to highlight performance differences. Additionally, to showcase the practical application of depth models trained on our dataset, we utilize the Sea-thru algorithm \cite{seathur2019akkaynak}, originally designed for underwater image enhancement using depth maps obtained from stereo images or video sequences, and apply it to single images with estimated depth. Finally, we conduct a comparison to show smaller domain gap of our dataset compared to previous synthetic dataset.
Due to limited space, we provide more visual results in supplementary material.

\paragraph{Experimental Setup.} We focus on two models: iDisc \cite{idisc2023piccinelli} and NeWCRFs \cite{newcrfs2022yuan}. Both models were trained from scratch on our dataset and evaluated against their official versions pretrained on the KITTI \cite{kitti} and NYU Depthv2 \cite{nyudepthv2} datasets. Each utilizes the SwinL model model \cite{swin} pretrained on ImageNet 22k \cite{imagenet} for encoder initialization. 
Quantitatively, we conducted evaluations using the Sea-thru's D3 and D5 subsets \cite{seathur2019akkaynak} and the SQUID dataset \cite{squid2020berman}, which includes underwater images with depth maps obtained via Structure-from-Motion (SfM) algorithm.
For qualitative assessments, we addtionally selected the UIEB dataset's test set \cite{uieb2019li} to complement the diversity of tested scenes for visual comparison.
The metrics used for quantitative evaluation encompass root mean square error ($RMSE$) and its log variant ($RMSE_{log}$), absolute error in log-scale ($Log_{10}$), absolute ($A.Rel$) and squared ($S.Rel$) mean relative error, the percentage of inlier pixels ($\delta_i$) with threshold $1.25^i$, and scale-invariant error in log-scale ($SI_{log}$): $100\sqrt{Var(\epsilon_{log})}$.

\subsection{Quantitative Results}
The results, as detailed in Tables \ref{tab:quantitative1} and \ref{tab:quantitative2}, demonstrate a significant domain gap for models pretrained on terrestrial datasets of KITTI \cite{kitti} and NYU Depthv2 \cite{nyudepthv2} when applied to underwater images. This domain gap, which adversely affects performance across most metrics, was evident in both iDisc \cite{idisc2023piccinelli} and NeWCRFs \cite{newcrfs2022yuan} models, underscoring the inherent challenges in applying supervised monocular depth models to underwater scenes.
Conversely, when these models were trained from scratch on our underwater depth dataset, both iDisc and NeWCRFs exhibited substantial improvements across the majority of quantitative metrics. This improvement was consistent across evaluations on the Sea-thru \cite{seathur2019akkaynak} and SQUID \cite{squid2020berman} datasets, affirming the efficacy of our dataset in enhancing monocular depth estimation for unseen underwater scenes. This outcome suggests that training on our dataset effectively reduces the domain gap.
It is noteworthy that our dataset, despite being smaller in size compared to the terrestrial datasets, has shown significant potential in this context. This indicates that expanding the dataset further or employing hybrid training approaches could yield even more pronounced improvements.

\subsection{Qualitative Results}
Figures \ref{fig:uieb} and \ref{fig:seathru} showcase visual comparisons that highlight the stark contrast in depth estimation performance. For underwater images, pretrained models on terrestrial datasets, including both iDisc \cite{idisc2023piccinelli} and NeWCRFs \cite{newcrfs2022yuan}, produce significantly erroneous results. These inaccuracies manifest as heavy haze artifacts in water body areas and incorrect relative scene layout distances.
In sharp contrast, when trained on our specifically designed underwater depth dataset, both models exhibit a remarkable improvement in interpreting underwater scenes. Notably, they accurately identify and appropriately distance water body areas, demonstrating enhanced discrimination capabilities. The transitions in scene content are marked by clear borders, and the models adeptly handle transparent water with varying color casts. Overall, the layout of underwater scenes is more accurately rendered, and depth ambiguities, particularly in water bodies, are substantially reduced.
This improvement underscores the effectiveness of our dataset in enabling depth estimation models to better differentiate water bodies and adapt to diverse underwater conditions, including color casts and lighting variations. It's important to note that the underwater images used in these comparisons were not part of the training dataset. This further emphasizes the generalization capability of our dataset in training robust depth estimation models that effectively adapt to real underwater scenes.

\subsection{Sea-thru Enhancement with Depth Models}
We demonstrate the application of our dataset in training reliable underwater depth models for effective underwater image enhancement using the Sea-thru algorithm \cite{seathur2019akkaynak}. Known for its ability to remove water effects with precise range maps derived from stereo pairs or video sequences, Sea-thru's capabilities are extended to single underwater images using depth maps estimated by models trained on our dataset.  
As depicted in Figure \ref{fig:downstream}, the Sea-thru algorithm, when equipped with depth estimates from our models, produces impressive underwater image enhancements. These results not only showcase the models' accuracy in depth estimation but also reaffirm the practical utility and effectiveness of our dataset in real-world applications.

\subsection{Domain Gap from LLVM Perspective}
\label{sec:domain}

The advent of Large Language Vision Models (LLVM) \cite{clip, llava} has revolutionized the alignment between textual and visual features, opening new avenues for synthetic data analysis. We utilize recent LLAVA v1.5 model \cite{llava} to reveal a disconnect in how LLVM perceives synthetic underwater images generated using conventional depth and image formation models.
\footnotetext{https://github.com/hainh/sea-thru} As depicted in Figure \ref{fig:llava}\textcolor{red}{(a)}, these images are often not recognized as underwater scenes, signaling a gap in the current synthesis approach. However, SD \cite{stablediffusion} and ControlNet \cite{controlnet2023zhang} stand out in their ability to generate highly realistic images guided by textual prompts, a testament to the advancements in aligning natural language with visual content. Figure \ref{fig:llava}\textcolor{red}{(b)} illustrates how LLAVA effectively recognizes images generated through this method as authentic underwater scenes, confirming smaller domain gap of our proposed dataset.

\section{Conclustion}

In this paper, we introduced a novel pipeline utilizing Stable Diffusion and a customized ControlNet for generating realistic underwater images with accurate depth. We proposed a dataset, Atlantis, to enable the training of terrestrial depth models for underwater depth estimation, which significantly enhances their performance on underwater scenes. The proposed dataset features easy acquisition, realistic underwater images and accurate depth, large diversity and theoretically unlimited scale. Our experiments, encompassing both quantitative and qualitative analyses, demonstrated the superiority of models trained on our dataset compared to those pretrained on terrestrial datasets. Notably, the application of these models in the Sea-thru algorithm for single underwater image enhancement showcased their practical utility and highlighted the value of our dataset. Our study reveals the potential of SD to be a new source of high quality training data.
As future work, expanding the dataset and exploring hybrid training approaches could unlock greater improvements in model performance and generalization.

%% file: main.bbl
\begin{thebibliography}{49}
\providecommand{\natexlab}[1]{#1}
\providecommand{\url}[1]{\texttt{#1}}
\expandafter\ifx\csname urlstyle\endcsname\relax
  \providecommand{\doi}[1]{doi: #1}\else
  \providecommand{\doi}{doi: \begingroup \urlstyle{rm}\Url}\fi

\bibitem[Akkaynak and Treibitz(2018)]{ifm2018akkaynak}
Derya Akkaynak and Tali Treibitz.
\newblock A revised underwater image formation model.
\newblock In \emph{CVPR}, 2018.

\bibitem[Akkaynak and Treibitz(2019)]{seathur2019akkaynak}
Derya Akkaynak and Tali Treibitz.
\newblock Sea-thru: A method for removing water from underwater images.
\newblock In \emph{CVPR}, 2019.

\bibitem[Amitai et~al.(2023)Amitai, Klein, and Treibitz]{depth_self2023amitai}
Shlomi Amitai, Itzik Klein, and Tali Treibitz.
\newblock Self-supervised monocular depth underwater.
\newblock In \emph{ICRA}, 2023.

\bibitem[Bailey and Flemming(2008)]{marine1}
Geoffrey~N Bailey and Nicholas~C Flemming.
\newblock Archaeology of the continental shelf: marine resources, submerged
  landscapes and underwater archaeology.
\newblock \emph{Quaternary Science Reviews}, 27\penalty0 (23-24):\penalty0
  2153--2165, 2008.

\bibitem[Berman et~al.(2020)Berman, Levy, Avidan, and
  Treibitz]{squid2020berman}
Dana Berman, Deborah Levy, Shai Avidan, and Tali Treibitz.
\newblock Underwater single image color restoration using haze-lines and a new
  quantitative dataset.
\newblock \emph{IEEE TPAMI}, 43\penalty0 (8):\penalty0 2822--2837, 2020.

\bibitem[Bhat et~al.(2021)Bhat, Alhashim, and Wonka]{adabins2021bhat}
Shariq~Farooq Bhat, Ibraheem Alhashim, and Peter Wonka.
\newblock Adabins: Depth estimation using adaptive bins.
\newblock In \emph{CVPR}, 2021.

\bibitem[Bhat et~al.(2023)Bhat, Birkl, Wofk, Wonka, and M{\"u}ller]{zoedepth}
Shariq~Farooq Bhat, Reiner Birkl, Diana Wofk, Peter Wonka, and Matthias
  M{\"u}ller.
\newblock Zoedepth: Zero-shot transfer by combining relative and metric depth.
\newblock \emph{arXiv preprint arXiv:2302.12288}, 2023.

\bibitem[Blidberg(2001)]{auv2}
D~Richard Blidberg.
\newblock The development of autonomous underwater vehicles (auv); a brief
  summary.
\newblock In \emph{Ieee Icra}, 2001.

\bibitem[Chiang and Chen(2011)]{chiang2011underwater}
John~Y Chiang and Ying-Ching Chen.
\newblock Underwater image enhancement by wavelength compensation and dehazing.
\newblock \emph{IEEE TIP}, 21\penalty0 (4):\penalty0 1756--1769, 2011.

\bibitem[Coleman et~al.(2000)Coleman, Newman, and Ballard]{marine2}
Dwight~F Coleman, James~B Newman, and Robert~D Ballard.
\newblock Design and implementation of advanced underwater imaging systems for
  deep sea marine archaeological surveys.
\newblock In \emph{OCEANS MTS/IEEE Conference and Exhibition}, 2000.

\bibitem[Deng et~al.(2009)Deng, Dong, Socher, Li, Li, and Fei-Fei]{imagenet}
Jia Deng, Wei Dong, Richard Socher, Li-Jia Li, Kai Li, and Li Fei-Fei.
\newblock Imagenet: A large-scale hierarchical image database.
\newblock In \emph{CVPR}, 2009.

\bibitem[Drews et~al.(2016{\natexlab{a}})Drews, Nascimento, Botelho, and
  Campos]{depth_uie2016drews}
Paulo~LJ Drews, Erickson~R Nascimento, Silvia~SC Botelho, and Mario
  Fernando~Montenegro Campos.
\newblock Underwater depth estimation and image restoration based on single
  images.
\newblock \emph{IEEE computer graphics and applications}, 36\penalty0
  (2):\penalty0 24--35, 2016{\natexlab{a}}.

\bibitem[Drews et~al.(2016{\natexlab{b}})Drews, Nascimento, Botelho, and
  Campos]{udcp}
Paulo~LJ Drews, Erickson~R Nascimento, Silvia~SC Botelho, and Mario
  Fernando~Montenegro Campos.
\newblock Underwater depth estimation and image restoration based on single
  images.
\newblock \emph{IEEE computer graphics and applications}, 36\penalty0
  (2):\penalty0 24--35, 2016{\natexlab{b}}.

\bibitem[Duntley(1963)]{duntley1963light}
Seibert~Q Duntley.
\newblock Light in the sea.
\newblock \emph{JOSA}, 53\penalty0 (2):\penalty0 214--233, 1963.

\bibitem[Eigen et~al.(2014)Eigen, Puhrsch, and Fergus]{eigen2014depth}
David Eigen, Christian Puhrsch, and Rob Fergus.
\newblock Depth map prediction from a single image using a multi-scale deep
  network.
\newblock \emph{NeurIPS}, 2014.

\bibitem[Filisetti et~al.(2018)Filisetti, Marouchos, Martini, Martin, and
  Collings]{lidar2}
Andrew Filisetti, Andreas Marouchos, Andrew Martini, Tara Martin, and Simon
  Collings.
\newblock Developments and applications of underwater lidar systems in support
  of marine science.
\newblock In \emph{OCEANS MTS/IEEE Charleston}, 2018.

\bibitem[Fu et~al.(2018)Fu, Gong, Wang, Batmanghelich, and Tao]{dorn}
Huan Fu, Mingming Gong, Chaohui Wang, Kayhan Batmanghelich, and Dacheng Tao.
\newblock Deep ordinal regression network for monocular depth estimation.
\newblock In \emph{CVPR}, 2018.

\bibitem[Geiger et~al.(2012)Geiger, Lenz, and Urtasun]{kitti}
Andreas Geiger, Philip Lenz, and Raquel Urtasun.
\newblock Are we ready for autonomous driving? the kitti vision benchmark
  suite.
\newblock In \emph{CVPR}, 2012.

\bibitem[Gibson et~al.(2016)Gibson, Atkinson, and Gordon]{marine3}
R Gibson, R Atkinson, and J Gordon.
\newblock A review of underwater stereo-image measurement for marine biology
  and ecology applications.
\newblock \emph{Oceanography and marine biology: an annual review},
  47:\penalty0 257--292, 2016.

\bibitem[Godard et~al.(2017)Godard, Mac~Aodha, and Brostow]{monodepth}
Cl{\'e}ment Godard, Oisin Mac~Aodha, and Gabriel~J Brostow.
\newblock Unsupervised monocular depth estimation with left-right consistency.
\newblock In \emph{CVPR}, 2017.

\bibitem[Godard et~al.(2019)Godard, Mac~Aodha, Firman, and Brostow]{monodepth2}
Cl{\'e}ment Godard, Oisin Mac~Aodha, Michael Firman, and Gabriel~J Brostow.
\newblock Digging into self-supervised monocular depth estimation.
\newblock In \emph{ICCV}, 2019.

\bibitem[Gupta and Mitra(2019)]{uwnet2019gupta}
Honey Gupta and Kaushik Mitra.
\newblock Unsupervised single image underwater depth estimation.
\newblock In \emph{ICIP}, 2019.

\bibitem[Hambarde et~al.(2021)Hambarde, Murala, and Dhall]{depth_uwgan}
Praful Hambarde, Subrahmanyam Murala, and Abhinav Dhall.
\newblock Uw-gan: Single-image depth estimation and image enhancement for
  underwater images.
\newblock \emph{IEEE Transactions on Instrumentation and Measurement},
  70:\penalty0 1--12, 2021.

\bibitem[He et~al.(2010)He, Sun, and Tang]{dcp}
Kaiming He, Jian Sun, and Xiaoou Tang.
\newblock Single image haze removal using dark channel prior.
\newblock \emph{IEEE TPAMI}, 33\penalty0 (12):\penalty0 2341--2353, 2010.

\bibitem[Jaffe(1990)]{jaffe1990computer}
Jules~S Jaffe.
\newblock Computer modeling and the design of optimal underwater imaging
  systems.
\newblock \emph{IEEE Journal of Oceanic Engineering}, 15\penalty0 (2):\penalty0
  101--111, 1990.

\bibitem[Li et~al.(2019)Li, Guo, Ren, Cong, Hou, Kwong, and Tao]{uieb2019li}
Chongyi Li, Chunle Guo, Wenqi Ren, Runmin Cong, Junhui Hou, Sam Kwong, and
  Dacheng Tao.
\newblock An underwater image enhancement benchmark dataset and beyond.
\newblock \emph{IEEE TIP}, 29:\penalty0 4376--4389, 2019.

\bibitem[Li et~al.(2020)Li, Anwar, and Porikli]{uwcnn}
Chongyi Li, Saeed Anwar, and Fatih Porikli.
\newblock Underwater scene prior inspired deep underwater image and video
  enhancement.
\newblock \emph{PR}, 98:\penalty0 107038, 2020.

\bibitem[Li et~al.(2023)Li, Li, Savarese, and Hoi]{blip2}
Junnan Li, Dongxu Li, Silvio Savarese, and Steven Hoi.
\newblock Blip-2: Bootstrapping language-image pre-training with frozen image
  encoders and large language models.
\newblock \emph{arXiv preprint arXiv:2301.12597}, 2023.

\bibitem[Liu et~al.(2023)Liu, Li, Wu, and Lee]{llava}
Haotian Liu, Chunyuan Li, Qingyang Wu, and Yong~Jae Lee.
\newblock Visual instruction tuning.
\newblock \emph{arXiv preprint arXiv:2304.08485}, 2023.

\bibitem[Liu et~al.(2021)Liu, Lin, Cao, Hu, Wei, Zhang, Lin, and Guo]{swin}
Ze Liu, Yutong Lin, Yue Cao, Han Hu, Yixuan Wei, Zheng Zhang, Stephen Lin, and
  Baining Guo.
\newblock Swin transformer: Hierarchical vision transformer using shifted
  windows.
\newblock In \emph{ICCV}, 2021.

\bibitem[McGlamery(1980)]{mcglamery1980computer}
BL McGlamery.
\newblock A computer model for underwater camera systems.
\newblock In \emph{Ocean Optics VI}, pages 221--231. SPIE, 1980.

\bibitem[Paull et~al.(2013)Paull, Saeedi, Seto, and Li]{auv1}
Liam Paull, Sajad Saeedi, Mae Seto, and Howard Li.
\newblock Auv navigation and localization: A review.
\newblock \emph{IEEE Journal of oceanic engineering}, 39\penalty0 (1):\penalty0
  131--149, 2013.

\bibitem[Piccinelli et~al.(2023)Piccinelli, Sakaridis, and
  Yu]{idisc2023piccinelli}
Luigi Piccinelli, Christos Sakaridis, and Fisher Yu.
\newblock idisc: Internal discretization for monocular depth estimation.
\newblock In \emph{CVPR}, 2023.

\bibitem[Poggi et~al.(2020)Poggi, Aleotti, Tosi, and Mattoccia]{uncertainty}
Matteo Poggi, Filippo Aleotti, Fabio Tosi, and Stefano Mattoccia.
\newblock On the uncertainty of self-supervised monocular depth estimation.
\newblock In \emph{CVPR}, 2020.

\bibitem[Radford et~al.(2021)Radford, Kim, Hallacy, Ramesh, Goh, Agarwal,
  Sastry, Askell, Mishkin, Clark, et~al.]{clip}
Alec Radford, Jong~Wook Kim, Chris Hallacy, Aditya Ramesh, Gabriel Goh,
  Sandhini Agarwal, Girish Sastry, Amanda Askell, Pamela Mishkin, Jack Clark,
  et~al.
\newblock Learning transferable visual models from natural language
  supervision.
\newblock 2021.

\bibitem[Ranftl et~al.(2020)Ranftl, Lasinger, Hafner, Schindler, and
  Koltun]{midas}
Ren{\'e} Ranftl, Katrin Lasinger, David Hafner, Konrad Schindler, and Vladlen
  Koltun.
\newblock Towards robust monocular depth estimation: Mixing datasets for
  zero-shot cross-dataset transfer.
\newblock \emph{IEEE TPAMI}, 44\penalty0 (3):\penalty0 1623--1637, 2020.

\bibitem[Ranftl et~al.(2021)Ranftl, Bochkovskiy, and Koltun]{dpt}
Ren{\'e} Ranftl, Alexey Bochkovskiy, and Vladlen Koltun.
\newblock Vision transformers for dense prediction.
\newblock In \emph{ICCV}, 2021.

\bibitem[Rombach et~al.(2022)Rombach, Blattmann, Lorenz, Esser, and
  Ommer]{stablediffusion}
Robin Rombach, Andreas Blattmann, Dominik Lorenz, Patrick Esser, and Bj{\"o}rn
  Ommer.
\newblock High-resolution image synthesis with latent diffusion models.
\newblock In \emph{CVPR}, 2022.

\bibitem[Silberman et~al.(2012)Silberman, Hoiem, Kohli, and Fergus]{nyudepthv2}
Nathan Silberman, Derek Hoiem, Pushmeet Kohli, and Rob Fergus.
\newblock Indoor segmentation and support inference from rgbd images.
\newblock In \emph{ECCV}, 2012.

\bibitem[Song et~al.(2018)Song, Wang, Huang, and
  Tjondronegoro]{depth_uie2018song}
Wei Song, Yan Wang, Dongmei Huang, and Dian Tjondronegoro.
\newblock A rapid scene depth estimation model based on underwater light
  attenuation prior for underwater image restoration.
\newblock In \emph{Advances in Multimedia Information Processing--PCM 2018:
  19th Pacific-Rim Conference on Multimedia, Hefei, China, September 21-22,
  2018, Proceedings, Part I 19}, 2018.

\bibitem[Varghese et~al.(2023)Varghese, Kumar, and
  Rajagopalan]{depth_self2023varghese}
Nisha Varghese, Ashish Kumar, and AN Rajagopalan.
\newblock Self-supervised monocular underwater depth recovery, image
  restoration, and a real-sea video dataset.
\newblock In \emph{ICCV}, 2023.

\bibitem[Vasiljevic et~al.(2019)Vasiljevic, Kolkin, Zhang, Luo, Wang, Dai,
  Daniele, Mostajabi, Basart, Walter, et~al.]{diode}
Igor Vasiljevic, Nick Kolkin, Shanyi Zhang, Ruotian Luo, Haochen Wang, Falcon~Z
  Dai, Andrea~F Daniele, Mohammadreza Mostajabi, Steven Basart, Matthew~R
  Walter, et~al.
\newblock Diode: A dense indoor and outdoor depth dataset.
\newblock \emph{arXiv preprint arXiv:1908.00463}, 2019.

\bibitem[von Platen et~al.(2022)von Platen, Patil, Lozhkov, Cuenca, Lambert,
  Rasul, Davaadorj, and Wolf]{diffusers}
Patrick von Platen, Suraj Patil, Anton Lozhkov, Pedro Cuenca, Nathan Lambert,
  Kashif Rasul, Mishig Davaadorj, and Thomas Wolf.
\newblock Diffusers: State-of-the-art diffusion models.
\newblock \url{https://github.com/huggingface/diffusers}, 2022.

\bibitem[Yang et~al.(2022)Yang, Zhang, Wang, Xin, and Hu]{depth_self2022yang}
Xuewen Yang, Xing Zhang, Nan Wang, Guoling Xin, and Wenjie Hu.
\newblock Underwater self-supervised depth estimation.
\newblock \emph{Neurocomputing}, 514:\penalty0 362--373, 2022.

\bibitem[Yu et~al.(2023)Yu, Wu, and Islam]{udepth2023yu}
Boxiao Yu, Jiayi Wu, and Md~Jahidul Islam.
\newblock Udepth: Fast monocular depth estimation for visually-guided
  underwater robots.
\newblock In \emph{ICRA}, 2023.

\bibitem[Yuan et~al.(2022)Yuan, Gu, Dai, Zhu, and Tan]{newcrfs2022yuan}
Weihao Yuan, Xiaodong Gu, Zuozhuo Dai, Siyu Zhu, and Ping Tan.
\newblock Neural window fully-connected crfs for monocular depth estimation.
\newblock In \emph{CVPR}, 2022.

\bibitem[Yuh and West(2001)]{auv3}
Junku Yuh and Michael West.
\newblock Underwater robotics.
\newblock \emph{Advanced Robotics}, 15\penalty0 (5):\penalty0 609--639, 2001.

\bibitem[Zhang et~al.(2023)Zhang, Rao, and Agrawala]{controlnet2023zhang}
Lvmin Zhang, Anyi Rao, and Maneesh Agrawala.
\newblock Adding conditional control to text-to-image diffusion models.
\newblock In \emph{ICCV}, 2023.

\bibitem[Zhou et~al.(2021)Zhou, Li, Zhang, Liu, Zhou, and Zhan]{lidar1}
Guoqing Zhou, Chenyang Li, Dianjun Zhang, Dequan Liu, Xiang Zhou, and Jie Zhan.
\newblock Overview of underwater transmission characteristics of oceanic lidar.
\newblock \emph{IEEE Journal of Selected Topics in Applied Earth Observations
  and Remote Sensing}, 14:\penalty0 8144--8159, 2021.

\end{thebibliography}
